\documentclass[]{youtu} 

\usepackage{mathpazo}
\usepackage{graphicx}
\usepackage[numbers]{natbib}
\usepackage{float}
\usepackage{enumitem}
\usepackage{graphicx}
\usepackage[most]{tcolorbox}
\usepackage{subcaption}

\usepackage{tcolorbox}
\usepackage{enumitem}
\usepackage[dvipsnames]{xcolor}
\usepackage{fontawesome}
\usepackage{multirow}
\usepackage{stfloats}
\usepackage{needspace}

\usepackage[utf8]{inputenc} 
\usepackage[T1]{fontenc}    
\usepackage{hyperref}       
\usepackage{url}            
\usepackage{booktabs}       
\usepackage{amsfonts}       
\usepackage{nicefrac}       
\usepackage{microtype}      
\usepackage{xcolor}         
\usepackage{amsmath}
\usepackage{enumitem}
\usepackage[most]{tcolorbox}
\usepackage{makecell}

\usepackage[utf8]{inputenc} 
\usepackage[T1]{fontenc}    
\usepackage{hyperref}       
\usepackage{url}            
\usepackage{booktabs}       
\usepackage{amsfonts}       
\usepackage{nicefrac}       
\usepackage{microtype}      
\usepackage{xcolor}         
\usepackage{amsmath}
\usepackage[most]{tcolorbox}
\usepackage{wrapfig}

\tcbuselibrary{breakable}
\usepackage{listings}
\usepackage{courier}

\definecolor{lightgray}{gray}{0.9}
\definecolor{stepgray}{gray}{0.8}
\definecolor{PineGreen}{HTML}{007C4A} 

\makeatletter
\renewcommand\tableofcontents{%
  \@starttoc{toc}%
}

\let\origaddcontentsline\addcontentsline
\newcommand{\DisableTOC}{\let\addcontentsline\@gobblethree}
\newcommand{\EnableTOC}{\let\addcontentsline\origaddcontentsline}
\makeatother

\usepackage{booktabs}
\usepackage{makecell}
\usepackage{graphicx}
\usepackage{pifont}
\usepackage{adjustbox}
\usepackage{xcolor}
\usepackage{multirow}
\usepackage{caption}
\usepackage{calc} 
\usepackage{tabularx}

\usepackage[table]{xcolor}

\definecolor{headerblue}{RGB}{25, 60, 130}      
\definecolor{rowblueA}{RGB}{245, 248, 255}      
\definecolor{rowblueB}{RGB}{235, 242, 252}      
\definecolor{highlightblue}{RGB}{210, 225, 250} 
\definecolor{checkcolor}{RGB}{40, 120, 200}     
\definecolor{crosscolor}{RGB}{160, 170, 180}    

\definecolor{tableblue}{RGB}{205, 220, 240}
\definecolor{tablegreen}{RGB}{210, 240, 210} 
\definecolor{tablepink}{RGB}{245, 215, 225}

\usepackage{arydshln} 
\definecolor{myborder}{RGB}{73, 86, 102}
\definecolor{myRed}{RGB}{240, 48, 159}
\definecolor{mylightblue}{RGB}{235, 245, 255}

\tcbuselibrary{skins}

\newcommand{\colorcatI}[1]{\cellcolor{cyan!#1!white!60}}   
\newcommand{\colorcatII}[1]{\cellcolor{green!#1!white!60}} 
\newcommand{\colorcatIII}[1]{\cellcolor{orange!#1!white!60}}
\newcommand{\coloravg}[1]{\cellcolor{red!#1!white!60}}    

\title{When Seeing Is Not Believing---A Benchmark for Search-Grounded Video Misinformation Detection}
\author{
Tao Yu\textsuperscript{$1,2,3 * \S \spadesuit$}, 
Yujia Yang\textsuperscript{$2 *$}, 
Shenghua Chai\textsuperscript{$1 *\dag$}, 
Jinshuai Zhang\textsuperscript{$1 *\dag$}, 
Haopeng Jin\textsuperscript{$1 \dag$}, \\
Hao Wang\textsuperscript{$1 \dag$}, 
Minghui Zhang\textsuperscript{$1 \dag$}, 
Zhongtian Luo\textsuperscript{$1 \dag$}, 
Yuchen Long\textsuperscript{$1 \dag$}, 
Xinlong Chen\textsuperscript{$1,2$}, \\
Jiabing Yang\textsuperscript{$1,2$}, 
Zhaolu Kang\textsuperscript{$5$}, 
Yuxuan Zhou\textsuperscript{$4$}, 
Zhengyu Man\textsuperscript{$1 \dag$}, 
Xinming Wang\textsuperscript{$1,2$}, \\
Hongzhu Yi\textsuperscript{$2 \ddagger$}, 
Zheqi He\textsuperscript{$3 \ddagger$}, 
Xi Yang\textsuperscript{$3$}, 
Yan Huang\textsuperscript{$1,2 \ddagger$}, 
Liang Wang\textsuperscript{$1,2$}
}

\vspace{-5mm}

\affiliation{
\textsuperscript{$1$}CASIA \ 
\textsuperscript{$2$}UCAS \ 
\textsuperscript{$3$}BAAI \
\textsuperscript{$4$}Tsinghua University \
\textsuperscript{$5$}Peking University
}

\date{Jun. 2, 2026}
\sourcecode{https://github.com/yutao1024/EVID-Bench}
\data{https://huggingface.co/datasets/Kirito-Lab/EVID-Bench/}

\begin{document}

\DisableTOC

\abstract{Video misinformation increasingly operates at the semantic and evidential level: authentic footage may be selectively edited, temporally reordered, spliced across sources, or augmented with AI-generated content to construct false narratives. Such evidence-dependent manipulations cannot be reliably verified from the input video alone, because the missing, reordered, replaced, or recontextualized evidence lies outside the video itself. We introduce \textbf{EVID-Bench}, a benchmark for search-grounded video misinformation detection, where a system must search the open web for related videos and identify what information is false through cross-video comparison. EVID-Bench comprises 222 videos spanning 9 manipulation types across 3 categories: AI generation, single-source editing, and multi-source editing. All samples are verified to be undetectable by frontier models through visual inspection alone. We evaluate nine frontier multimodal models using a retrieval-augmented verification baseline. The best system achieves only 61.43\% point-level accuracy and 43.24\% video-level accuracy, while AI-generated manipulations remain especially challenging. Error analysis reveals recurring challenges: models fixate on irrelevant anchors, misattribute synthetic content to editorial splicing, and terminate search prematurely before fully explaining the manipulation.}
\maketitle

\renewcommand{\thefootnote}{*}
\footnotetext{Equal contribution.}

\renewcommand{\thefootnote}{\dag}
\footnotetext{Work done during an internship at CASIA.}

\renewcommand{\thefootnote}{\S}
\footnotetext[0]{Work done during an internship at BAAI.}

\renewcommand{\thefootnote}{\textdaggerdbl}
\footnotetext{Corresponding author.}

\renewcommand{\thefootnote}{\ensuremath{\spadesuit}}
\footnotetext{Project leader.}
\renewcommand{\thefootnote}{\arabic{footnote}}

\vspace{-.1em}

\section{Introduction}

Video has become a dominant medium of information on social platforms, news media, and instant messaging applications, making it a highly influential vehicle for misinformation. Compared with text or images, viewers are often more inclined to regard video as a direct record of reality~\cite{gunasekara2025influence,sundar2021seeing}. A carefully manipulated video may shape public opinion, intensify social tensions, or damage individual reputations before factual clarifications can spread.

One particularly dangerous category of video misinformation derives its core harm from the distortion of event semantics and evidential relationships. Authentic footage can be selectively trimmed, temporally reordered, or stitched across sources to construct narratives that never occurred~\cite{diwan2024systematic,wang2025fmnv}; meanwhile, realistic synthetic faces, objects, or event segments can be embedded into real video contexts. For example, removing the trigger of a protest conflict may frame the remaining footage as ``unprovoked violence,'' reordered speech clips may suggest a ``reversal of position,'' and face replacement in a real interview may make viewers believe that another public figure appeared and made the remarks. In such cases, misinformation often parasitizes authentic material: every frame may be real while the relations between segments are false, or only a local identity, object, or event segment may be synthetic while the overall narrative is altered. We refer to this setting as evidence-dependent video misinformation, where authenticity cannot be fully determined from the input video alone but must be judged against external event evidence.

Existing video authenticity detection paradigms are not designed for this setting. Traditional forensic methods~\cite{singh2024copy,fatima2026enhanced} rely on low-level traces such as compression inconsistencies, optical flow irregularities, or pixel statistics, which may not exist when authentic footage is edited or recontextualized. Deepfake detectors~\cite{chou2026improving,qian2020thinkingfrequencyfaceforgery,hernandezortega2020deepfakesonphysdeepfakesdetectionbased} can identify local generative artifacts, but they cannot recover the original content before replacement or explain how a local modification changes identity, causality, or responsibility. Multimodal misinformation methods~\cite{abdali2024multimodalmisinformationdetectionapproaches,tian2026exposingcrossmodalconsistencyfake,shopnil2026meritmodularframeworkmultimodal,aneja2021cosmoscatchingoutofcontextmisinformation} often rely on explicit textual claims, whereas video misinformation can convey false narratives purely through visual editing. Thus, the key challenge is not merely detecting whether a video contains artifacts, but identifying what information has been falsified relative to evidence about the real event. This requires retrieving external video evidence~\cite{yu2026shotfinder,yu2026beyond}
and reasoning over cross-video discrepancies. Recent findings that non-adversarial noise can reduce the benefits of agent
skill self-evolution~\cite{yu2026nonadversarialnoisehurtslearning} further
motivate examining how verification agents handle imperfect retrieved evidence.

To address this challenge, we introduce \textbf{EVID-Bench}, a benchmark for \textit{search-grounded video misinformation detection}. Given a potentially manipulated video, a system must search the open web for related or corroborating videos and identify what information is false through cross-video evidence comparison. This formulation mirrors how human fact-checkers debunk video misinformation: by finding related footage and showing what was omitted, reordered, inserted, or recontextualized.
\begin{figure*}[!b]
    \centering
    \includegraphics[width=\linewidth]{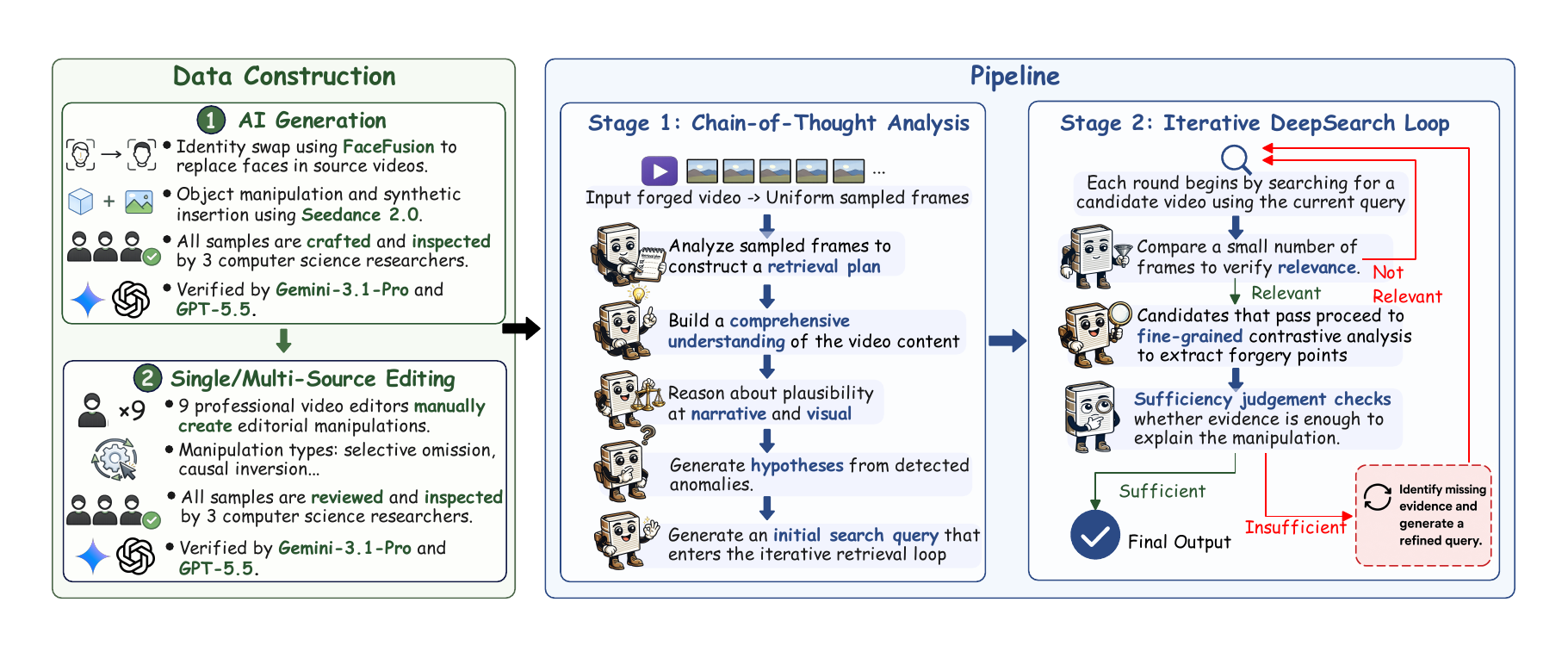}
    \vspace{-20pt}
    \caption{Overview of EVID-Bench construction and the search-grounded video misinformation detection pipeline. The benchmark constructs hard-to-detect video misinformation through AI generation and professional single-source and multi-source editing, followed by human inspection and model-based verification. The detection pipeline analyzes sampled frames, forms retrieval hypotheses, iteratively searches for external video evidence, compares candidate sources with the input video, and refines the search until sufficient evidence is obtained for the final misinformation explanation.}
    \label{fig:1}
    \vspace{-10pt}
\end{figure*}

EVID-Bench contains 222 manipulated videos spanning 9 forgery types in 3 categories: AI generation, single-source editing, and multi-source editing, across 6 real-world topic domains. All samples undergo human review and model-based filtering to ensure that frontier multimodal models cannot identify the manipulation through visual inspection alone. We further propose a retrieval-augmented verification baseline with chain-of-thought analysis and an iterative search-verify-reflect loop. Experiments on nine frontier models show that EVID-Bench remains challenging: the best model achieves only 61.43\% point-level accuracy and 43.24\% video-level accuracy, with AI-generated manipulations remaining especially difficult.

Our contributions are as follows:
\begin{itemize}
    \item We formalize search-grounded video misinformation detection, where systems must identify false information through external video retrieval and cross-video reasoning.
    \item We construct EVID-Bench, a quality-assured benchmark of 222 videos covering 9 forgery types and 6 topic domains.
    \item We evaluate nine frontier models with a retrieval-augmented verification baseline, showing that even the best system achieves only 43.24\% video-level accuracy and revealing key challenges in retrieval, contrastive reasoning, and sufficiency judgment.
\end{itemize}

\section{Related Works}

\subsection{Video Forgery Detection}

Traditional methods detect low-level artifacts: compression inconsistencies~\cite{singh2024copy}, irregular optical flow~\cite{fatima2026enhanced}, or face-swap traces through physiological signals and frequency analysis~\cite{chou2026improving,qian2020thinkingfrequencyfaceforgery,hernandezortega2020deepfakesonphysdeepfakesdetectionbased}. These assume manipulation leaves detectable pixel-level traces---an assumption that breaks down for semantic-level forgeries where every frame is authentic and the manipulation lies in editorial choices. EVID-Bench targets this class of forgeries, where detection requires external evidence.

\subsection{Multimodal Misinformation Detection}

Another line of work verifies textual claims against visual content~\cite{abdali2024multimodalmisinformationdetectionapproaches,tian2026exposingcrossmodalconsistencyfake}, retrieves textual evidence for fact-checking~\cite{shopnil2026meritmodularframeworkmultimodal}, or detects out-of-context media~\cite{aneja2021cosmoscatchingoutofcontextmisinformation}. These face two limitations: most assume an explicit textual claim, whereas manipulated videos convey false narratives purely through visual editing; and retrieval operates in the text domain rather than comparing video evidence directly. EVID-Bench requires searching for corroborating videos and performing cross-video contrastive reasoning.

\section{EVID-Bench}

\subsection{Task Definition and Design Principles} 

We introduce \textbf{search-grounded video misinformation detection} as the core task of EVID-Bench. Given a potentially misleading video, a system must search the open web for original or related videos and identify what information is false through cross-video evidence comparison. The task is designed around three principles:

\textbf{Evidence-dependent misinformation.} Misleading videos in our benchmark distort event meaning by altering identities, reordering events, changing context, or embedding AI-generated content into authentic footage. Such misinformation cannot be reliably verified from the input video alone, because the missing, replaced, reordered, or recontextualized evidence lies outside the video itself.

\textbf{Reasoning-driven search.} Since the input video may encode a false narrative, naively summarizing it into search queries risks searching for the misinformation rather than the truth. The system must reason about which elements are likely authentic and searchable, use them as retrieval anchors, and iteratively refine its queries as partial evidence is gathered.

\textbf{Cross-video contrastive verification.} The final judgment must be grounded in differences between the input video and retrieved evidence. This requires the system to align videos at the level of entities, actions, temporal order, and narrative context, and identify where they diverge. The output is not a binary label, but a structured explanation of what is false and which evidence supports that conclusion.

Together, these principles make EVID-Bench a joint evaluation of video understanding, web search, and cross-video reasoning---capabilities not jointly assessed by existing forensics or misinformation benchmarks.

\begin{figure*}[!b]
    \centering
    \includegraphics[width=0.85\linewidth]{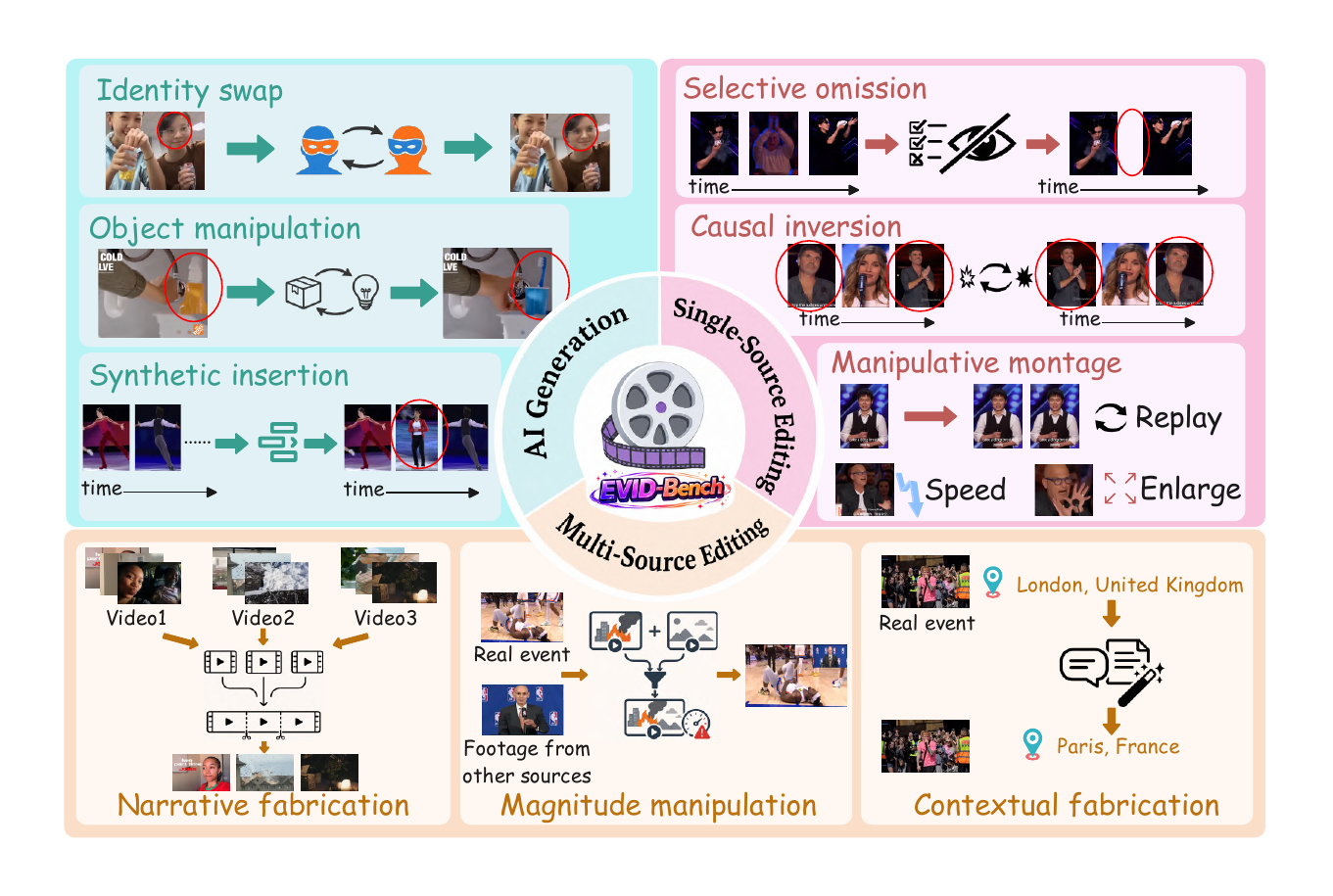}
    \vspace{-10pt}
    \caption{Taxonomy of search-grounded video misinformation in EVID-Bench covering AI Generation, Single-Source Editing, and Multi-Source Editing. These mechanisms manipulate identities, objects, temporal order, narrative structure, event magnitude, or contextual information, making the video misleading not necessarily through visible artifacts, but through its discrepancy from external event evidence.}
    \label{fig:2}
    \vspace{-10pt}
\end{figure*}

\subsection{Task Taxonomy} 
Drawing on prior work~\cite{rossler2019faceforensics++,ho2022video,fischinger2025df,wardle2017information,gangal2022nareor,yin2024text,pei2024multi,bu2023combating,luo2021newsclippings}, we organize manipulations into three categories based on the forgery mechanism, each with three subcategories (nine task types in total). 

\subsubsection{AI Generation} 

Synthetic content is introduced into authentic footage. 

\textbf{Identity swap.} A person's face is replaced with a different identity using deepfake techniques while the body, background, and context remain authentic. 

\textbf{Synthetic insertion.} A contiguous AI-generated video segment is embedded into real footage, introducing an entirely fabricated temporal sequence of events. 

\textbf{Object manipulation.} Objects within the scene are added, removed, or replaced through generative editing, altering what is physically present without changing the overall spatial structure.

\subsubsection{Single-Source Editing} 

A single authentic video is editorially altered to change its narrative meaning. 

\textbf{Selective omission.} Critical portions are removed so that the remaining footage conveys a different or opposite meaning from the original. 

\textbf{Causal inversion.} The temporal order of events is reordered to flip causality---e.g., transforming a victim's retaliation into an unprovoked attack by swapping the sequence of actions. 

\textbf{Manipulative montage.} Shots are rearranged and re-paced to create an emotional or narrative impression that the original footage does not support, without reversing any specific causal sequence. 

\subsubsection{Multi-Source Editing} 

Footage from multiple authentic videos is combined to construct false information. 

\textbf{Narrative fabrication.} Segments from unrelated videos are spliced together to tell a story that none of the individual sources depict. 

\textbf{Magnitude manipulation.} A real event is combined with footage from other sources to distort its severity---either exaggerating or downplaying its actual impact. 

\textbf{Contextual fabrication.} Authentic footage of a real event is placed into a false context (different time, location, or attribution) through splicing, misleading viewers about when, where, or to whom the event occurred.


























\subsection{Topic Categories} 
To ensure our benchmark reflects the domains where video misinformation most commonly arises, we select six topic categories based on prior studies~\cite{papadopoulou2019corpus,wang2024official,huang2026probing,bu2023combating,wang2025fmnv,bevendorff2024product} on video-based false information: 

\textbf{Daily life.} Vlogs, workplace recordings, campus life, travel clips, food, and pet videos. 

\textbf{Education \& skills.} Online tutorials, DIY crafts, fashion and beauty, fitness, and parenting content. 

\textbf{On-Scene footage.} Dashcam and surveillance recordings, disaster scenes, and live incident documentation. 

\textbf{Sports.} Professional and amateur athletic events, match highlights, and training footage. 

\textbf{Performance \& talent.} Music, dance, comedy, and other talent showcases. 

\textbf{Reviews.} Product unboxings, brand comparisons, and store/restaurant visits.

\begin{wrapfigure}{r}{0.5\textwidth} 
    \centering
    \includegraphics[width=\linewidth]{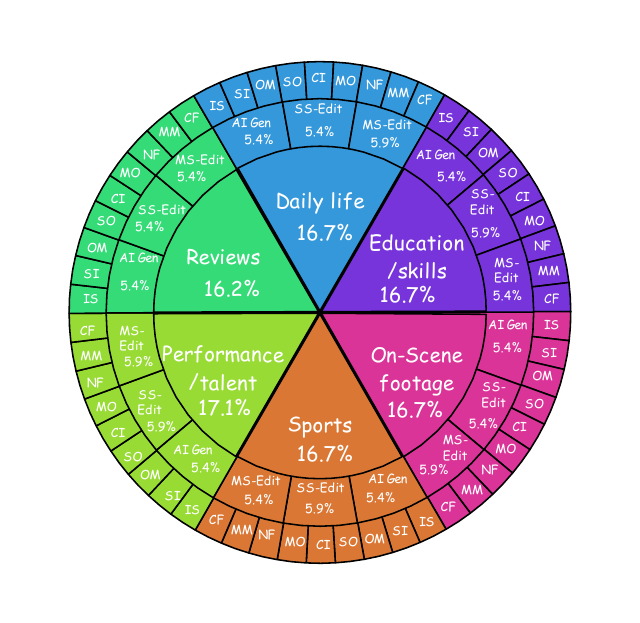} 
    \vspace{-15pt}
    \caption{Distribution of EVID-Bench across task types and topics. Task types are abbreviated as follows: AI Gen (AI generation), SS-Edit (Single-Source Editing), MS-Edit (Multi-Source Editing), IS (Identity Swap), SI (Synthetic Insertion), OM (Object Manipulation), SO (Selective Omission), CI (Causal Inversion), MO (Manipulative Montage), NF (Narrative Fabrication), MM (Magnitude Manipulation), and CF (Contextual Fabrication).}
    \label{fig:4}
    \vspace{-50pt}
\end{wrapfigure}%
\subsection{Data Construction} 

\subsubsection{AI Generation} 

For identity swap, we use FaceFusion\footnote{\url{https://github.com/facefusion/facefusion}} to replace faces in source videos. For object manipulation and synthetic insertion, we generate content using Seedance 2.0 \cite{seedance2026seedance20advancingvideo}. All AI-generated samples are manually crafted and inspected by three computer science researchers to ensure visual coherence. Each sample is then verified by Gemini-3.1-Pro \cite{gemini3pro2025} and GPT-5.5 \cite{singh2025openai}: we confirm that neither model can identify the manipulation through direct visual reasoning over the forged video alone. 

\subsubsection{Single-Source and Multi-Source Editing} 

We recruit nine professional video editors to manually produce all editorial manipulations, including selective omission, causal inversion, manipulative montage, narrative fabrication, magnitude manipulation, and contextual fabrication. Each edited video is first reviewed by three computer science researchers to ensure the manipulation is coherent and undetectable. We then present the edited videos to Gemini-3.1-Pro and GPT-5.5, verifying that the models interpret the video content consistently with the intended manipulated narrative---confirming that the editorial forgery successfully conveys its false information without raising suspicion. 

\subsubsection{Quality Assurance} 


Our verification process combines human review and model-based filtering. Three computer science researchers first examine each sample for technical quality and narrative coherence; we then conduct visual verification using Gemini-3.1-Pro and GPT-5.5. For AI-generated forgeries, both models are required to fail to identify the manipulation; for editorial forgeries, both models are required to interpret the video according to the intended false narrative. Samples that fail either stage are revised or discarded, ensuring that the final 222 samples cannot be identified as manipulated by frontier models through visual analysis alone. The data generation process is shown in Figures \ref{fig:1} and \ref{fig:2}. Details of this process and the prompts used are provided in Appendix \ref{sec:appendix_verification}.

\subsection{Data Statistics}

Our benchmark comprises 222 manipulated videos spanning 3 major categories, 9 task types, and 6 topic categories. Figure~\ref{fig:4} shows the distribution across task types and topics. A detailed description of the data is provided in Appendix \ref{sec:appendix_data}.

\section{Experiments}

\subsection{Baseline}

\label{1}

\textbf{Stage 1: Chain-of-Thought Analysis.} A VLM analyzes uniformly sampled frames via chain-of-thought reasoning to construct a retrieval plan. The fundamental challenge is that the video's apparent narrative cannot be taken at face value---it may be the product of manipulation. The model first builds a comprehensive understanding of the video content: identifying key entities (people, locations, events, temporal cues), segmenting distinct scenes, and detecting discontinuities in environment, production style, or visual quality that suggest assembly from multiple sources. It then reasons about plausibility at two levels---at the narrative level, whether the implied facts are internally consistent and whether the associations between entities across scenes are natural or artificially constructed; at the visual level, whether any elements exhibit signs of AI generation such as contextually incongruent objects or inconsistent facial identity between segments. These detected anomalies inform hypothesis generation: the model infers what the unmanipulated reality likely depicts and what claims most urgently require external verification. Suspected synthetic elements are explicitly excluded from retrieval anchors, ensuring that queries are grounded in authentic, verifiable cues rather than potentially fabricated content. The output is an initial search query that enters the iterative retrieval loop.

\textbf{Stage 2: Iterative DeepSearch Loop.} The system enters an iterative retrieval loop. Each round begins by searching for a candidate video using the current query. To avoid expensive analysis on irrelevant results, a coarse relevance filter first compares a small number of frames from both videos, assessing whether they share the same event, scene family, or verification-relevant content. Candidates that pass proceed to fine-grained contrastive analysis, where a larger set of frames from both videos is compared to extract narrative-level forgery points---structured descriptions of how the input video distorts, reframes, or fabricates information relative to the retrieved evidence. Candidates that fail are discarded, and the model directly generates a new query for the next round. After each successful comparison, a sufficiency judgment evaluates whether the accumulated evidence adequately explains the manipulation. If not, the model identifies what evidence is still missing and generates a refined query to target that gap. 
\begin{wrapfigure}{r}{0.6\textwidth} 
    \centering
    \includegraphics[width=\linewidth]{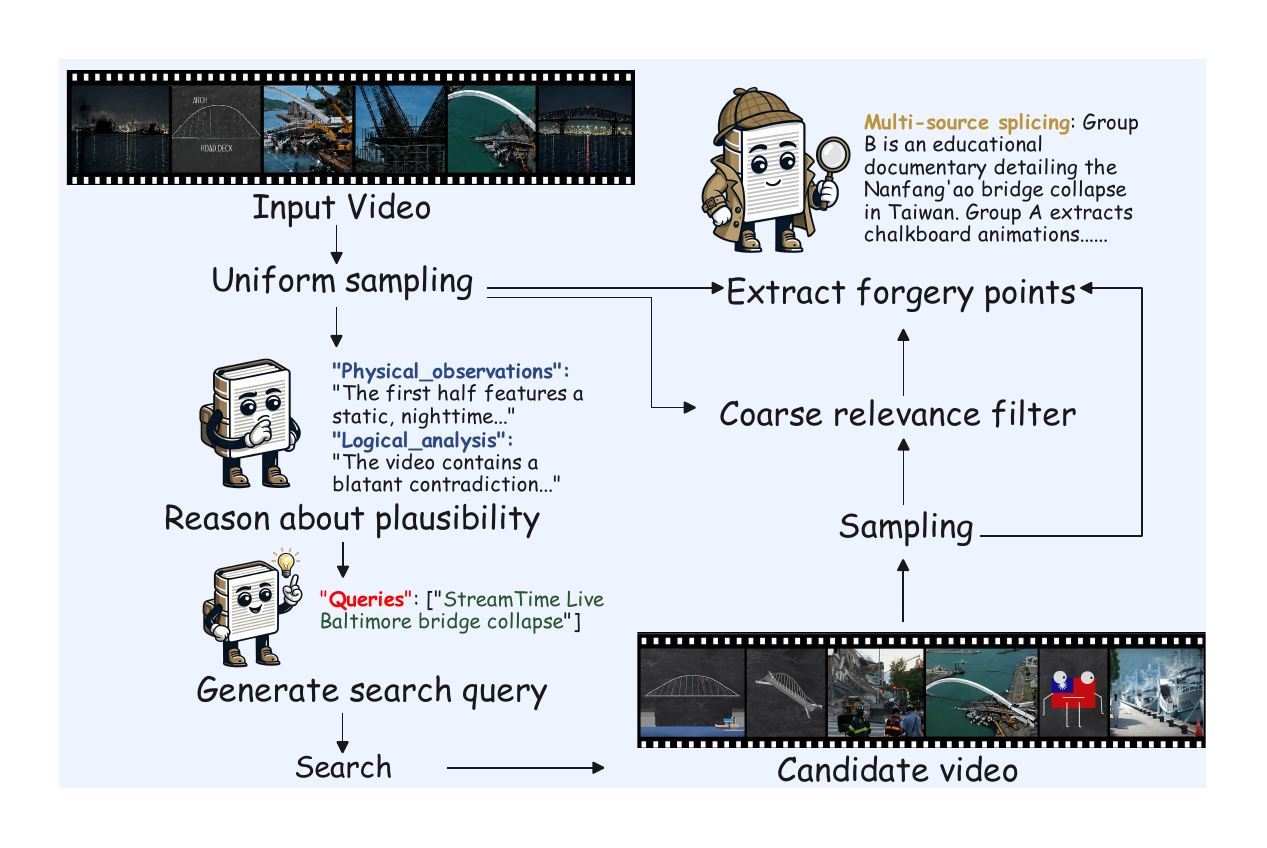}
    \vspace{-20pt}
    \caption{Case study of EVID-Bench.}
    \label{fig:3}
    \vspace{-60pt}
\end{wrapfigure}
This iterative search-verify-reflect cycle continues until the evidence is deemed sufficient or the round budget is exhausted. A case study is shown in Figure \ref{fig:3}. Details of this part and the prompts used are provided in Appendix \ref{sec:appendix_pipeline}.

\subsection{Evaluation Metrics and Settings}

\textbf{Metrics.}
We evaluate performance at two levels. \textbf{Point-level accuracy} measures the fraction of ground-truth misinformation points correctly identified by the system, where each video contains 3--5 annotated points. \textbf{Video-level accuracy} counts a video as correct only when all ground-truth points are identified, measuring whether the system fully explains the manipulation.

\textbf{Experimental setup.}
We evaluate nine frontier multimodal models: GPT-5.5, GPT-5.4~\cite{singh2025openai5.4}, GPT-5.4-mini~\cite{singh2025openai5.4}, Claude Opus 4.6~\cite{claudeops}, Claude Sonnet 4.6~\cite{claudeson}, Gemini-3.1-Pro, Gemini-3-Flash~\cite{gemini3pro2025}, Qwen3-VL-235B-A22B-Instruct~\cite{bai2025qwen3vltechnicalreport}, and Qwen-3.5-Plus~\cite{qwen3.5}. Parameter details are provided in Appendix \ref{sec:appendix_setup}.

\textbf{Evaluation protocol.}
Because outputs are natural-language descriptions, we use three LLM judges---GPT-5.4, Gemini-3.1-Pro, and Claude Sonnet 4.6---to assess whether each predicted point matches a ground-truth point. Correctness is decided by majority vote. The similarity analysis between LLM-based evaluation and human evaluation is provided in Appendix \ref{sec:appendix_human_agreement}.

\begin{table*}[!b]
\centering
\caption{Point-level accuracy of evaluated models across fine-grained manipulation task types. All metrics are reported in percentages (\%).}
\vspace{-8pt}
\label{tab:1}
\setlength{\tabcolsep}{6pt} 
\renewcommand{\arraystretch}{1.3}
\resizebox{\textwidth}{!}{
\begin{tabular}{l *{9}{c} c}
\toprule
\multirow{3}{*}{\textbf{Model}} & \multicolumn{3}{c}{\textbf{\makecell{AI Generation}}} & \multicolumn{3}{c}{\textbf{\makecell{Single-Source Editing}}} & \multicolumn{3}{c}{\textbf{\makecell{Multi-Source Editing}}} & \multirow{3}{*}{\textbf{Avg.}} \\
\cmidrule(lr){2-4} \cmidrule(lr){5-7} \cmidrule(lr){8-10}
& \textbf{\makecell{Identity\\Swap}} & \textbf{\makecell{Synthetic\\Insertion}} & \textbf{\makecell{Object\\Manipulation}} & \textbf{\makecell{Selective\\Omission}} & \textbf{\makecell{Causal\\Inversion}} & \textbf{\makecell{Manipulative\\Montage}} & \textbf{\makecell{Narrative\\Fabrication}} & \textbf{\makecell{Magnitude\\Manipulation}} & \textbf{\makecell{Contextual\\Fabrication}} & \\
\midrule
\multicolumn{11}{l}{\textbf{GPT Series}} \\
\addlinespace[3pt]
\hdashline
\addlinespace[3pt]
GPT-5.5 & \colorcatI{48}11.11 ($\pm$2.1) & \colorcatI{100}16.67 ($\pm$1.6) & \colorcatI{100}13.89 ($\pm$0.8) & \colorcatII{100}72.92 ($\pm$1.6) & \colorcatII{100}55.32 ($\pm$1.2) & \colorcatII{100}74.00 ($\pm$1.5) & \colorcatIII{100}82.24 ($\pm$1.0) & \colorcatIII{100}98.91 ($\pm$0.0) & \colorcatIII{84}88.46 ($\pm$0.6) & \coloravg{100}61.43 ($\pm$0.9) \\
GPT-5.4 & \colorcatI{20}0.00 ($\pm$0.0) & \colorcatI{56}9.72 ($\pm$1.6) & \colorcatI{80}11.11 ($\pm$0.8) & \colorcatII{79}64.58 ($\pm$1.6) & \colorcatII{77}45.74 ($\pm$0.0) & \colorcatII{77}66.00 ($\pm$0.6) & \colorcatIII{79}73.83 ($\pm$1.7) & \colorcatIII{93}95.65 ($\pm$1.1) & \colorcatIII{60}78.85 ($\pm$1.5) & \coloravg{73}53.77 ($\pm$0.9) \\
GPT-5.4-Mini & \colorcatI{23}1.39 ($\pm$0.0) & \colorcatI{38}6.94 ($\pm$0.0) & \colorcatI{40}5.56 ($\pm$1.4) & \colorcatII{60}57.29 ($\pm$1.0) & \colorcatII{67}41.49 ($\pm$1.1) & \colorcatII{54}58.00 ($\pm$1.7) & \colorcatIII{55}64.49 ($\pm$1.7) & \colorcatIII{71}85.87 ($\pm$1.3) & \colorcatIII{32}67.31 ($\pm$2.2) & \coloravg{49}46.97 ($\pm$1.0) \\
\midrule
\multicolumn{11}{l}{\textbf{Claude Series}} \\
\addlinespace[3pt]
\hdashline
\addlinespace[3pt]
Claude-Opus-4.6 & \colorcatI{41}8.33 ($\pm$2.1) & \colorcatI{82}13.89 ($\pm$0.8) & \colorcatI{80}11.11 ($\pm$1.6) & \colorcatII{92}69.79 ($\pm$1.8) & \colorcatII{95}53.19 ($\pm$0.6) & \colorcatII{89}70.00 ($\pm$0.6) & \colorcatIII{86}76.64 ($\pm$1.5) & \colorcatIII{95}96.74 ($\pm$0.6) & \colorcatIII{79}86.54 ($\pm$0.6) & \coloravg{89}58.34 ($\pm$1.0) \\
Claude-Sonnet-4.6 & \colorcatI{34}5.56 ($\pm$0.8) & \colorcatI{73}12.50 ($\pm$1.6) & \colorcatI{70}9.72 ($\pm$1.6) & \colorcatII{84}66.67 ($\pm$1.2) & \colorcatII{87}50.00 ($\pm$0.6) & \colorcatII{80}67.00 ($\pm$2.3) & \colorcatIII{76}72.90 ($\pm$1.0) & \colorcatIII{90}94.57 ($\pm$1.3) & \colorcatIII{62}79.81 ($\pm$1.0) & \coloravg{78}55.13 ($\pm$1.0) \\
\midrule
\multicolumn{11}{l}{\textbf{Gemini Series}} \\
\addlinespace[3pt]
\hdashline
\addlinespace[3pt]
Gemini-3.1-Pro & \colorcatI{100}30.60 ($\pm$1.8) & \colorcatI{82}13.90 ($\pm$1.5) & \colorcatI{70}9.70 ($\pm$1.1) & \colorcatII{87}67.70 ($\pm$0.8) & \colorcatII{92}52.10 ($\pm$0.8) & \colorcatII{77}66.00 ($\pm$1.6) & \colorcatIII{72}71.00 ($\pm$1.3) & \colorcatIII{64}82.60 ($\pm$2.4) & \colorcatIII{59}77.90 ($\pm$1.1) & \coloravg{81}55.90 ($\pm$1.0) \\
Gemini-3-Flash & \colorcatI{55}13.90 ($\pm$1.7) & \colorcatI{65}11.10 ($\pm$1.9) & \colorcatI{20}2.80 ($\pm$0.0) & \colorcatII{81}65.60 ($\pm$0.6) & \colorcatII{97}54.30 ($\pm$1.3) & \colorcatII{89}70.00 ($\pm$1.1) & \colorcatIII{76}72.90 ($\pm$2.2) & \colorcatIII{59}80.40 ($\pm$1.3) & \colorcatIII{52}75.00 ($\pm$1.1) & \coloravg{72}53.60 ($\pm$1.0) \\
\midrule
\multicolumn{11}{l}{\textbf{Qwen Series}} \\
\addlinespace[3pt]
\hdashline
\addlinespace[3pt]
Qwen-3.5-Plus & \colorcatI{76}22.20 ($\pm$1.3) & \colorcatI{82}13.90 ($\pm$0.0) & \colorcatI{70}9.70 ($\pm$0.9) & \colorcatII{63}58.30 ($\pm$1.6) & \colorcatII{75}44.70 ($\pm$2.0) & \colorcatII{71}64.00 ($\pm$2.1) & \colorcatIII{79}73.80 ($\pm$1.5) & \colorcatIII{76}88.00 ($\pm$0.7) & \colorcatIII{48}74.00 ($\pm$0.6) & \coloravg{72}53.40 ($\pm$1.0) \\
Qwen3-VL-235B-A22B-Instruct & \colorcatI{37}6.90 ($\pm$1.6) & \colorcatI{73}12.50 ($\pm$1.0) & \colorcatI{30}4.20 ($\pm$1.8) & \colorcatII{71}61.50 ($\pm$1.9) & \colorcatII{72}43.60 ($\pm$0.7) & \colorcatII{60}60.00 ($\pm$0.7) & \colorcatIII{72}71.00 ($\pm$1.5) & \colorcatIII{44}73.90 ($\pm$0.7) & \colorcatIII{41}71.20 ($\pm$1.1) & \coloravg{56}48.80 ($\pm$1.0) \\
\bottomrule
\end{tabular}
}
\vspace{-8pt}
\end{table*}

\subsection{Main Results}

\textbf{EVID-Bench remains challenging for frontier models.}
The best model, GPT-5.5, achieves 61.43\% point-level but only 43.24\% video-level accuracy (Tables~\ref{tab:1} and~\ref{tab:2}). Claude Opus 4.6 and Gemini-3.1-Pro follow at 58.34\%/38.74\% and 55.90\%/36.90\%. Even frontier models only partially identify misinformation in most videos.

\textbf{Complete verification is substantially harder than partial detection.}
All models show a clear gap between point-level and video-level accuracy. Models can retrieve relevant evidence and identify some manipulation points but struggle to integrate multiple pieces of evidence and fully explain all false information.

\begin{table*}[t]
\centering
\caption{Video-level accuracy of evaluated models across fine-grained manipulation task types. All metrics are reported in percentages (\%).}
\vspace{-8pt}
\label{tab:2}
\setlength{\tabcolsep}{6pt} 
\renewcommand{\arraystretch}{1.3}
\resizebox{\textwidth}{!}{
\begin{tabular}{l *{9}{c} c}
\toprule
\multirow{3}{*}{\textbf{Model}} & \multicolumn{3}{c}{\textbf{\makecell{AI Generation}}} & \multicolumn{3}{c}{\textbf{\makecell{Single-Source Editing}}} & \multicolumn{3}{c}{\textbf{\makecell{Multi-Source Editing}}} & \multirow{3}{*}{\textbf{Avg.}} \\
\cmidrule(lr){2-4} \cmidrule(lr){5-7} \cmidrule(lr){8-10}
& \textbf{\makecell{Identity\\Swap}} & \textbf{\makecell{Synthetic\\Insertion}} & \textbf{\makecell{Object\\Manipulation}} & \textbf{\makecell{Selective\\Omission}} & \textbf{\makecell{Causal\\Inversion}} & \textbf{\makecell{Manipulative\\Montage}} & \textbf{\makecell{Narrative\\Fabrication}} & \textbf{\makecell{Magnitude\\Manipulation}} & \textbf{\makecell{Contextual\\Fabrication}} & \\
\midrule
\multicolumn{11}{l}{\textbf{GPT Series}} \\
\addlinespace[3pt]
\hdashline
\addlinespace[3pt]
GPT-5.5 & \colorcatI{36}4.17 ($\pm$0.0) & \colorcatI{100}8.33 ($\pm$0.0) & \colorcatI{100}4.17 ($\pm$0.0) & \colorcatII{100}60.00 ($\pm$0.0) & \colorcatII{80}36.00 ($\pm$2.3) & \colorcatII{100}52.00 ($\pm$2.3) & \colorcatIII{91}51.85 ($\pm$2.2) & \colorcatIII{100}86.96 ($\pm$0.0) & \colorcatIII{92}84.00 ($\pm$0.0) & \coloravg{100}43.24 ($\pm$0.7) \\
GPT-5.4 & \colorcatI{20}0.00 ($\pm$0.0) & \colorcatI{60}4.17 ($\pm$0.0) & \colorcatI{100}4.17 ($\pm$0.0) & \colorcatII{73}48.00 ($\pm$0.0) & \colorcatII{70}32.00 ($\pm$2.3) & \colorcatII{70}40.00 ($\pm$2.3) & \colorcatIII{64}40.74 ($\pm$2.2) & \colorcatIII{87}78.26 ($\pm$1.3) & \colorcatIII{68}72.00 ($\pm$2.3) & \coloravg{75}35.59 ($\pm$0.7) \\
GPT-5.4-Mini & \colorcatI{20}0.00 ($\pm$0.0) & \colorcatI{60}4.17 ($\pm$0.0) & \colorcatI{20}0.00 ($\pm$0.0) & \colorcatII{47}36.00 ($\pm$0.0) & \colorcatII{40}20.00 ($\pm$0.0) & \colorcatII{40}28.00 ($\pm$0.0) & \colorcatIII{38}29.63 ($\pm$2.2) & \colorcatIII{60}60.87 ($\pm$1.3) & \colorcatIII{36}56.00 ($\pm$0.0) & \coloravg{44}26.13 ($\pm$0.3) \\
\midrule
\multicolumn{11}{l}{\textbf{Claude Series}} \\
\addlinespace[3pt]
\hdashline
\addlinespace[3pt]
Claude-Opus-4.6 & \colorcatI{36}4.17 ($\pm$0.0) & \colorcatI{60}4.17 ($\pm$0.0) & \colorcatI{100}4.17 ($\pm$0.0) & \colorcatII{82}52.00 ($\pm$0.0) & \colorcatII{70}32.00 ($\pm$0.0) & \colorcatII{80}44.00 ($\pm$0.0) & \colorcatIII{82}48.15 ($\pm$0.0) & \colorcatIII{87}78.26 ($\pm$1.3) & \colorcatIII{84}80.00 ($\pm$0.0) & \coloravg{85}38.74 ($\pm$0.3) \\
Claude-Sonnet-4.6 & \colorcatI{36}4.17 ($\pm$0.0) & \colorcatI{60}4.17 ($\pm$0.0) & \colorcatI{20}0.00 ($\pm$0.0) & \colorcatII{73}48.00 ($\pm$0.0) & \colorcatII{60}28.00 ($\pm$0.0) & \colorcatII{70}40.00 ($\pm$2.0) & \colorcatIII{64}40.74 ($\pm$0.0) & \colorcatIII{80}73.91 ($\pm$1.3) & \colorcatIII{60}68.00 ($\pm$0.0) & \coloravg{70}34.23 ($\pm$0.7) \\
\midrule
\multicolumn{11}{l}{\textbf{Gemini Series}} \\
\addlinespace[3pt]
\hdashline
\addlinespace[3pt]
Gemini-3.1-Pro & \colorcatI{100}20.80 ($\pm$0.0) & \colorcatI{60}4.20 ($\pm$0.0) & \colorcatI{20}0.00 ($\pm$0.0) & \colorcatII{73}48.00 ($\pm$0.0) & \colorcatII{80}36.00 ($\pm$0.0) & \colorcatII{80}44.00 ($\pm$2.1) & \colorcatIII{64}40.70 ($\pm$1.3) & \colorcatIII{67}65.20 ($\pm$1.6) & \colorcatIII{68}72.00 ($\pm$1.3) & \coloravg{79}36.90 ($\pm$1.4) \\
Gemini-3-Flash & \colorcatI{36}4.20 ($\pm$1.3) & \colorcatI{20}0.00 ($\pm$0.0) & \colorcatI{20}0.00 ($\pm$0.0) & \colorcatII{73}48.00 ($\pm$0.0) & \colorcatII{100}44.00 ($\pm$0.0) & \colorcatII{70}40.00 ($\pm$2.3) & \colorcatIII{56}37.00 ($\pm$0.0) & \colorcatIII{67}65.20 ($\pm$1.3) & \colorcatIII{68}72.00 ($\pm$0.0) & \coloravg{72}34.70 ($\pm$0.3) \\
\midrule
\multicolumn{11}{l}{\textbf{Qwen Series}} \\
\addlinespace[3pt]
\hdashline
\addlinespace[3pt]
Qwen-3.5-Plus & \colorcatI{100}20.80 ($\pm$0.0) & \colorcatI{100}8.30 ($\pm$0.0) & \colorcatI{20}0.00 ($\pm$0.0) & \colorcatII{47}36.00 ($\pm$0.0) & \colorcatII{70}32.00 ($\pm$0.0) & \colorcatII{60}36.00 ($\pm$2.3) & \colorcatIII{73}44.40 ($\pm$2.2) & \colorcatIII{73}69.60 ($\pm$1.3) & \colorcatIII{44}60.00 ($\pm$0.0) & \coloravg{70}34.20 ($\pm$1.0) \\
Qwen3-VL-235B-A22B-Instruct & \colorcatI{20}0.00 ($\pm$0.0) & \colorcatI{20}0.00 ($\pm$0.0) & \colorcatI{20}0.00 ($\pm$0.0) & \colorcatII{56}40.00 ($\pm$0.0) & \colorcatII{60}28.00 ($\pm$0.0) & \colorcatII{90}48.00 ($\pm$0.0) & \colorcatIII{100}55.60 ($\pm$2.0) & \colorcatIII{73}69.60 ($\pm$0.0) & \colorcatIII{36}56.00 ($\pm$2.4) & \coloravg{67}33.30 ($\pm$0.6) \\
\bottomrule
\end{tabular}
}
\vspace{-8pt}
\end{table*}

\textbf{AI-generated manipulations are largely unsolved.}
AI Generation is extremely difficult: GPT-5.5 achieves only 11--17\% point-level and near-zero video-level accuracy (Tables~\ref{tab:1} and~\ref{tab:2}). While the pipeline can retrieve related videos, identifying which specific identity, object, or segment was replaced through cross-video comparison remains unsolved.

\textbf{Multi-Source Editing benefits from retrieval but is not solved.}
Models achieve higher scores on Multi-Source Editing tasks (GPT-5.5: 82--99\% point-level). Cross-source splicing provides stronger retrieval anchors, but video-level accuracy remains much lower, indicating partial rather than complete explanation.

\textbf{Topic difficulty reflects retrieval anchor availability.}
Daily Life and Reviews are the hardest domains, while On-Scene Footage and Education \& Skills achieve higher accuracy (Tables~\ref{tab:3} and~\ref{tab:4}), reflecting the availability of identifiable events or named individuals versus generic content.

\textbf{Human performance.}
On a 54-video subset, three researchers with video-editing experience achieve 57.50\% point-level and 37.04\% video-level accuracy under a search-allowed setting. Performance is lowest on Multi-Source Editing, with 48.00\% point-level and 27.78\% video-level accuracy, showing that cross-source verification remains challenging even for experienced human evaluators. Detailed results are provided in Table~\ref{tab:human-performance} of Appendix~\ref{sec:human-performance-baseline}.

\textbf{Authentic-video evaluation.}
On 60 topic-matched authentic videos, Gemini-3.1-Pro correctly leaves 83.3\% unflagged, while Claude Opus 4.6 and Claude Sonnet 4.6 both achieve 95.0\%, corresponding to false-positive rates of 16.7\%, 5.0\%, and 5.0\%, respectively. False positives occur more frequently on Daily Life and Reviews, suggesting that routine cuts and transitions may trigger over-flagging. Detailed results are provided in Table~\ref{tab:authentic-video-results} of Appendix~\ref{sec:authentic-video-evaluation}.

\textbf{Visual-only forensic baselines.}
We evaluate AIDE~\cite{yan2025sanity} and UnivFD~\cite{ojha2024universal} on all 222 manipulated videos and 60 authentic controls. Mean and majority aggregation produce near-zero detection rates, while max aggregation increases detection but also flags 55.0\% and 78.3\% of authentic videos, respectively. These results show that the tested forensic detectors cannot reliably distinguish semantic manipulations from authentic videos. Detailed results are provided in Table~\ref{tab:forensic-detector-results} of Appendix~\ref{sec:forensic-detector-baselines}.

\textbf{Reverse-image retrieval baseline.}
We compare our iterative text retrieval with Google Lens using 16 uniformly sampled frames and the top-10 results per frame. Despite examining up to 160 URLs per video, Google Lens achieves only 26\% source recall, whereas the best text-based retrieval reaches 60\% with at most six candidate videos. Complete results are provided in Table~\ref{tab:source-recall-results} of Appendix~\ref{sec:reverse-image-search}.






\subsection{Ablation Study}

\begin{figure*}[t]
  \centering
  \includegraphics[width=\linewidth]{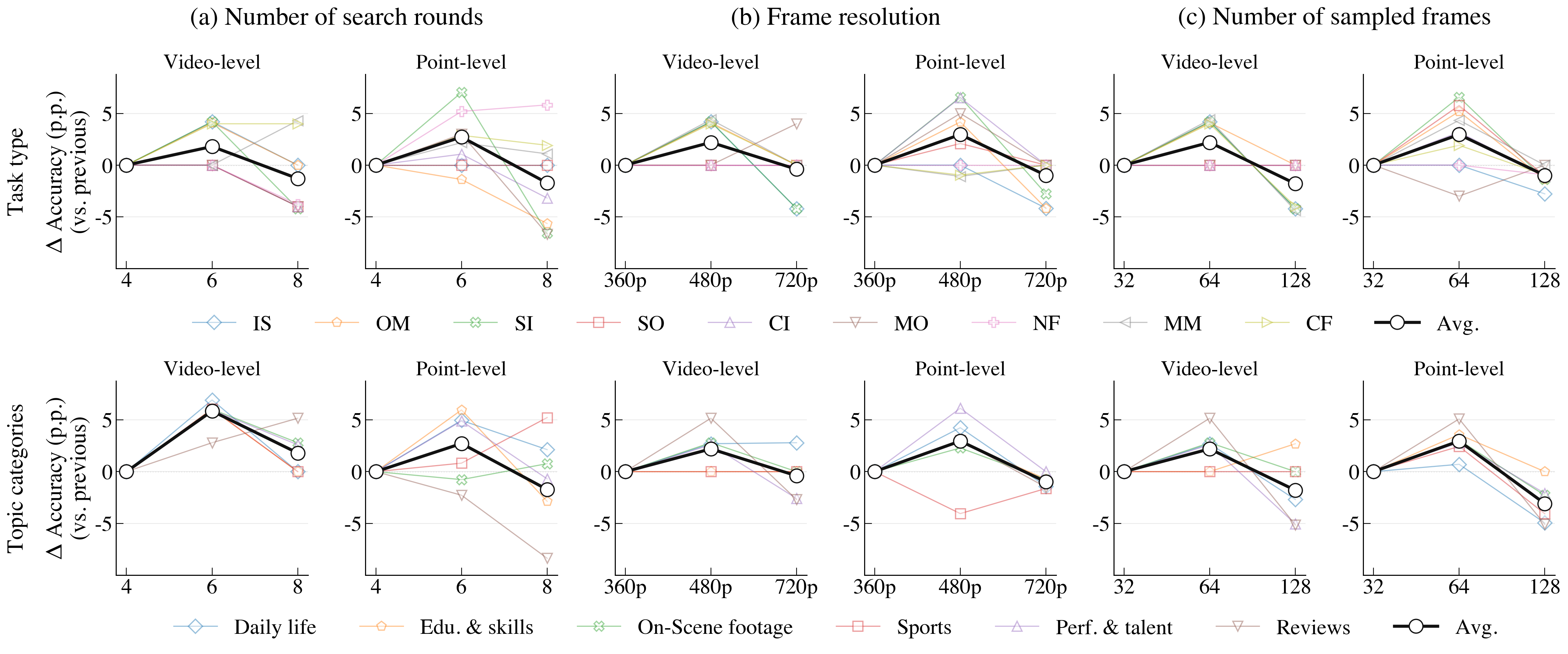}
  \caption{Ablation on GPT-5.5 over search rounds, frame resolution, and sampled frames.
  Video-level and point-level accuracy are broken down by task type (top) and topic category (bottom).
  Curves plot stepwise accuracy change in percentage points: $\Delta_1=0$ at the leftmost setting and $\Delta_j=\mathrm{Acc}_j-\mathrm{Acc}_{j-1}$ for $j\ge 2$.}
  \label{fig:ablation_gpt}
  \vspace{-10pt}
\end{figure*}

We ablate three key hyperparameters using GPT-5.5 and Qwen-3.5-Plus as representative models. Figure~\ref{fig:ablation_gpt} shows the fine-grained breakdown for GPT-5.5. The corresponding results for Qwen-3.5-Plus are provided in Appendix~\ref{sec:appendix_ablation_qwen}.

\textbf{Number of search rounds.}
Performance peaks at 6 rounds for both models (GPT-5.5: 61.43\%; Qwen-3.5-Plus: 53.4\%). Fewer rounds miss relevant evidence, particularly for Single-Source and Multi-Source Editing; more rounds introduce irrelevant results that derail the analysis (see Appendix~\ref{app:case_ablation_rounds}).

\textbf{Frame resolution.}
Both models perform best at 480p. Lower resolution limits fine-grained cross-video comparison; higher resolution causes fixation on low-level details at the expense of semantic understanding (see Appendix~\ref{app:case_ablation_resolution}).

\textbf{Number of sampled frames.}
Both models peak at 64 frames compared to 32 and 128 frames. Increasing from 32 to 64 provides meaningful improvement by capturing more temporal information, but further increases can expose fine-grained differences that mislead the model away from narrative-level reasoning (see case study in Appendix~\ref{app:case_ablation_frames}).

\subsection{Error Analysis}

We conduct a systematic error analysis across model outputs to identify recurring failure patterns. We organize findings into two categories: model-level reasoning failures and pipeline-level structural failures.

\subsubsection{Model-Level Failures}

\textbf{Attention captured by salient but irrelevant anchors.}
Models are frequently drawn to visually or textually prominent elements---video titles, brand names, on-screen slogans---at the expense of understanding the actual manipulation. Rather than reasoning about how footage has been restructured, models latch onto conspicuous keywords and build explanations around them (see Appendix~\ref{app:case_keyword_hijack}).

\textbf{Inability to recognize AI-generated content.}
When videos contain synthetic insertions, models fail to recognize them as generated content. They detect that extra material is present but consistently interpret it as footage sourced from other real videos rather than content that was never filmed (see Appendix~\ref{app:case_ai_generated_content_blindness}).

\textbf{Misinformation type misattribution.}
Even when models correctly detect manipulation, they often misclassify its granularity. Specific mechanisms such as identity swap or contextual fabrication are collapsed into coarser categories like editorial montage or generic narrative fabrication. Models sense that something changed but cannot pinpoint which layer---identity, object, temporal structure, or contextual attribution---was altered (see Appendix~\ref{app:case_method_misattribution}).

\textbf{Insensitivity to small-scale visual modifications.}
Object-level manipulations (replaced textures, altered clothing details, swapped handheld items) are frequently missed entirely. When the overall scene structure and activity flow remain unchanged, models conclude that no manipulation has occurred, even when objects have been visibly substituted (see Appendix~\ref{app:case_small_object_insensitivity}).

\textbf{Factual over emotional reasoning.}
Models excel at detecting factual fabrications (wrong dates, misattributed events) but struggle with manipulations that alter emotional tone or narrative arc. When a video's temporal reordering transforms a positive story into a negative one, models tend to focus on verifiable metadata rather than recognizing the shifted narrative direction (see Appendix~\ref{app:case_emotional_reframing_blindness}).

\textbf{Detail fixation over narrative comprehension.}
Even when relevant evidence is retrieved, models often pursue low-level discrepancies (subtitle formatting, logo placement, parameter specifications) rather than identifying the structural manipulation that changes the video's meaning (see Appendix~\ref{app:case_detail_chasing_wrong_goal}).

\subsubsection{Pipeline-Level Failures}

\textbf{Source decomposition errors.}
When the initial chain-of-thought incorrectly estimates the number of sources or misidentifies which source is most critical, all subsequent retrieval rounds are constrained to the wrong search space (see Appendix~\ref{app:pipeline_stage_a_source_decomposition}).

\textbf{Query reflection trapped in error clusters.}
The iterative refinement mechanism sometimes fails to escape incorrect semantic neighborhoods. Successive queries drift within the same wrong cluster rather than exploring fundamentally different search directions (see Appendix~\ref{app:pipeline_stage_b_query_reflection}).

\textbf{Forgery extraction misaligned with ground truth.}
The most prevalent pipeline failure occurs at the fine-grained comparison stage: the correct source video is retrieved, but the contrastive analysis identifies the wrong manipulation axis. Models describe a coherent but incorrect forgery narrative, mistaking structural reordering for subtitle errors or causal manipulation for branding alterations (see Appendix~\ref{app:pipeline_stage_d_fine_extraction}).

\textbf{Premature stopping.}
The sufficiency judgment conflates ``source found'' with ``manipulation understood.'' Once a matching source is retrieved, the pipeline often terminates without verifying whether the identified forgery points actually cover the true manipulation (see Appendix~\ref{app:pipeline_stage_e_stop_decision}).



\section{Conclusion}

We presented EVID-Bench, a benchmark for search-grounded video misinformation detection with 222 videos across 9 manipulation types. Experiments on nine frontier models show that the best system reaches only 61.43\% point-level and 43.24\% video-level accuracy, with AI-generated manipulations remaining especially difficult. These results highlight the need for systems that verify videos through external evidence retrieval and cross-video reasoning, rather than relying only on visual artifact detection.

\section*{Limitations}

The benchmark contains 222 videos, which limits statistical power for fine-grained comparisons across individual task types. Our evaluation depends on web search APIs whose results may change over time, affecting reproducibility. The quality assurance filters based on current frontier models; future models may find some samples detectable without retrieval. Finally, the LLM-based evaluation protocol, while more robust than single-model judgment, may introduce systematic biases in semantic matching.

\section*{Ethical Considerations}

EVID-Bench is developed solely for defensive research on video misinformation. All source videos are collected from publicly accessible online sources, primarily YouTube; no private, confidential, or non-consensually obtained footage or additional personal information is used. Some samples involve celebrities or public figures as representative facial-manipulation cases, but they are not intended to target, defame, or attribute misleading claims or actions to any individual.

We acknowledge that manipulated faces and detailed detection vulnerabilities may be misused for impersonation or more deceptive content. To mitigate these risks, the benchmark will be released for non-commercial academic research under the stated license and usage policy, which prohibit malicious use. Since EVID-Bench is English-centric and contains 222 curated cases, its pipeline should not be treated as a deployable moderation system; we recommend human-in-the-loop review in practical applications.

\section*{Acknowledgments}
Thank these students for their contributions during the video editing phase: Siyi Cao, Yuqing Wu, Peifan Wang, Tianyang Dang, Minghui Guo, Hanyu Wang, Hai Huang, Xingyu Yang, Xinya Sun.

This work was jointly supported by National Natural Science Foundation of China (62322607, 62236010 and 62276261), Beijing Natural Science Foundation (L252033).

\setcitestyle{numbers,square}
\bibliography{citation}

@article{yan2025sanity,
  title={A Sanity Check for {AI}-Generated Image Detection},
  author={Yan, Shilin and Li, Ouxiang and Cai, Jiayin and Hao, Yanbin and Jiang, Xiaolong and Hu, Yao and Xie, Weidi},
  journal={arXiv preprint arXiv:2406.19435},
  year={2025}
}

@article{ojha2024universal,
  title={Towards Universal Fake Image Detectors That Generalize across Generative Models},
  author={Ojha, Utkarsh and Li, Yuheng and Lee, Yong Jae},
  journal={arXiv preprint arXiv:2302.10174},
  year={2024}
}

@article{singh2024copy,
  title={Copy-move video forgery detection techniques: A systematic survey with comparisons, challenges and future directions},
  author={Singh, Gurvinder and Singh, Kulbir},
  journal={Wireless Personal Communications},
  volume={134},
  number={3},
  pages={1863--1913},
  year={2024},
  publisher={Springer}
}

@article{fatima2026enhanced,
  title={Enhanced inter-frame video forgery detection using convolutional network and stacking ensemble},
  author={Fatima, Baheesa and Bakhshi, Asim Dilawar and Ghafoor, Abdul},
  journal={Multimedia Tools and Applications},
  volume={85},
  number={5},
  pages={497},
  year={2026},
  publisher={Springer}
}

@misc{hernandezortega2020deepfakesonphysdeepfakesdetectionbased,
      title={DeepFakesON-Phys: DeepFakes Detection based on Heart Rate Estimation}, 
      author={Javier Hernandez-Ortega and Ruben Tolosana and Julian Fierrez and Aythami Morales},
      year={2020},
      eprint={2010.00400},
      archivePrefix={arXiv},
      primaryClass={cs.CV},
      url={https://arxiv.org/abs/2010.00400}, 
}

@misc{qian2020thinkingfrequencyfaceforgery,
      title={Thinking in Frequency: Face Forgery Detection by Mining Frequency-aware Clues}, 
      author={Yuyang Qian and Guojun Yin and Lu Sheng and Zixuan Chen and Jing Shao},
      year={2020},
      eprint={2007.09355},
      archivePrefix={arXiv},
      primaryClass={cs.CV},
      url={https://arxiv.org/abs/2007.09355}, 
}

@misc{abdali2024multimodalmisinformationdetectionapproaches,
      title={Multi-modal Misinformation Detection: Approaches, Challenges and Opportunities}, 
      author={Sara Abdali and Sina shaham and Bhaskar Krishnamachari},
      year={2024},
      eprint={2203.13883},
      archivePrefix={arXiv},
      primaryClass={cs.LG},
      url={https://arxiv.org/abs/2203.13883}, 
}

@misc{tian2026exposingcrossmodalconsistencyfake,
      title={Exposing Cross-Modal Consistency for Fake News Detection in Short-Form Videos}, 
      author={Chong Tian and Yu Wang and Chenxu Yang and Junyi Guan and Zheng Lin and Yuhan Liu and Xiuying Chen and Qirong Ho},
      year={2026},
      eprint={2603.14992},
      archivePrefix={arXiv},
      primaryClass={cs.AI},
      url={https://arxiv.org/abs/2603.14992}, 
}

@misc{shopnil2026meritmodularframeworkmultimodal,
      title={MERIT: Modular Framework for Multimodal Misinformation Detection with Web-Grounded Reasoning}, 
      author={Mir Nafis Sharear Shopnil and Sharad Duwal and Abhishek Tyagi and Adiba Mahbub Proma},
      year={2026},
      eprint={2510.17590},
      archivePrefix={arXiv},
      primaryClass={cs.AI},
      url={https://arxiv.org/abs/2510.17590}, 
}

@misc{aneja2021cosmoscatchingoutofcontextmisinformation,
      title={COSMOS: Catching Out-of-Context Misinformation with Self-Supervised Learning}, 
      author={Shivangi Aneja and Chris Bregler and Matthias Nießner},
      year={2021},
      eprint={2101.06278},
      archivePrefix={arXiv},
      primaryClass={cs.CV},
      url={https://arxiv.org/abs/2101.06278}, 
}

@inproceedings{gunasekara2025influence,
  title={The Influence of Content Modality on Perceptions of Online Misinformation},
  author={Gunasekara, Suwani and Pareek, Saumya and Kelly, Ryan M and Goncalves, Jorge},
  booktitle={Proceedings of the 2025 CHI Conference on Human Factors in Computing Systems},
  pages={1--10},
  year={2025}
}

@article{sundar2021seeing,
  title={Seeing is believing: Is video modality more powerful in spreading fake news via online messaging apps?},
  author={Sundar, S Shyam and Molina, Maria D and Cho, Eugene},
  journal={Journal of Computer-Mediated Communication},
  volume={26},
  number={6},
  pages={301--319},
  year={2021},
  publisher={Oxford University Press}
}

@article{diwan2024systematic,
  title={Systematic analysis of video tampering and detection techniques},
  author={Diwan, Anjali and Dixit, Saurav and Subbiah, Ram and Mahadeva, Rajesh},
  journal={Cogent Engineering},
  volume={11},
  number={1},
  pages={2424466},
  year={2024},
  publisher={Taylor \& Francis}
}

@inproceedings{wang2025fmnv,
  title={Fmnv: A dataset of media-published news videos for fake news detection},
  author={Wang, Yihao and Qian, Zhong and Li, Peifeng},
  booktitle={International Conference on Intelligent Computing},
  pages={321--332},
  year={2025},
  organization={Springer}
}

@article{yu2026shotfinder,
  title={ShotFinder: Imagination-Driven Open-Domain Video Shot Retrieval via Web Search},
  author={Yu, Tao and Jin, Haopeng and Wang, Hao and Chai, Shenghua and Yang, Yujia and Gong, Junhao and Guo, Jiaming and Zhang, Minghui and Chen, Xinlong and Zhang, Zhenghao and others},
  journal={arXiv preprint arXiv:2601.23232},
  year={2026}
}

@article{yu2026beyond,
  title={Beyond Closed-Pool Video Retrieval: A Benchmark and Agent Framework for Real-World Video Search and Moment Localization},
  author={Yu, Tao and Yang, Yujia and Jin, Haopeng and Gong, Junhao and Chen, Xinlong and Zhou, Yuxuan and Zhang, Shanbin and Yang, Jiabing and Wang, Xinming and Yi, Hongzhu and others},
  journal={arXiv preprint arXiv:2602.10159},
  year={2026}
}

@inproceedings{chou2026improving,
  title={Improving Deepfake Detection with Reinforcement Learning-Based Adaptive Data Augmentation},
  author={Chou, Yuxuan and Yu, Tao and Huang, Wen and Dai, Tao and Xia, Shu-Tao and others},
  booktitle={Proceedings of the AAAI Conference on Artificial Intelligence},
  volume={40},
  pages={3381--3389},
  year={2026}
}

@misc{seedance2026seedance20advancingvideo,
      title={Seedance 2.0: Advancing Video Generation for World Complexity}, 
      author={Team Seedance and De Chen and Liyang Chen and Xin Chen and Ying Chen and Zhuo Chen and Zhuowei Chen and Feng Cheng and Tianheng Cheng and Yufeng Cheng and Mojie Chi and Xuyan Chi and Jian Cong and Qinpeng Cui and Fei Ding and Qide Dong and Yujiao Du and Haojie Duanmu and Junliang Fan and Jiarui Fang and Jing Fang and Zetao Fang and Chengjian Feng and Yu Gao and Diandian Gu and Dong Guo and Hanzhong Guo and Qiushan Guo and Boyang Hao and Hongxiang Hao and Haoxun He and Jiaao He and Qian He and Tuyen Hoang and Heng Hu and Ruoqing Hu and Yuxiang Hu and Jiancheng Huang and Weilin Huang and Zhaoyang Huang and Zhongyi Huang and Jishuo Jin and Ming Jing and Ashley Kim and Shanshan Lao and Yichong Leng and Bingchuan Li and Gen Li and Haifeng Li and Huixia Li and Jiashi Li and Ming Li and Xiaojie Li and Xingxing Li and Yameng Li and Yiying Li and Yu Li and Yueyan Li and Chao Liang and Han Liang and Jianzhong Liang and Ying Liang and Wang Liao and J. H. Lien and Shanchuan Lin and Xi Lin and Feng Ling and Yue Ling and Fangfang Liu and Jiawei Liu and Jihao Liu and Jingtuo Liu and Shu Liu and Sichao Liu and Wei Liu and Xue Liu and Zuxi Liu and Ruijie Lu and Lecheng Lyu and Jingting Ma and Tianxiang Ma and Xiaonan Nie and Jingzhe Ning and Junjie Pan and Xitong Pan and Ronggui Peng and Xueqiong Qu and Yuxi Ren and Yuchen Shen and Guang Shi and Lei Shi and Yinglong Song and Fan Sun and Li Sun and Renfei Sun and Wenjing Tang and Boyang Tao and Zirui Tao and Dongliang Wang and Feng Wang and Hulin Wang and Ke Wang and Qingyi Wang and Rui Wang and Shuai Wang and Shulei Wang and Weichen Wang and Xuanda Wang and Yanhui Wang and Yue Wang and Yuping Wang and Yuxuan Wang and Zijie Wang and Ziyu Wang and Guoqiang Wei and Meng Wei and Di Wu and Guohong Wu and Hanjie Wu and Huachao Wu and Jian Wu and Jie Wu and Ruolan Wu and Shaojin Wu and Xiaohu Wu and Xinglong Wu and Yonghui Wu and Ruiqi Xia and Xin Xia and Xuefeng Xiao and Shuang Xu and Bangbang Yang and Jiaqi Yang and Runkai Yang and Tao Yang and Yihang Yang and Zhixian Yang and Ziyan Yang and Fulong Ye and Bingqian Yi and Xing Yin and Yongbin You and Linxiao Yuan and Weihong Zeng and Xuejiao Zeng and Yan Zeng and Siyu Zhai and Zhonghua Zhai and Bowen Zhang and Chenlin Zhang and Heng Zhang and Jun Zhang and Manlin Zhang and Peiyuan Zhang and Shuo Zhang and Xiaohe Zhang and Xiaoying Zhang and Xinyan Zhang and Xinyi Zhang and Yichi Zhang and Zixiang Zhang and Haiyu Zhao and Huating Zhao and Liming Zhao and Yian Zhao and Guangcong Zheng and Jianbin Zheng and Xiaozheng Zheng and Zerong Zheng and Kuan Zhu and Feilong Zuo},
      year={2026},
      eprint={2604.14148},
      archivePrefix={arXiv},
      primaryClass={cs.CV},
      url={https://arxiv.org/abs/2604.14148}, 
}

@misc{gemini3pro2025,
  title={Gemini 3},
  author={{Google DeepMind}},
  year={2025},
  howpublished={\url{https://blog.google/products/gemini/gemini-3/}}
}

@misc{singh2025openai,
  title={Openai gpt-5.5 system card},
  author       = {{OpenAI}},
  year={2026},
  howpublished={\url{https://openai.com/index/gpt-5-5-system-card/}}
}

@misc{singh2025openai5.4,
  title={Openai gpt-5.4 system card},
  author       = {{OpenAI}},
  year={2026},
  howpublished={\url{https://openai.com/index/gpt-5-4-thinking-system-card/}}
}

@misc{claudeops,
  title={Claude Opus 4.6},
  author       = {{Anthropic}},
  year={2026},
  howpublished={\url{https://www.anthropic.com/news/claude-opus-4-6}}
}

@misc{claudeson,
  title={Claude Sonnet 4.6},
  author       = {{Anthropic}},
  year={2026},
  howpublished={\url{https://www.anthropic.com/news/claude-sonnet-4-6}}
}

@misc{bai2025qwen3vltechnicalreport,
      title={Qwen3-VL Technical Report}, 
      author={Shuai Bai and Yuxuan Cai and Ruizhe Chen and Keqin Chen and Xionghui Chen and Zesen Cheng and Lianghao Deng and Wei Ding and Chang Gao and Chunjiang Ge and Wenbin Ge and Zhifang Guo and Qidong Huang and Jie Huang and Fei Huang and Binyuan Hui and Shutong Jiang and Zhaohai Li and Mingsheng Li and Mei Li and Kaixin Li and Zicheng Lin and Junyang Lin and Xuejing Liu and Jiawei Liu and Chenglong Liu and Yang Liu and Dayiheng Liu and Shixuan Liu and Dunjie Lu and Ruilin Luo and Chenxu Lv and Rui Men and Lingchen Meng and Xuancheng Ren and Xingzhang Ren and Sibo Song and Yuchong Sun and Jun Tang and Jianhong Tu and Jianqiang Wan and Peng Wang and Pengfei Wang and Qiuyue Wang and Yuxuan Wang and Tianbao Xie and Yiheng Xu and Haiyang Xu and Jin Xu and Zhibo Yang and Mingkun Yang and Jianxin Yang and An Yang and Bowen Yu and Fei Zhang and Hang Zhang and Xi Zhang and Bo Zheng and Humen Zhong and Jingren Zhou and Fan Zhou and Jing Zhou and Yuanzhi Zhu and Ke Zhu},
      year={2025},
      eprint={2511.21631},
      archivePrefix={arXiv},
      primaryClass={cs.CV},
      url={https://arxiv.org/abs/2511.21631}, 
}

@misc{qwen3.5,
    title  = {{Qwen3.5}: Towards Native Multimodal Agents},
    author = {{Qwen Team}},
    year   = {2026},
    month  = {February},
    url    = {https://qwen.ai/blog?id=qwen3.5}
}

@inproceedings{rossler2019faceforensics++,
  title={Faceforensics++: Learning to detect manipulated facial images},
  author={Rossler, Andreas and Cozzolino, Davide and Verdoliva, Luisa and Riess, Christian and Thies, Justus and Nie{\ss}ner, Matthias},
  booktitle={Proceedings of the IEEE/CVF international conference on computer vision},
  pages={1--11},
  year={2019}
}

@article{ho2022video,
  title={Video diffusion models},
  author={Ho, Jonathan and Salimans, Tim and Gritsenko, Alexey and Chan, William and Norouzi, Mohammad and Fleet, David J},
  journal={Advances in neural information processing systems},
  volume={35},
  pages={8633--8646},
  year={2022}
}

@article{fischinger2025df,
  title={DF-Net: The digital forensics network for image forgery detection},
  author={Fischinger, David and Boyer, Martin},
  journal={arXiv preprint arXiv:2503.22398},
  year={2025}
}

@book{wardle2017information,
  title={Information disorder: Toward an interdisciplinary framework for research and policymaking},
  author={Wardle, Claire and Derakhshan, Hossein},
  volume={27},
  year={2017},
  publisher={Council of Europe Strasbourg}
}

@inproceedings{gangal2022nareor,
  title={Nareor: The narrative reordering problem},
  author={Gangal, Varun and Feng, Steven Y and Alikhani, Malihe and Mitamura, Teruko and Hovy, Eduard},
  booktitle={Proceedings of the AAAI Conference on Artificial Intelligence},
  volume={36},
  pages={10645--10653},
  year={2022}
}

@article{yin2024text,
  title={Text-video multi-grained integration for video moment montage},
  author={Yin, Zhihui and Ma, Ye and Cao, Xipeng and Wang, Bo and Chen, Quan and Jiang, Peng},
  journal={arXiv preprint arXiv:2412.09276},
  year={2024}
}

@article{pei2024multi,
  title={Multi-View Inconsistency Analysis for Video Object-Level Splicing Localization},
  author={Pei, Pengfei and Liang, Guoqing and Luan, Tao},
  journal={International Journal of Emerging Technologies and Advanced Applications},
  volume={1},
  number={3},
  pages={1--5},
  year={2024}
}

@inproceedings{bu2023combating,
  title={Combating online misinformation videos: Characterization, detection, and future directions},
  author={Bu, Yuyan and Sheng, Qiang and Cao, Juan and Qi, Peng and Wang, Danding and Li, Jintao},
  booktitle={Proceedings of the 31st ACM International Conference on Multimedia},
  pages={8770--8780},
  year={2023}
}

@inproceedings{luo2021newsclippings,
  title={Newsclippings: Automatic generation of out-of-context multimodal media},
  author={Luo, Grace and Darrell, Trevor and Rohrbach, Anna},
  booktitle={Proceedings of the 2021 Conference on Empirical Methods in Natural Language Processing},
  pages={6801--6817},
  year={2021}
}

@article{papadopoulou2019corpus,
  title={A corpus of debunked and verified user-generated videos},
  author={Papadopoulou, Olga and Zampoglou, Markos and Papadopoulos, Symeon and Kompatsiaris, Ioannis},
  journal={Online information review},
  volume={43},
  number={1},
  pages={72--88},
  year={2019},
  publisher={Emerald Publishing Limited}
}

@article{wang2024official,
  title={Official-NV: An LLM-Generated News Video Dataset for Multimodal Fake News Detection},
  author={Wang, Yihao and Chen, Lizhi and Qian, Zhong and Li, Peifeng},
  journal={arXiv preprint arXiv:2407.19493},
  year={2024}
}

@article{huang2026probing,
  title={Probing Multimodal Large Language Models on Cognitive Biases in Chinese Short-Video Misinformation},
  author={Huang, Jen-tse and Chen, Chang and Lai, Shiyang and Wang, Wenxuan and Kaufman, Michelle R and Dredze, Mark},
  journal={arXiv preprint arXiv:2601.06600},
  year={2026}
}

@inproceedings{bevendorff2024product,
  title={Product spam on youtube: A case study},
  author={Bevendorff, Janek and Wiegmann, Matti and Potthast, Martin and Stein, Benno},
  booktitle={Proceedings of the 2024 conference on human information interaction and retrieval},
  pages={358--363},
  year={2024}
}

@misc{yu2026nonadversarialnoisehurtslearning,
      title={When Non-Adversarial Noise Hurts Learning: Benchmarking Skill Self-Evolution in Agents}, 
      author={Tao Yu and Zhongtian Luo and Yuchen Long and Yiheng Lu and Yuxuan Zhou and Yanpei Gong and Siyi Cao and Yuqing Wu and Haopeng Jin and Zhenghao Zhang and Yujia Yang and Jingmin Zhu and Yunyun Hou and Xinming Wang and Hongzhu Yi and Jiachen Liu and Sirui Wang and Nan Duan and Yan Huang and Liang Wang and Tianjin Huang and Yanyan Zou},
      year={2026},
      eprint={2610.19910},
      archivePrefix={arXiv},
      primaryClass={cs.LG},
      url={https://arxiv.org/abs/2610.19910}, 
}

\newpage

\newpage
\EnableTOC
\clearpage
\appendix

\section*{Appendix}
\begingroup
\setlength{\parskip}{0pt}
\small
\setlength{\baselineskip}{10pt}
\setcounter{tocdepth}{2}  
\tableofcontents
\endgroup
\clearpage 

\begin{table}[H]
\centering
\caption{Point-level accuracy of evaluated models across topic categories. All metrics are reported in percentages (\%).}
\label{tab:3}
\setlength{\tabcolsep}{0pt}
\renewcommand{\arraystretch}{1.25} 
\resizebox{0.95\textwidth}{!}{
\begin{tabular}{l @{\hspace{16pt}} *{7}{>{\centering\arraybackslash}m{0.125\textwidth}}}
\toprule
\textbf{Model} & \textbf{\makecell{Daily\\life}} & \textbf{\makecell{Education\\\& skills}} & \textbf{\makecell{On-Scene\\footage}} & \textbf{Sports} & \textbf{\makecell{Performance\\\& talent}} & \textbf{Reviews} & \textbf{Avg.} \\
\midrule
\multicolumn{8}{l}{\textbf{GPT Series}} \\
\addlinespace[3pt]
\hdashline
\addlinespace[3pt]
GPT-5.5 & \colorcatI{100}44.68 ($\pm$0.4) & \colorcatI{100}69.78 ($\pm$0.0) & \colorcatI{100}77.86 ($\pm$0.8) & \colorcatI{100}66.67 ($\pm$1.7) & \colorcatI{100}73.43 ($\pm$1.2) & \colorcatI{68}36.36 ($\pm$1.9) & \coloravg{100}61.43 ($\pm$0.9) \\
GPT-5.4 & \colorcatI{60}35.46 ($\pm$1.1) & \colorcatI{78}61.15 ($\pm$1.1) & \colorcatI{75}68.70 ($\pm$0.9) & \colorcatI{74}56.10 ($\pm$1.2) & \colorcatI{67}65.03 ($\pm$2.1) & \colorcatI{68}36.36 ($\pm$0.0) & \coloravg{73}53.77 ($\pm$0.9) \\
GPT-5.4-Mini & \colorcatI{48}32.62 ($\pm$1.7) & \colorcatI{53}51.80 ($\pm$0.7) & \colorcatI{49}58.78 ($\pm$0.9) & \colorcatI{56}48.78 ($\pm$0.5) & \colorcatI{31}55.94 ($\pm$1.3) & \colorcatI{60}34.09 ($\pm$2.0) & \coloravg{49}46.97 ($\pm$1.0) \\
\midrule
\multicolumn{8}{l}{\textbf{Claude Series}} \\
\addlinespace[3pt]
\hdashline
\addlinespace[3pt]
Claude-Opus-4.6 & \colorcatI{85}41.13 ($\pm$1.7) & \colorcatI{91}66.19 ($\pm$0.4) & \colorcatI{90}74.05 ($\pm$1.3) & \colorcatI{88}61.79 ($\pm$0.9) & \colorcatI{92}71.33 ($\pm$1.1) & \colorcatI{65}35.61 ($\pm$0.9) & \coloravg{89}58.34 ($\pm$1.0) \\
Claude-Sonnet-4.6 & \colorcatI{72}38.30 ($\pm$1.3) & \colorcatI{80}61.87 ($\pm$0.4) & \colorcatI{77}69.47 ($\pm$1.2) & \colorcatI{78}57.72 ($\pm$0.9) & \colorcatI{81}68.53 ($\pm$1.7) & \colorcatI{63}34.85 ($\pm$0.9) & \coloravg{78}55.13 ($\pm$1.0) \\
\midrule
\multicolumn{8}{l}{\textbf{Gemini Series}} \\
\addlinespace[3pt]
\hdashline
\addlinespace[3pt]
Gemini-3.1-Pro & \colorcatI{81}40.40 ($\pm$0.7) & \colorcatI{83}63.30 ($\pm$1.4) & \colorcatI{73}67.90 ($\pm$1.4) & \colorcatI{76}56.90 ($\pm$0.9) & \colorcatI{78}67.80 ($\pm$1.3) & \colorcatI{76}38.60 ($\pm$0.9) & \coloravg{81}55.90 ($\pm$1.0) \\
Gemini-3-Flash & \colorcatI{72}38.30 ($\pm$1.6) & \colorcatI{85}64.00 ($\pm$0.4) & \colorcatI{51}59.50 ($\pm$0.9) & \colorcatI{78}57.70 ($\pm$1.2) & \colorcatI{45}59.40 ($\pm$1.5) & \colorcatI{92}43.20 ($\pm$0.8) & \coloravg{73}53.60 ($\pm$1.0) \\
\midrule
\multicolumn{8}{l}{\textbf{Qwen Series}} \\
\addlinespace[3pt]
\hdashline
\addlinespace[3pt]
Qwen-3.5-Plus & \colorcatI{81}40.40 ($\pm$0.5) & \colorcatI{96}68.30 ($\pm$0.8) & \colorcatI{40}55.70 ($\pm$1.3) & \colorcatI{60}50.40 ($\pm$0.6) & \colorcatI{75}67.10 ($\pm$1.4) & \colorcatI{71}37.10 ($\pm$1.6) & \coloravg{72}53.40 ($\pm$1.0) \\
Qwen3-VL-235B-A22B-Instruct & \colorcatI{69}37.60 ($\pm$1.6) & \colorcatI{50}50.40 ($\pm$1.4) & \colorcatI{69}66.40 ($\pm$1.6) & \colorcatI{66}52.80 ($\pm$1.3) & \colorcatI{58}62.90 ($\pm$0.7) & \colorcatI{20}22.70 ($\pm$1.7) & \coloravg{56}48.80 ($\pm$1.0) \\
\bottomrule
\end{tabular}
}
\end{table}

\begin{table}[H]
\centering
\caption{Video-level accuracy of evaluated models across topic categories. All metrics are reported in percentages (\%).}
\label{tab:4}
\setlength{\tabcolsep}{0pt}
\renewcommand{\arraystretch}{1.25} 
\resizebox{0.95\textwidth}{!}{
\begin{tabular}{l @{\hspace{16pt}} *{7}{>{\centering\arraybackslash}m{0.125\textwidth}}}
\toprule
\textbf{Model} & \textbf{\makecell{Daily\\life}} & \textbf{\makecell{Education\\\& skills}} & \textbf{\makecell{On-Scene\\footage}} & \textbf{Sports} & \textbf{\makecell{Performance\\\& talent}} & \textbf{Reviews} & \textbf{Avg.} \\
\midrule
\multicolumn{8}{l}{\textbf{GPT Series}} \\

\hdashline

GPT-5.5 & \colorcatI{100}48.65 ($\pm$1.6) & \colorcatI{100}59.46 ($\pm$0.0) & \colorcatI{100}51.35 ($\pm$0.0) & \colorcatI{100}54.05 ($\pm$1.6) & \colorcatI{62}36.84 ($\pm$1.5) & \colorcatI{20}8.33 ($\pm$0.0) & \coloravg{100}43.24 ($\pm$0.7) \\
GPT-5.4 & \colorcatI{73}35.14 ($\pm$0.0) & \colorcatI{75}48.65 ($\pm$1.6) & \colorcatI{71}40.54 ($\pm$1.6) & \colorcatI{73}43.24 ($\pm$0.0) & \colorcatI{46}31.58 ($\pm$1.5) & \colorcatI{46}13.89 ($\pm$0.0) & \coloravg{75}35.59 ($\pm$0.7) \\
GPT-5.4-Mini & \colorcatI{63}29.73 ($\pm$0.0) & \colorcatI{45}35.14 ($\pm$1.6) & \colorcatI{42}29.73 ($\pm$0.0) & \colorcatI{46}32.43 ($\pm$0.0) & \colorcatI{20}21.05 ($\pm$1.5) & \colorcatI{20}8.33 ($\pm$0.0) & \coloravg{44}26.13 ($\pm$0.3) \\
\midrule
\multicolumn{8}{l}{\textbf{Claude Series}} \\

\hdashline

Claude-Opus-4.6 & \colorcatI{89}43.24 ($\pm$0.0) & \colorcatI{88}54.05 ($\pm$0.0) & \colorcatI{93}48.65 ($\pm$0.0) & \colorcatI{80}45.95 ($\pm$0.0) & \colorcatI{46}31.58 ($\pm$0.0) & \colorcatI{20}8.33 ($\pm$1.6) & \coloravg{85}38.74 ($\pm$0.3) \\
Claude-Sonnet-4.6 & \colorcatI{73}35.14 ($\pm$0.0) & \colorcatI{75}48.65 ($\pm$0.0) & \colorcatI{78}43.24 ($\pm$0.0) & \colorcatI{67}40.54 ($\pm$1.6) & \colorcatI{39}28.95 ($\pm$1.3) & \colorcatI{20}8.33 ($\pm$0.0) & \coloravg{70}34.23 ($\pm$0.7) \\
\midrule
\multicolumn{8}{l}{\textbf{Gemini Series}} \\

\hdashline

Gemini-3.1-Pro & \colorcatI{49}29.70 ($\pm$0.0) & \colorcatI{75}48.60 ($\pm$1.9) & \colorcatI{78}43.20 ($\pm$1.6) & \colorcatI{57}35.10 ($\pm$1.8) & \colorcatI{100}47.40 ($\pm$1.4) & \colorcatI{60}16.70 ($\pm$1.9) & \coloravg{79}36.90 ($\pm$1.4) \\
Gemini-3-Flash & \colorcatI{42}27.00 ($\pm$1.8) & \colorcatI{69}45.90 ($\pm$0.0) & \colorcatI{71}40.50 ($\pm$0.0) & \colorcatI{75}43.20 ($\pm$1.6) & \colorcatI{46}31.60 ($\pm$1.6) & \colorcatI{73}19.40 ($\pm$0.0) & \coloravg{72}34.70 ($\pm$0.3) \\
\midrule
\multicolumn{8}{l}{\textbf{Qwen Series}} \\

\hdashline
Qwen-3.5-Plus & \colorcatI{33}24.30 ($\pm$0.0) & \colorcatI{63}43.20 ($\pm$1.6) & \colorcatI{42}29.70 ($\pm$1.7) & \colorcatI{60}37.80 ($\pm$0.0) & \colorcatI{100}44.70 ($\pm$1.8) & \colorcatI{100}25.00 ($\pm$0.0) & \coloravg{70}34.20 ($\pm$1.0) \\
Qwen3-VL-235B-A22B-Instruct & \colorcatI{41}27.00 ($\pm$1.9) & \colorcatI{45}35.10 ($\pm$0.0) & \colorcatI{85}45.90 ($\pm$1.6) & \colorcatI{53}35.10 ($\pm$0.0) & \colorcatI{100}44.70 ($\pm$0.0) & \colorcatI{33}11.10 ($\pm$1.9) & \coloravg{67}33.30 ($\pm$0.6) \\
\bottomrule
\end{tabular}
}
\end{table}

\clearpage 
\setlength{\parskip}{0.5\baselineskip}
\pretocmd{\section}{\FloatBarrier\Needspace{8\baselineskip}}{}{}
\pretocmd{\subsection}{\FloatBarrier\Needspace{10\baselineskip}}{}{}

\section{Verification Process Details}
\label{sec:appendix_verification}

Our quality assurance pipeline has two stages, human review followed by model-based filtering. Below we describe each stage and list the exact prompts used.

\textbf{Human Review.} Three computer science researchers independently examine each sample. They check technical quality (resolution, frame rate, and whether compression artifacts are consistent with naturally recorded video) and narrative coherence (whether the intended manipulation is logically self-consistent and free of obvious editing seams). Samples that fail human review are revised or discarded before model-based filtering.

\textbf{Model-Based Filtering.} All samples then go through model-based filtering using Gemini-3.1-Pro and GPT-5.5 via their official web interfaces for direct visual reasoning, without providing any retrieval tools. The filtering strategy differs across the three forgery categories. For AI-generated forgeries (identity swap, object manipulation, and synthetic insertion), we require both models to fail to identify the manipulation through visual analysis. We use three complementary prompts for this purpose, each targeting a different aspect of potential artifacts. For single-source and multi-source editing (selective omission, causal inversion, manipulative montage, narrative fabrication, magnitude manipulation, and contextual fabrication), every frame is authentic so visual artifact detection alone is not enough. Instead we first ask Gemini to summarize the video content, then ask GPT to compare the summary against the editor's intended false narrative. Both models must interpret the video consistently with the manipulated narrative for the sample to pass.

\subsection{Forgery-Point Annotation and Cross-Check}
\label{sec:forgery-point-annotation}

\paragraph{Annotation procedure.}
The forgery points for each video are initially defined by the editor who constructs the manipulated video. Because the benchmark samples are deliberately constructed rather than collected as naturally occurring misinformation, the editor has direct knowledge of every modification applied to the source footage. Each forgery point corresponds to a distinct and intentional editing operation, such as replacing one identity with another during a specific segment, removing a causally important event, or splicing footage from different sources to create a false relationship.

The editing operations are converted into 3--5 natural-language forgery points according to three guidelines:

\begin{enumerate}[leftmargin=*, label=\arabic*.]
    \item \textbf{Specificity:} each point must identify the manipulated entity, segment, temporal relation, or contextual attribution and describe how it differs from the source footage;

    \item \textbf{Granularity consistency:} related edits that express the same falsification are grouped into one point, whereas independently verifiable manipulations are described as separate points;

    \item \textbf{External verifiability:} each point must, in principle, be verifiable by retrieving and comparing the original or related video evidence.
\end{enumerate}

\paragraph{Cross-check procedure.}
After the initial annotation, the manipulated video, its source video or videos, the editor's construction rationale, and the proposed forgery points are jointly reviewed by the annotation team. The review checks: (1) completeness, ensuring that no intentional manipulation is omitted; (2) granularity consistency, ensuring that points across videos are expressed at comparable levels of detail; and (3) verifiability, ensuring that each point can be supported through external video evidence. Points that are incomplete, redundant, overly broad, or not independently verifiable are revised, merged, or split during the review. The annotation is finalized only after the team agrees that the resulting set covers all intended manipulations at a consistent granularity.

This construction-aware annotation process grounds the labels in the known provenance of each manipulation rather than requiring third-party annotators to infer hidden edits from the forged video alone. The subsequent group review further reduces individual subjectivity and improves completeness and consistency across the benchmark.

\subsection{Prompt 1: Perceptual Quality Assessment}

\begin{tcolorbox}[
    title=\textbf{Prompt 1: Perceptual Quality Assessment},
    fonttitle=\bfseries,
    breakable,
]
You are a video content understanding and visual quality assessment assistant. Please watch the input video and complete the following task. Your task is to determine whether the video content is clear, natural, and coherent, and whether there are any visually perceptible abnormalities. Please focus on observing:

(1) \textbf{Subject naturalness} Whether the main subjects appear natural across different shots, angles, expressions, and lighting conditions.

(2) \textbf{Facial blending} Whether the face, facial features, skin, hair, neck, ears, glasses, hat, and other regions blend naturally with the body and scene.

(3) \textbf{Facial stability} Whether the face exhibits obvious discontinuities, flickering, drifting, edge artifacts, or identity changes when the subject speaks, turns, lowers their head, is occluded, or engages in rapid motion.

(4) \textbf{Overall authenticity} Whether the video overall resembles a normally recorded video or exhibits obvious post-production compositing, local replacement, frame stitching, or AI-generated artifacts.

(5) \textbf{Reporting principle} If no obvious abnormalities are observed, do not speculate about whether the video has been processed, simply state that the video appears natural overall.

Please output strictly in the following JSON format, with no additional text:

\begin{lstlisting}
{
  "content_summary": "<brief summary of video content>",
  "visible_abnormality": true/false,
  "abnormality_type": [
    "<no obvious abnormality / facial abnormality / body-face mismatch / edge blending abnormality / expression or lip-sync abnormality / flickering or drifting / identity instability / compositing artifact / other>"
  ],
  "abnormality_description": "<describe which subject, region, and manifestation if abnormality exists; otherwise 'no visually perceptible abnormality observed'>",
  "suspicion_of_manipulation": "<no obvious suspicion / slight suspicion / obvious suspicion>",
  "would_mention_as_problem_in_caption": true/false,
  "final_judgment": "<naturally convincing / mostly natural with minor flaws / obvious visual abnormality / suspected post-processing>",
  "reason": "<brief justification>"
}
\end{lstlisting}
\end{tcolorbox}

\subsection{Prompt 2: Temporal Coherence and Clip Insertion Detection}

\begin{tcolorbox}[
    title=\textbf{Prompt 2: Temporal Coherence and Clip Insertion Detection},
    fonttitle=\bfseries,
    breakable,
]
You are a video content understanding and visual quality assessment assistant. Please watch the input video and complete the following task. Your task is to determine whether the video content is clear, natural, and coherent, and whether there are any visually perceptible abnormalities or abrupt clip transitions. Please focus on observing:

(1) \textbf{Overall coherence} Whether the video content appears to depict the same continuous event or the same type of material, and whether different segments are consistent in scene, subjects, actions, filming style, and narrative logic.

(2) \textbf{Transition naturalness} Whether shot transitions are natural, and whether any abruptly inserted segments exhibit noticeable differences in visual style, resolution, color tone, lighting, camera motion, or compression artifacts.

(3) \textbf{Generative artifacts} Whether any segments exhibit obvious AI-generated appearance, animation-like quality, plastic texture, over-smoothing, melting visuals, or inauthentic details.

(4) \textbf{Temporal consistency} Whether subjects, objects, and backgrounds remain stable over time, and whether there is flickering, jumping, drifting, local deformation, or sudden appearance/disappearance.

(5) \textbf{Semantic plausibility} Whether the overall event depicted is clear, and whether any segments are semantically unrelated, logically disconnected, or stylistically inconsistent with the rest.

(6) \textbf{Reporting principle} If no obvious abnormalities exist, do not speculate. Only report abnormalities when segments are clearly inconsistent with the overall video or exhibit clear generative artifacts.

Please output strictly in the following JSON format, with no additional text:

\begin{lstlisting}
{
  "content_summary": "<brief summary of video content>",
  "visible_abnormality": true/false,
  "abnormality_type": [
    "<no obvious abnormality / abrupt clip transition / inconsistent visual style / resolution or quality shift / color or lighting shift / subject abnormality / limb or hand abnormality / facial abnormality / unnatural motion / object abnormality / background abnormality / text garbling / temporal discontinuity / semantic logic disruption / obvious AI-generated artifact / other>"
  ],
  "abnormality_description": "<describe which segment, subject, object, background, or time range if abnormality exists; otherwise 'no visually perceptible abnormality observed'>",
  "suspicion_of_ai_generation_or_insertion": "<no obvious suspicion / slight suspicion / obvious suspicion>",
  "would_mention_as_problem_in_caption": true/false,
  "final_judgment": "<naturally convincing / mostly natural with minor flaws / obvious visual abnormality / suspected AI generation or segment insertion>",
  "reason": "<brief justification>"
}
\end{lstlisting}
\end{tcolorbox}

\subsection{Prompt 3: Spatial and Object-Level Consistency}

\begin{tcolorbox}[
    title=\textbf{Prompt 3: Spatial and Object-Level Consistency},
    fonttitle=\bfseries,
    breakable,
]
You are a video content understanding and visual quality assessment assistant. Please watch the input video and complete the following task. Your task is to determine whether the video content is clear, natural, and coherent, and whether there are any visually perceptible abnormalities. Please focus on observing:

(1) \textbf{Spatial relationships} Whether subjects, objects, backgrounds, and spatial relationships appear natural.

(2) \textbf{Object plausibility} Whether any objects exhibit unreasonable position, size, perspective, occlusion, or motion trajectory.

(3) \textbf{Sudden changes} Whether there are sudden appearances, disappearances, flickering, drifting, abnormal shape changes, or poorly aligned edges.

(4) \textbf{Local artifacts} Whether there are local texture repetitions, blurry patches, repair traces, abrupt color changes, or inconsistent lighting and shadows.

(5) \textbf{Overall authenticity} Whether the video overall resembles a normally recorded video or exhibits obvious post-production compositing, local insertion, deletion, replacement, or AI-generated artifacts.

(6) \textbf{Reporting principle} If no obvious abnormalities are observed, do not speculate, simply state that the video appears natural overall.

Please output strictly in the following JSON format, with no additional text:

\begin{lstlisting}
{
  "content_summary": "<brief summary of video content>",
  "visible_abnormality": true/false,
  "abnormality_type": [
    "<no obvious abnormality / object position abnormality / scale or perspective abnormality / occlusion abnormality / motion trajectory abnormality / sudden appearance or disappearance / edge blending abnormality / local repair traces / lighting or shadow abnormality / compositing artifact / other>"
  ],
  "abnormality_description": "<describe which object, region, and manifestation if abnormality exists; otherwise 'no visually perceptible abnormality observed'>",
  "suspicion_of_manipulation": "<no obvious suspicion / slight suspicion / obvious suspicion>",
  "would_mention_as_problem_in_caption": true/false,
  "final_judgment": "<naturally convincing / mostly natural with minor flaws / obvious visual abnormality / suspected post-processing>",
  "reason": "<brief justification>"
}
\end{lstlisting}
\end{tcolorbox}

\subsection{Step 1: Content Summarization (Editorial Forgeries)}

\begin{tcolorbox}[
    title=\textbf{Step 1: Content Summarization},
    fonttitle=\bfseries,
    breakable,
]
Audio has been incorrectly imported; please ignore the audio. Based on the video content, visual frames, and subtitles as a whole, describe and explain the content of this video. Please note: do not describe the visual content and subtitles separately.
\end{tcolorbox}

\subsection{Step 2: Narrative Alignment Verification (Editorial Forgeries)}

\begin{tcolorbox}[
    title=\textbf{Step 2: Narrative Alignment Verification},
    fonttitle=\bfseries,
    breakable,
]
I am constructing a benchmark for misleading video detection research. Specifically, real videos are edited into corresponding misleading videos through the following methods: selective presentation editing, narrative reordering, montage manipulation, cross-video splicing, contextual misuse (visual--text mismatch), and contextual misdirection. These are compiled into a benchmark for scientific research purposes.

Below I will provide two pieces of content: the first is the model's summary of the misleading video content; the second is the editor's actual editing rationale and a brief description of the intended video content. As a professional video editor, please compare these two parts and objectively judge whether the editor's intended goal has been achieved.
\end{tcolorbox}

\section{Taxonomy, Topic Selection, and Data Statistics Details}
\label{sec:appendix_data}

This appendix provides additional motivation for the task taxonomy and topic selection described in Sections~3.2 and~3.3, followed by detailed statistics of EVID-Bench.

\subsection{Taxonomy Rationale}
\label{sec:appendix_taxonomy_rationale}

The nine task types in EVID-Bench are motivated by recurring manipulation operations documented in prior studies of video misinformation, multimedia forensics, and synthetic media. For generation-based forgeries, prior work covers face swapping, synthetic video generation, and object-level visual tampering~\cite{rossler2019faceforensics++,ho2022video,fischinger2025df}. For single-source editing, existing studies describe misleading omission, temporal reordering, and montage-based narrative construction~\cite{wardle2017information,gangal2022nareor,yin2024text}. For multi-source editing, related work studies video splicing, severity distortion, and out-of-context reuse of authentic media~\cite{pei2024multi,bu2023combating,luo2021newsclippings}. These prior observations support our organization of search-grounded video misinformation into three major categories, each with three fine-grained task types.

\subsection{Comparison with FMNV}
\label{sec:comparison-fmnv}

FMNV~\cite{wang2025fmnv} and EVID-Bench both contain manipulated video content, but they target different detection paradigms. FMNV contains 2,393 media-published news videos, including 893 real and 1,500 fake videos. Its fake samples are constructed through four augmentation strategies involving title modification, video-segment removal, synthetic voiceover, and common-sense-violating claims. Detection is formulated as binary classification using the title, video, and audio modalities of a single sample.

EVID-Bench instead focuses on evidence-dependent visual-semantic manipulation. Its samples remain internally coherent after identity, object, temporal, narrative, or contextual modification, and therefore require retrieval and comparison of external video evidence. The output is a natural-language explanation of the specific forgery points rather than a binary real/fake label. Table~\ref{tab:fmnv-comparison} summarizes these differences.

The two benchmarks are therefore complementary. FMNV evaluates whether a model can identify contradictions or implausibility within the modalities of an individual news video, whereas EVID-Bench evaluates whether a model can retrieve external evidence and explain semantic manipulations that remain coherent within the input video itself.

\begin{table}[H]
\centering
\caption{Comparison between EVID-Bench and FMNV. Although both benchmarks contain manipulated videos, they evaluate different and complementary detection paradigms.}
\label{tab:fmnv-comparison}
\small
\setlength{\tabcolsep}{5pt}
\renewcommand{\arraystretch}{1.15}
\begin{tabularx}{\textwidth}{@{} l X X @{}}
\toprule
\textbf{Dimension} & \textbf{EVID-Bench} & \textbf{FMNV} \\
\midrule
Scale
& 222 manipulated videos
& 2,393 videos: 893 real and 1,500 fake \\

Source
& YouTube videos across 6 general topic domains
& News videos from 27 media outlets on YouTube and Twitter across 12 topics \\

Construction
& Professional manual video editing and AI-based visual generation
& LLM-assisted augmentation of titles, video segments, and audio \\

Manipulation taxonomy
& 9 visual-semantic types covering AI generation, single-source editing, and multi-source editing
& 4 types covering contextual dishonesty, cherry-picked editing, synthetic voiceover, and contrived absurdity \\

Detection input
& Input-video frames and externally retrieved video evidence
& Title, video frames, and audio transcript from a single video \\

External evidence
& Required for cross-video verification
& Not required by the benchmark formulation \\

Evaluation target
& Natural-language explanation of specific forgery points
& Binary real/fake classification \\

Core challenge
& Evidence retrieval and cross-video contrastive reasoning
& Within-video cross-modal inconsistency and common-sense reasoning \\
\bottomrule
\end{tabularx}
\end{table}

\subsection{Topic Selection Rationale}
\label{sec:appendix_topic_rationale}

The six topic categories are selected to cover domains where video-based false information has been observed in prior work. User-generated fake-video corpora report topics such as video blogging, comedy, entertainment, and music~\cite{papadopoulou2019corpus}, supporting our coverage of daily-life and performance-oriented videos. News and short-video misinformation studies include lifestyle, education, sports, accidents, and epidemics~\cite{wang2024official,huang2026probing,bu2023combating,wang2025fmnv}, motivating our daily-life, education, sports, and on-scene footage categories. Product-related misinformation studies further show that review-style videos can disguise misleading claims as seemingly informative evaluations~\cite{bevendorff2024product}, motivating our reviews category.

\subsection{Source Video Collection}
\label{sec:source-video-collection}

The original, unmanipulated videos are independently sourced from publicly accessible YouTube content by the professional editors responsible for constructing the benchmark. Candidate videos are selected according to three criteria: (1) coverage of one of the six predefined topic domains; (2) sufficient visual and narrative content to support at least one of the nine manipulation types; and (3) sufficient duration and visual quality for subsequent editing and evidence comparison.

Editors adopt a bottom-up sourcing strategy rather than searching with a single fixed keyword list. This reduces concentration around a narrow set of predefined events, entities, or content patterns. During collection, we monitor the resulting pool against the task--topic matrix and add candidates to underrepresented cells to maintain broad coverage. For each selected video, we record its YouTube ID, source URL, topic category, intended manipulation type, and collection timestamp.

Because source selection requires sufficient editability and retrievability, EVID-Bench should be viewed as a controlled evaluation benchmark rather than a random sample of all YouTube videos. The balanced task--topic coverage mitigates concentration bias, but the dataset is not intended to estimate the real-world prevalence of different manipulation types.

\subsection{Dataset Distribution}
\label{sec:appendix_dataset_distribution}

Table~\ref{tab:task-topic-distribution} shows the detailed distribution of the 222 samples across the 9 task types and 6 topic categories. The samples are distributed as evenly as possible across topics within each task type, with most cells containing 4 samples. Category 1 (AI Generation) contains 72 samples, Category 2 (Single-Source Editing) contains 75 samples, and Category 3 (Multi-Source Editing) contains 75 samples.

\begin{table}[H]
\centering
\caption{Distribution of the 222 samples across task types and topic categories.}
\label{tab:task-topic-distribution}
\setlength{\tabcolsep}{3pt} 
\renewcommand{\arraystretch}{1.0}
\begin{tabular}{@{} l ccccccc @{}}
\toprule
\textbf{Task Type} & \textbf{\makecell{Daily\\life}} & \textbf{\makecell{Education\\\& skills}} & \textbf{\makecell{On-Scene\\footage}} & \textbf{Sports} & \textbf{\makecell{Performance\\\& talent}} & \textbf{Reviews} & \textbf{Total} \\
\midrule
\addlinespace[2pt]
\quad 1.1 Identity Swap       & 4 & 4 & 4 & 4 & 4 & 4 & 24 \\
\quad 1.2 Object Manipulation & 4 & 4 & 4 & 4 & 4 & 4 & 24 \\
\quad 1.3 Synthetic Insertion & 4 & 4 & 4 & 4 & 4 & 4 & 24 \\
\midrule
\addlinespace[2pt]
\quad 2.1 Selective Omission  & 4 & 4 & 4 & 4 & 5 & 4 & 25 \\
\quad 2.2 Causal Inversion    & 4 & 5 & 4 & 4 & 4 & 4 & 25 \\
\quad 2.3 Manipulative Montage& 4 & 4 & 4 & 5 & 4 & 4 & 25 \\
\midrule
\addlinespace[2pt]
\quad 3.1 Narrative Fabrication & 4 & 4 & 7 & 4 & 4 & 4 & 27 \\
\quad 3.2 Magnitude Manipulation& 5 & 4 & 2 & 4 & 4 & 4 & 23 \\
\quad 3.3 Contextual Fabrication& 4 & 4 & 4 & 4 & 5 & 4 & 25 \\
\midrule
\textbf{Total} & 37 & 37 & 37 & 37 & 38 & 36 & \textbf{222} \\
\bottomrule
\end{tabular}
\end{table}

\FloatBarrier
\Needspace{0.90\textheight}
\section{Experimental Setup}
\label{sec:appendix_setup}

Table~\ref{tab:pipeline-params} summarizes detection-pipeline hyperparameters used in all main experiments and ablations. Table~\ref{tab:eval-params} summarizes the LLM-as-judge evaluation protocol. Table~\ref{tab:evaluated-models} lists the nine frontier VLMs under test and their decoding temperature.

\begin{table}[H]
\centering
\caption{Hyperparameters of the detection pipeline.}
\label{tab:pipeline-params}
\setlength{\tabcolsep}{6pt}
\renewcommand{\arraystretch}{1.2}
\begin{tabularx}{\columnwidth}{@{} l X @{}}
\toprule
\textbf{Setting} & \textbf{Value} \\
\midrule
Frames (fine / coarse) & 64 / 16 \\
Frame height & 480p (input \& candidate) \\
Sampling & Uniform \\
Retrieval & SerpAPI \\
VLM temperature & 0.0 \\
Max rounds & 6 \\
\bottomrule
\end{tabularx}
\end{table}

\begin{table}[H]
\centering
\caption{Hyperparameters of the LLM-as-judge evaluation protocol.}
\label{tab:eval-params}
\setlength{\tabcolsep}{6pt}
\renewcommand{\arraystretch}{1.2}
\begin{tabularx}{\columnwidth}{@{} l X @{}}
\toprule
\textbf{Setting} & \textbf{Value} \\
\midrule
Benchmark split & Full EVID-Bench (222 videos) \\
Points per video & 3-5 \\
Judge models & GPT-5.4, Gemini-3.1-Pro, Claude Sonnet 4.6 \\
Judge input & Text only \\
Judge temperature & 0.0 \\
Matching & One-to-one assignment \\
Final label & Majority vote ($\geq$2/3 judges) \\
\bottomrule
\end{tabularx}
\end{table}

\begin{table}[H]
\centering
\caption{Models under test and decoding temperature. All models share the video-input settings described below.}
\label{tab:evaluated-models}
\setlength{\tabcolsep}{6pt}
\renewcommand{\arraystretch}{1.2}
\begin{tabular}{@{} l c @{}}
\toprule
\textbf{Model} & \textbf{$T$} \\
\midrule
GPT-5.5 & 0.0 \\
GPT-5.4 & 0.0 \\
GPT-5.4-mini & 0.0 \\
Claude Opus 4.6 & 0.0 \\
Claude Sonnet 4.6 & 0.0 \\
Gemini-3.1-Pro & 0.0 \\
Gemini-3-Flash & 0.0 \\
Qwen3-VL-235B-A22B-Instruct & 0.0 \\
Qwen-3.5-Plus & 0.0 \\
\bottomrule
\end{tabular}
\end{table}

\section{Multiple Experimental Runs and Variance Reporting}
\label{sec:appendix_runs_std}

For each model, we run the full benchmark three times under the same settings in Tables~\ref{tab:pipeline-params} and~\ref{tab:eval-params}: the detection pipeline on all 222 videos, followed by LLM-as-judge scoring. Main results (Tables~\ref{tab:1}--\ref{tab:4}) report the mean accuracy over these three runs; the $\pm$ values are the sample standard deviation ($n{=}3$) within each slice.

\section{Detection Pipeline Details}
\label{sec:appendix_pipeline}

Table~\ref{tab:pipeline-params} lists the key hyperparameters of the detection pipeline. Below we describe the round-level procedure and provide the exact prompts used in Stage~1 and Stage~2.

\textbf{Procedure.} (1)~Uniformly sample frames from the input video at 480p resolution. (2)~Run Stage~1 chain-of-thought analysis over 64 frames to obtain an initial YouTube query and retrieval context. (3)~Iterate for up to six DeepSearch rounds. Each round retrieves one video from SerpAPI, downloads the candidate, and runs coarse relevance filtering with 16 frames from each video. If the candidate is relevant, compare 64 frames per side to extract narrative forgery points, deduplicate them, and accumulate the result, then apply a text-only sufficiency check on the source-like evidence collected so far. (4)~When search, download, or coarse filtering fails, run multimodal reflect on the 64 input frames using failed candidate titles to propose a replacement query. Only the first returned query is executed, and negative keywords from reflect are not passed to SerpAPI. (5)~If sufficiency is not met, the sufficiency module supplies the query for the next round. Multimodal comparison prompts always present forged-video frames before candidate frames.

\paragraph{Retrieval interface and design rationale.}
The retrieval interface is kept minimal and identical across all evaluated models. In each round, the model generates one search query, which is submitted to SerpAPI. The resulting candidate video is then processed by the same coarse relevance filter and fine-grained comparison procedure for every model. The process continues for at most six rounds.

This lightweight interface is intended to evaluate model capability rather than retrieval-system engineering. In particular, performance depends on the model's ability to formulate informative queries, assess the relevance of retrieved evidence, compare the input with candidate sources, and determine whether the accumulated evidence is sufficient. Using the same retrieval interface and search budget for all models prevents differences in model-specific retrieval infrastructure from confounding the comparison.

\subsection{Expected Evidence and Evidence-Based Reasoning}
\label{sec:expected-evidence}

Both the original unmanipulated source video and other videos depicting the same event or scene can serve as valid evidence. The former enables direct comparison with the manipulated input and usually reveals the specific content that has been altered. The latter provides complementary information or alternative viewpoints that can expose semantic inconsistencies even when the exact source video is unavailable.

Our manual inspection of successful GPT-5.5 cases shows that most detections rely on retrieving the original source video or a relevant segment thereof, while a smaller proportion use other videos of the same event or scene. This pattern is consistent across AI Generation, Single-Source Editing, and Multi-Source Editing, as the benchmark manipulations are constructed from real source materials that generally remain publicly accessible online.

In successful cases, the model typically follows a coherent evidence-based reasoning process: it identifies a potentially inconsistent element, formulates a targeted search query, retrieves relevant evidence, and compares the input with the evidence to identify the specific manipulation. When this process fails, the cause can arise at different stages, including ineffective query formulation, irrelevant retrieval results, or hallucinated conclusions during evidence comparison. These failure modes are further examined in the error analysis in the main text.

\subsection{Stage 1: Chain-of-Thought Analysis}

\noindent\textbf{Inputs:} 64 uniformly sampled forged-video frames.

\begin{tcolorbox}[
    title=\textbf{Chain-of-Thought Retrieval Planning},
    fonttitle=\bfseries,
    breakable,
]
You are a video forensic analyst. You will see 64 frames uniformly sampled from a video that may contain misinformation.

Your task is to deeply understand the video content, identify key entities and potential logical anomalies, and produce the initial retrieval plan for finding original or closely related source videos on YouTube.

Think step by step (write your full reasoning in the `reasoning` field):

== Step 1: Content Understanding ==

Examine the 64 frames as a timeline:

a) What different scenes, environments, or camera setups appear in the video? Briefly describe each one.

b) Identify the key entities and information in the video:
   - People: Who appears? Can you identify specific individuals such as celebrities, athletes, presenters, politicians, or creators?
   - Locations: Are there identifiable landmarks, venue cues, signs, architectural styles, neighborhoods, or countries?
   - Events: What is happening? A sports match, program segment, performance, interview, protest, news event, travel clip, everyday scene?
   - Time: Can you infer dates, seasons, eras, or broadcast periods?
   - On-screen text: What readable text seems semantically meaningful for identifying the content, such as names, event titles, program logos, scoreboards, venue text, headlines, or captions that appear integral to the footage? Ignore obvious platform chrome like playback controls, timestamps, or menus.

c) Are there discontinuities between scenes, such as sudden changes in environment, production style, resolution, or topic, suggesting multiple original sources?

== Step 2: Logical Reasoning ==

Combine visual information, text information, and common sense to look for potential anomalies or verification targets. Consider ALL of the following:

NARRATIVE-LEVEL ANOMALIES:
- Are the facts claimed or implied by the video internally consistent?
- Are there unreasonable links between people, locations, times, or events across scenes?
- Are there scenes that look edited together from different sources?

AI-GENERATION ANOMALIES:
- Do any objects appear visually inconsistent with their surroundings (wrong lighting, unnatural texture, mismatched brand/style for the context)?
- Are there objects or products that seem contextually out of place (e.g., a branded product appearing in a setting where it would not naturally exist)?
- Do any faces look inconsistent with the person's body, clothing, or prior appearance in other frames?
- Are there segments where visual quality, motion dynamics, or rendering style suddenly differs from the rest?

VERIFICATION TARGET:
- Is there a specific claim, identity, location, or event that is most worth verifying first?
- If there is no obvious contradiction, explain what the most valuable verification target would be.
- If AI-generation is suspected, note which elements may be synthetic so they are NOT used as retrieval anchors.

== Step 3: Retrieval Planning ==

Based on Step 1 and Step 2, determine the single best initial search query to begin retrieval:

- What is the most promising verification target to search for first, and why?
- Estimate how many distinct original sources are likely present (this informs later rounds, but you only need ONE query now).
- Produce exactly ONE high-value YouTube search query of 5-12 words targeting the most verifiable and specific element you identified.

Query strategy:
- Prioritize specific people, event names, program names, competition names, locations, dates, and distinctive scene cues.
- If a named person + event is identifiable, combine them directly.
- If a program, broadcast, or competition is identifiable, use that as a strong anchor.
- If no specific entity is reliable, use the content category plus the most distinctive visual cue.
- The query should sound like a plausible YouTube title fragment or high-value search phrase, not a forensic note.
- Do NOT generate multiple queries. Subsequent queries will be generated adaptively based on retrieval results from this first attempt.

OUTPUT JSON FORMAT:
\begin{lstlisting}
{
  "reasoning": "Step 1: <content understanding>. Step 2: <logical reasoning>. Step 3: <why this query is the best starting point>.",
  "entities": {
    "people": ["identified people"],
    "locations": ["identified locations"],
    "events": ["identified events/programs/competitions"],
    "text_claims": ["key facts or labels conveyed by on-screen text"]
  },
  "logical_analysis": "Logical analysis of the video content, including contradictions found or the main fact that should be verified first.",
  "physical_observations": "Summarize the most searchable visual clues in 3-5 sentences.",
  "search_intent": "What this retrieval should verify or reconstruct first.",
  "estimated_sources": 1,
  "source_queries": [
    {
      "source_label": "source_1: brief description of the most promising verification target",
      "queries": ["single best first query (5-12 words)"]
    }
  ]
}
\end{lstlisting}
\end{tcolorbox}

\subsection{Stage 2: Coarse Relevance Filter}

\noindent\textbf{Inputs:} 16 forged + 16 candidate frames (Group~A first, then Group~B).

\begin{tcolorbox}[
    title=\textbf{Coarse Relevance Filter},
    fonttitle=\bfseries,
    breakable,
]
You are a coarse filter in a video retrieval pipeline.

You will see TWO frame groups:
- Group A: frames from the SUSPECTED FORGERY video.
- Group B: frames from a CANDIDATE YouTube video.

CONTEXT FROM PRIOR COT ANALYSIS:
- Physical observations: \{physical\_observations\}
- Logical analysis: \{logical\_analysis\}
- Search intent: \{search\_intent\}

YOUR TASK: judge whether Group B is useful retrieval evidence for Group A. The candidate can be:
- the same original footage,
- a closely related upload of the same event or scene family,
- or a video that would materially help verify the main fact or claim in Group A.

CRITICAL EVALUATION RULES:
1. This is a COARSE FILTER with HIGH RECALL. Prefer false positives over false negatives.
2. Do NOT rely on pixel-level similarity.
3. Ignore differences caused by compression, bitrate, resolution, color grading, subtitles, black bars, crop, mirror, small timing offset, and camera/viewpoint variation.
4. Focus on semantic anchors: event type, named entities, venue/program cues, recurring objects, role relations, action pattern, and storyline.
5. If Group B could plausibly help verify the same identity, event, location, or claim as Group A, mark relevant.
6. Only mark irrelevant when topics/scenes are clearly different.

OUTPUT JSON FORMAT:
\begin{lstlisting}
{
  "reasoning": "Brief semantic comparison between Group A and Group B.",
  "is_relevant": true
}
\end{lstlisting}
\end{tcolorbox}

\subsection{Stage 2: Fine-Grained Forgery Analysis}

\noindent\textbf{Inputs:} 64 forged + 64 candidate frames (Group~A first, then Group~B).

\begin{tcolorbox}[
    title=\textbf{Fine-Grained Forgery Analysis},
    fonttitle=\bfseries,
    breakable,
]
You are a forensic video analyst comparing:
- Group A: frames from a SUSPECTED FORGERY video
- Group B: frames from a CANDIDATE source video

PRIOR CONTEXT:
- Physical observations: \{physical\_observations\}
- Logical analysis: \{logical\_analysis\}
- Search intent: \{search\_intent\}
- Entities summary: \{entities\_summary\}

YOUR TASK: Perform a detailed cross-group comparison and extract forgery points that describe WHAT IS FALSE in Group A, using Group B as ground-truth evidence.

Forgeries fall into three broad categories --- consider ALL of them:

A) AI GENERATION: faces replaced by deepfake (identity swap), objects added/removed/replaced by generative editing, or contiguous AI-generated segments inserted into real footage. Clues: identity mismatch (same body/background but different face), objects that exist in A but not in B (or vice versa), segments in A with no corresponding content in B that show telltale synthetic continuity.

B) SINGLE-SOURCE EDITING: the same source footage is editorially altered --- scenes selectively omitted to change meaning, temporal order reordered to flip causality, or shots rearranged to create a misleading emotional impression.

C) MULTI-SOURCE SPLICING: footage from Group B is combined with footage from other unrelated sources to fabricate a narrative, distort the severity of an event, or place real footage into a false context (wrong time, location, or attribution).

THINK IN THIS ORDER (write reasoning in `source\_description`):
1. IDENTITY \& OBJECT ALIGNMENT: Compare the people, faces, and key objects between Group A and Group B. Are the same individuals present? Are faces consistent? Are objects added, removed, or replaced?
2. TEMPORAL \& STRUCTURAL ALIGNMENT: Do the scenes in Group A appear in the same order and completeness as in Group B? Are segments missing, reordered, or inserted?
3. CONTEXTUAL ALIGNMENT: Does Group A preserve the original context (time, location, event, attribution) shown in Group B, or does it reframe the footage into a different narrative?
4. SYNTHESIS: What is the overall false impression that Group A creates, and which category (AI generation / single-source editing / multi-source splicing) best describes the manipulation?
5. DEDUPLICATE: each point must cover a unique manipulation aspect.

OUTPUT JSON FORMAT (strict, no extra text, NO code fences):
\begin{lstlisting}
{
  "source_description": "1-2 sentences: semantic relation between Group A and Group B plus the primary forgery mechanism identified.",
  "points": [
    {
      "description": "Forgery point: what Group B shows as ground truth, what differs in Group A, and the false impression this creates."
    }
  ]
}
\end{lstlisting}

CONSTRAINTS:
1. `points` should contain 3 to 5 entries when evidence supports it.
2. No duplicate points with paraphrased wording.
3. Ignore pure quality/viewpoint/compression differences unless they support a manipulation claim.
4. At least one point should provide a high-level synthesis of the overall manipulation (not just one local edit or one replaced element).
5. For identity swaps: describe the original person's appearance in Group B (hair color/style, clothing, build, accessories) and state that Group A replaces their face with a different person's face via AI. Do this for EACH person whose face is swapped.
6. For object manipulation: describe the original object in Group B (what it is, where it appears, its role in the scene) and state what it has been replaced with or altered into in Group A. Do this for EACH manipulated object.
7. Output VALID JSON only. No markdown fences, no commentary.
\end{tcolorbox}

\subsection{Stage 2: Sufficiency Judgment and Next Query}

\noindent\textbf{Inputs:} text-only context fields; no visual input.

\begin{tcolorbox}[
    title=\textbf{Sufficiency Judgment and Next Query},
    fonttitle=\bfseries,
    breakable,
]
You are the decision-maker in a deep-search loop for video forgery analysis. You have been searching YouTube for original source videos and collecting forgery evidence.

CONTEXT (do NOT echo back):

- Round status:
\{round\_status\}

--- What the video looks like (from COT analysis of the input video):
- Physical observations: \{physical\_observations\}
- Logical analysis: \{logical\_analysis\}
- Search intent: \{search\_intent\}
- Entities summary: \{entities\_summary\}

--- Estimated source families / scene groups:
\{source\_descriptions\}

--- Forgery evidence collected so far:
\{collected\_points\}

--- Videos examined so far:
\{examined\_videos\}

--- Previous search queries tried:
\{prev\_queries\}

YOUR JOB (two tasks in one response):

TASK 1 --- SUFFICIENCY JUDGMENT: Compare what you KNOW about the input video against what you've COLLECTED. Judge whether the evidence is SUFFICIENT.

THINK STEP BY STEP for sufficiency:
1. Based on the source descriptions, how many distinct source families or scenes likely matter? Have we found evidence for each?
2. Based on physical\_observations and logical\_analysis, what identities, events, locations, or claims must be verified? Do the collected points cover them?
3. Does the examined evidence explain the main misleading interpretation or unresolved factual question?
4. Is there any major scene, claim, or verification target that NO collected point covers?

CRITICAL RULES:
- For a SINGLE dominant source: do NOT stop merely because you found a topically related video. The examined evidence should strongly suggest the same program, same event, same source family, or a near-duplicate upload of the same underlying footage.
- If the current evidence mainly highlights that Group A and Group B are from different eras, different shows, different productions, or different contexts, that is evidence of topical relatedness, NOT sufficient source resolution. Keep searching.
- For MULTI-SOURCE or MULTI-SCENE cases: you need evidence that covers EACH major source family or unresolved scene cluster.
- Prefer precision over early stopping. It is acceptable to continue searching when the current best evidence is only ``related but not same-source enough.''
- If the round status says this is the FINAL round, you MUST set `is\_sufficient=true`, leave `missing\_description` and `next\_keyword` empty, and do not request another search.

TASK 2 --- NEXT KEYWORD (only if NOT sufficient): Generate the SINGLE best YouTube search query to find the next missing original source.

THINK STEP BY STEP for keyword generation:
1. Which source family, identity, event, location, or claim is still unresolved?
2. What specific distinguishing cue would most likely retrieve useful evidence for it?
3. Generate ONE search query (5-12 words) using the strongest identifying anchor available.
4. Prefer a query that could plausibly match a real YouTube title or high-value search phrase.

OUTPUT JSON FORMAT:
\begin{lstlisting}
{
  "reasoning": "Step-by-step: what the forgery contains vs what evidence is collected, whether the current evidence is same-source enough, and what's still missing.",
  "is_sufficient": true | false,
  "missing_description": "If insufficient, describe exactly what evidence is missing. If sufficient, empty string.",
  "next_keyword": "If insufficient, ONE best search query (5-12 words). If sufficient, empty string."
}
\end{lstlisting}
\end{tcolorbox}

\subsection{Stage 2: Query Refinement (Reflect)}

\noindent\textbf{Inputs:} 64 forged-video frames; triggered when search, download, or coarse filtering fails.

\begin{tcolorbox}[
    title=\textbf{Query Refinement (Reflect)},
    fonttitle=\bfseries,
    breakable,
]
You are a video retrieval strategist working inside an iterative search loop. Previous YouTube queries did NOT find a useful source video.

CONTEXT OF FAILURE:
- Previously tried queries: \{prev\_queries\}
- WRONG candidate titles returned by those queries: \{candidate\_titles\}

PRIOR COT ANALYSIS OF THE INPUT VIDEO:
- Physical observations: \{physical\_observations\}
- Logical analysis: \{logical\_analysis\}
- Search intent: \{search\_intent\}
- Entities summary: \{entities\_summary\}

YOUR JOB: look at the input frames again, study the failed candidate titles, diagnose why retrieval drifted, and produce ONE stronger follow-up query.

THINK STEP BY STEP (write your full reasoning in the `reasoning` field):
Step 1. Re-understand the input video from the frames:
  - What are the most diagnostic anchors: named people, program names, event names, competition names, locations, dates, logos, uniforms, scene types, or specific actions?
  - Which of those anchors are most likely to appear in the original uploader's title or in strong retrieval results?

Step 2. Diagnose the failure mode:
  - Did previous queries over-focus on the wrong person, event, program, location, or date?
  - Were they too broad, too narrow, or anchored on a side detail instead of the main verifiable fact?
  - Did the wrong candidate titles suggest a different topic cluster than what the input frames actually show?

Step 3. Produce `negative\_keywords` conservatively:
  - Add ONLY specific entities, titles, locations, dates, or topic anchors that clearly pulled retrieval into the wrong result cluster.
  - Do NOT add broad genre words that could still belong to the right title.
  - When unsure, leave a term out. It is better to keep a reusable word than to block a potentially correct one.

Step 4. Generate 1 fresh search query:
  - Choose the single best angle that is MOST DIFFERENT from all previously tried queries.
  - You are free to use ANY retrieval angle: entity/event/title, location/date/program, visual cue, content category, or any combination.
  - The query must NOT repeat the same approach that already failed.

QUERY RULES:
- The query must be plain YouTube search text, 5-12 words.
- It should sound like a plausible title fragment or high-value search phrase, not a forensic note.
- Avoid reusing any term in `negative\_keywords` unless it is part of an explicit exclude operator like `-term`.

OUTPUT JSON FORMAT:
\begin{lstlisting}
{
  "reasoning": "Step 1: <what the video most likely shows>. Step 2: <why previous searches failed>. Step 3: <which specific anchors were treated as negatives>. Step 4: <how the new query improves retrieval>.",
  "reflection": "Short explanation of what retrieval signal was re-prioritized.",
  "negative_keywords": ["wrong_entity_or_topic_1", "wrong_entity_or_topic_2"],
  "new_queries": ["single best query"]
}
\end{lstlisting}
\end{tcolorbox}

\section{Evaluation Protocol Details}
\label{sec:appendix_eval}

Table~\ref{tab:eval-params} summarizes evaluation settings. Three text-only LLM judges independently compare each video's 3--5 predicted points with 3--5 ground-truth points using one-to-one semantic matching; each GT point is marked correct if at least two judges agree. The exact judge prompt is given below.

\subsection{Task Input, Output, and Metric Definitions}
\label{sec:task-metric-definitions}

\paragraph{Input.}
For each input video, the evaluated model receives 64 uniformly sampled visual frames. The audio channel is not provided to the model. Subtitles, captions, and other text embedded directly in the frames are retained because they form part of the visual content and may provide evidence relevant to the manipulation. The coarse relevance filter separately uses 16 frames from the input and candidate videos, as described in Appendix~\ref{sec:appendix_pipeline}.

\paragraph{Output.}
The expected output is a list of natural-language forgery-point descriptions, with each point describing one specific manipulation identified in the input video. For example, a point may state that ``the face of person A is replaced with the face of person B,'' or that ``clips from two unrelated events are spliced together to fabricate a causal relationship.'' The model is not required to predict the manipulation category or the temporal boundaries of the manipulated segment. The task focuses on explaining what information has been manipulated rather than assigning a predefined class or localizing an anomalous interval.

\paragraph{Metric definitions.}
For video $i$, let $g_i$ denote the number of ground-truth forgery points, where $g_i \in \{3,4,5\}$, and let $m_i$ denote the number of ground-truth points matched by the model output. Predicted and ground-truth points are matched through a one-to-one assignment using the LLM-as-judge protocol described below. Each ground-truth point can be matched at most once.

Point-level accuracy measures the proportion of ground-truth forgery points successfully matched across the benchmark:
\begin{equation}
\mathrm{PointAcc}
=
\frac{\sum_i m_i}{\sum_i g_i}.
\end{equation}

Video-level accuracy requires all ground-truth points of a video to be matched:
\begin{equation}
\mathrm{VideoAcc}
=
\frac{1}{N}
\sum_i
\mathbf{1}[m_i=g_i].
\end{equation}

Thus, point-level accuracy measures partial coverage of the annotated manipulations, whereas video-level accuracy measures whether the model provides a complete explanation of the manipulated video. Human validation of the LLM-based matching protocol is provided in Appendix~\ref{sec:appendix_human_agreement}.

\subsection{Stage 3: LLM-as-Judge Scoring}

\noindent\textbf{Inputs:} numbered GT and predicted forgery-point lists (text only; no visual input).

\begin{tcolorbox}[
    title=\textbf{LLM-as-Judge Forgery-Point Scoring},
    fonttitle=\bfseries,
    breakable,
]
You are a strict but fair evaluator for video forgery analysis. You will be given a ground-truth (GT) list of 3-5 forgery points and a predicted list of 3-5 forgery points. For each GT point, decide whether any prediction matches it.

Important framing:
- GT points and predictions may be written at different abstraction levels.
- Some items are narrative-level accusations: the edit constructs a false moral, political, social, or causal interpretation.
- Some items are event-level findings: the edit splices, reorders, re-contextualizes, or rebinds specific scenes, audio, or actions.
- A good prediction may express the same core misleading point either directly at the narrative level or indirectly through concrete event evidence.

Scoring rules (important; follow strictly):
1. Each GT point can be matched at most once. Use a one-to-one best assignment.
2. Evaluate every GT/pred pair on two dimensions:
   - Dimension A: manipulation method. Examples: splicing unrelated clips, reordering scenes, removing narration, binding unrelated audio, selective omission, context switching.
   - Dimension B: misleading point / false reframing. Examples: portraying someone as rule-breaking, constructing a negative personality arc, implying opportunistic recruitment, or creating a moral-judgment framing.
3. Dimension B is primary, but judge it flexibly across abstraction levels:
   - Direct narrative match: If the prediction states substantially the same false reframing or accusation as the GT, mark `matched\_dim\_misleading=true` and `verdict=1`.
   - Event-to-narrative support: If the GT is narrative-level, and the prediction identifies the concrete edited event(s) that materially support that same narrative reframing, you may still mark `matched\_dim\_misleading=true` and `verdict=1` even if the prediction is more concrete and less interpretive.
4. ``Roughly consistent'' allows paraphrase, synonymy, wording differences, partial compression of multiple scenes into one point, and different abstraction levels, as long as the core misleading implication is aligned.
5. Be conservative about over-crediting:
   - Do not give credit just because both items mention the same event.
   - Do not give credit just because both items describe editing manipulation in general.
   - Do give credit when the prediction captures the same accusation, reframing, or a concrete event-level mechanism that clearly underpins that accusation.
6. Report `matched\_dim\_method` truthfully even when `verdict=1` is driven mainly by Dimension B.
7. Every GT point should receive the best available predicted match. If multiple GT points compete for the same prediction, choose the assignment that gives the best overall scoring.
8. Output strict JSON only. Do not use markdown fences.

GT (3-5 points):
\{gt\_block\}

PREDICTION (3-5 points):
\{pred\_block\}

Output JSON:
\begin{lstlisting}
{
  "hits": <integer from 0 to number_of_gt_points>,
  "score": <hits / number_of_gt_points as a float>,
  "matches": [
    {
      "gt_idx": <0-based GT index>,
      "pred_idx": <0-based prediction index or null if no match>,
      "verdict": <0|1>,
      "matched_dim_method": <true|false>,
      "matched_dim_misleading": <true|false>,
      "reason": "One concise English sentence explaining whether the method matches, whether the misleading point matches, and the key mismatch if any."
    }
  ],
  "comment": "1-2 concise English sentences summarizing the biggest strength and biggest weakness of the predictions."
}
\end{lstlisting}
\end{tcolorbox}

\section{Human Evaluation Agreement}
\label{sec:appendix_human_agreement}

To check whether LLM-as-judge scoring tracks human judgment, we sample 50 EVID-Bench videos stratified across the nine task types (182 ground-truth points) and evaluate predictions from GPT-5.5 and Qwen-3.5-Plus. Three annotators independently label each ground-truth point using the same matching rubric as the LLM judges; labels are merged by majority vote.

\begin{figure}[H]
  \centering
  \includegraphics[width=\linewidth]{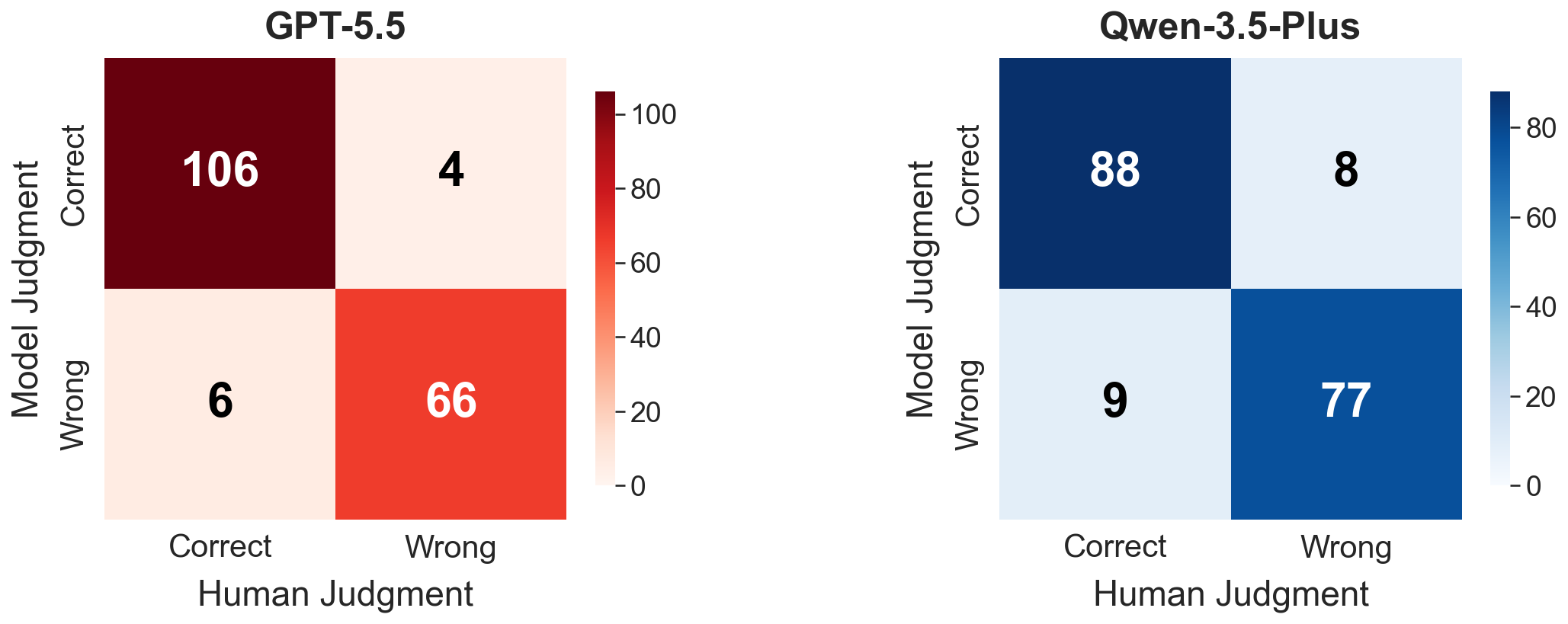}
  \caption{Point-level confusion matrices comparing LLM majority-vote labels with human majority labels on 182 forgery points from 50 stratified videos. \textbf{Left:} GPT-5.5 predictions. \textbf{Right:} Qwen-3.5-Plus predictions.}
  \label{fig:human_confusion_matrix}
\end{figure}

Figure~\ref{fig:human_confusion_matrix} compares these human labels with the LLM majority votes used in the main experiments. Agreement is close for both models.

Overall agreement reaches 94.51\% for GPT-5.5 predictions and 90.66\% for Qwen-3.5-Plus predictions. Cohen's $\kappa$ is 0.886 and 0.811, respectively, indicating strong agreement between human and LLM-based judgments.

Our judge prompt requires one-to-one matching with a specific manipulation mechanism. A generic statement that a video is misleading or manipulated is not sufficient unless it identifies the corresponding identity, object, temporal, narrative, or contextual modification. As shown in Figure~\ref{fig:human_confusion_matrix}, false positives---cases judged correct by the LLM majority but incorrect by the human majority---occur in only 4 of 182 GPT-5.5 judgments and 8 of 182 Qwen-3.5-Plus judgments. The corresponding false-negative counts are 6 and 9. This disagreement analysis suggests that over-crediting generic narrative falsehoods is uncommon under our evaluation protocol.

\section{Additional Validation Experiments}
\label{sec:additional-validation}

\subsection{Human Performance Baseline}
\label{sec:human-performance-baseline}

Three computer science researchers with video-editing experience independently evaluated 54 EVID-Bench videos, with six videos sampled from each of the nine manipulation types. Participants were allowed to use web search to locate original or related video evidence.

For video $i$, let $h_i$ denote the number of matched ground-truth forgery points and $g_i$ denote the total number of ground-truth points. Point-level accuracy is computed over all ground-truth points, while a video is counted as correct only when all of its ground-truth points are identified:
\begin{equation}
\mathrm{PointAcc}
=
\frac{\sum_i h_i}{\sum_i g_i},
\qquad
\mathrm{VideoAcc}
=
\frac{1}{N}
\sum_i
\mathbf{1}[h_i=g_i].
\end{equation}

\begin{table}[H]
\centering
\caption{Human performance (\%) on the 54-video subset.}
\label{tab:human-performance}
\small
\setlength{\tabcolsep}{7pt}
\renewcommand{\arraystretch}{1.1}
\begin{tabular}{@{}lcc@{}}
\toprule
\textbf{Category} & \textbf{Point Acc.} & \textbf{Video Acc.} \\
\midrule
AI Generation          & 64.81 & 44.44 \\
Single-Source Editing  & 61.97 & 38.89 \\
Multi-Source Editing   & 48.00 & 27.78 \\
\midrule
Overall                & 57.50 & 37.04 \\
\bottomrule
\end{tabular}
\end{table}

\subsection{Authentic-Video Evaluation}
\label{sec:authentic-video-evaluation}

We evaluate over-flagging using an auxiliary set of 60 authentic, unmanipulated videos, with 10 videos from each of the six topic domains. A prediction is counted as correct when the model produces no forgery points. This metric measures the proportion of authentic videos that are correctly left unflagged.

Claude Opus 4.6 and Claude Sonnet 4.6 both achieve 95.0\% accuracy on authentic videos, indicating a low over-flagging rate. Gemini-3.1-Pro achieves 83.3\%, with more false positives on Reviews and other casually edited videos. Across models, Daily Life and Reviews are more prone to false positives, suggesting that routine cuts, transitions, and other benign editing patterns may be mistaken for intentional manipulation.

\begin{table}[H]
\centering
\caption{Accuracy (\%) on the 60 authentic videos. A video is counted as correct when the model produces no forgery points.}
\label{tab:authentic-video-results}
\small
\setlength{\tabcolsep}{6pt}
\renewcommand{\arraystretch}{1.1}
\begin{tabular}{@{}lccc@{}}
\toprule
\textbf{Topic}
& \textbf{Gemini-3.1-Pro}
& \textbf{Claude Opus 4.6}
& \textbf{Claude Sonnet 4.6} \\
\midrule
On-Scene Footage   & 80  & 100 & 100 \\
Sports             & 80  & 100 & 100 \\
Performance/Talent & 90  & 100 & 100 \\
Education/Skills   & 100 & 100 & 100 \\
Daily Life         & 80  & 80  & 90  \\
Reviews            & 70  & 90  & 80  \\
\midrule
Overall            & 83.3 & 95.0 & 95.0 \\
\bottomrule
\end{tabular}
\end{table}

\subsection{Forensic Detector Baselines}
\label{sec:forensic-detector-baselines}

AIDE~\cite{yan2025sanity} and UnivFD~\cite{ojha2024universal} produce frame-level fake probabilities. We uniformly sample 64 frames from each video and aggregate frame-level predictions using three rules: mean probability at least 0.5, at least half of the frames at or above 0.5, and any frame at or above 0.5.

Under the mean and majority aggregation rules, both detectors achieve near-zero detection rates across the three manipulation categories. The max rule produces higher detection rates, but also yields high false-positive rates on authentic videos: AIDE flags 55.0\% of authentic videos, while UnivFD flags 78.3\%. Consequently, the increased sensitivity of the max rule does not translate into reliable discrimination between manipulated and authentic content.

These results provide a visual-only forensic sanity check rather than evidence that all visual detection is impossible. They show that the two tested detectors, which primarily target pixel-level generative artifacts, do not reliably detect the evidence-dependent semantic manipulations in EVID-Bench. We therefore scope our conclusion to current forensic detectors and frontier VLMs evaluated under the reported protocols.

\begin{table}[H]
\centering
\caption{Video-level detection rates (\%) under three frame-level aggregation rules.}
\label{tab:forensic-detector-results}
\small
\setlength{\tabcolsep}{5pt}
\renewcommand{\arraystretch}{1.1}
\begin{tabular}{@{}llccc@{}}
\toprule
\textbf{Detector} & \textbf{Category}
& \textbf{Mean} & \textbf{Majority} & \textbf{Max} \\
\midrule
AIDE & AI Generation         & 4.2 & 4.2 & 70.8 \\
AIDE & Single-Source Editing & 0.0 & 0.0 & 58.7 \\
AIDE & Multi-Source Editing  & 0.0 & 0.0 & 60.0 \\
AIDE & Authentic             & 0.0 & 1.7 & 55.0 \\
\midrule
UnivFD & AI Generation         & 0.0 & 0.0 & 69.4 \\
UnivFD & Single-Source Editing & 0.0 & 0.0 & 81.3 \\
UnivFD & Multi-Source Editing  & 0.0 & 0.0 & 92.0 \\
UnivFD & Authentic             & 0.0 & 0.0 & 78.3 \\
\bottomrule
\end{tabular}
\end{table}

\subsection{Reverse Image Search Baseline}
\label{sec:reverse-image-search}

We evaluate whether reverse image search can retrieve the original source videos independently of the downstream forgery analysis. For each input video, we uniformly sample 16 frames and submit each frame independently to Google Lens. We collect the top-10 results for each frame, yielding up to 160 candidate URLs per video after deduplication. Retrieval is considered successful if at least one returned URL corresponds to the original source video. For comparison, the iterative text-retrieval pipeline examines at most six candidate videos per input.

Despite examining substantially more candidate URLs, Google Lens achieves 26\% source recall, compared with up to 60\% for iterative text retrieval. Reverse image search succeeds when a query frame matches an indexed thumbnail, cover image, or other web-visible visual content, but arbitrary internal video frames are not reliably indexed as searchable video evidence. In contrast, text retrieval uses identifiable entities, events, and locations to search video titles, descriptions, and associated metadata. These results suggest that reverse image search is useful as a complementary retrieval channel but is insufficient as the primary retrieval method for EVID-Bench.

\begin{table}[H]
\centering
\caption{Source recall of reverse-image and iterative text-retrieval methods. Source recall measures whether the original source video is retrieved for an input video.}
\label{tab:source-recall-results}
\small
\setlength{\tabcolsep}{5pt}
\renewcommand{\arraystretch}{1.1}
\begin{tabularx}{\linewidth}{@{}Xcc@{}}
\toprule
\textbf{Method} & \textbf{Budget} & \textbf{Recall} \\
\midrule
Google Lens ($16\times$ top-10) & $\leq 160$ URLs & 26\% \\
Claude Opus 4.6                 & $\leq 6$ videos & 54\% \\
Claude Sonnet 4.6               & $\leq 6$ videos & 52\% \\
GPT-5.4                         & $\leq 6$ videos & 52\% \\
GPT-5.5                         & $\leq 6$ videos & 60\% \\
Gemini-3-Flash                  & $\leq 6$ videos & 30\% \\
Gemini-3.1-Pro                  & $\leq 6$ videos & 32\% \\
Qwen-3.5-Plus                   & $\leq 6$ videos & 31\% \\
Qwen3-VL-235B-A22B-Instruct     & $\leq 6$ videos & 23\% \\
\bottomrule
\end{tabularx}
\end{table}

\FloatBarrier
\Needspace{0.60\textheight}
\section{Ablation Study (Qwen-3.5-Plus)}
\label{sec:appendix_ablation_qwen}

\begin{figure}[H]
  \centering
  \includegraphics[width=\linewidth]{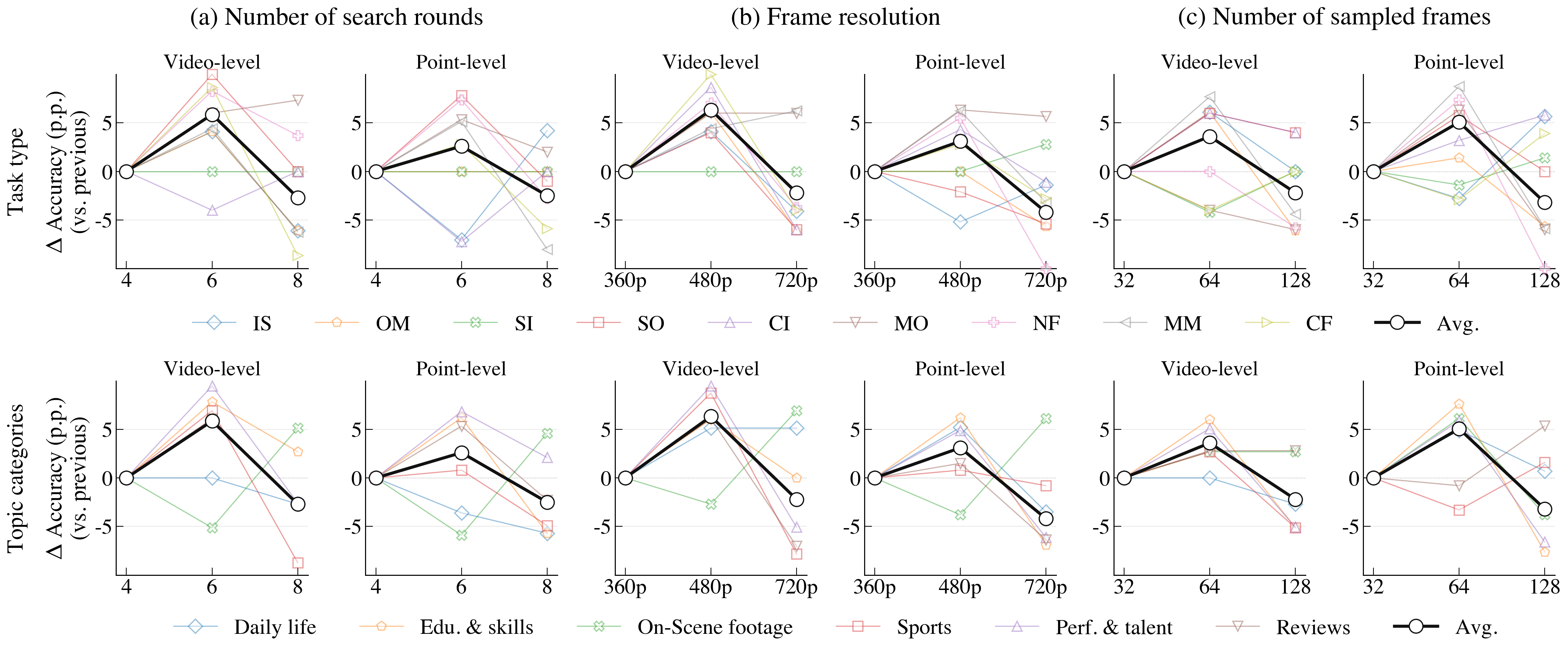}
  \caption{Same ablation layout as Figure~\ref{fig:ablation_gpt} for Qwen-3.5-Plus.
  Trends match GPT-5.5, peaking at 6 rounds, 480p, and 64 frames. Curves plot stepwise accuracy change in percentage points: $\Delta_1=0$ at the leftmost setting and $\Delta_j=\mathrm{Acc}_j-\mathrm{Acc}_{j-1}$ for $j\ge 2$.}
  \label{fig:ablation_qwen}
\end{figure}

The main text reports the GPT-5.5 ablation in Figure~\ref{fig:ablation_gpt}.
Figure~\ref{fig:ablation_qwen} presents the corresponding breakdown for Qwen-3.5-Plus using the same layout: three hyperparameters (search rounds, frame resolution, and sampled frames), split by video- and point-level accuracy across task types and topic categories.
Qwen-3.5-Plus follows the same overall trends as GPT-5.5, peaking at 6 rounds, 480p, and 64 frames, though with lower accuracy across most cells.

\FloatBarrier
\Needspace{0.70\textheight}
\section{Case Study}

\Needspace{0.50\textheight}
\subsection{Excess Search Rounds Introduce Irrelevant Evidence}

\begin{tcolorbox}[
    title=\textbf{Case Study: Excess search rounds introduce irrelevant evidence},
    fonttitle=\bfseries,
    breakable,
    fontupper=\small
]

\label{app:case_ablation_rounds}

\includegraphics[width=0.95\linewidth]{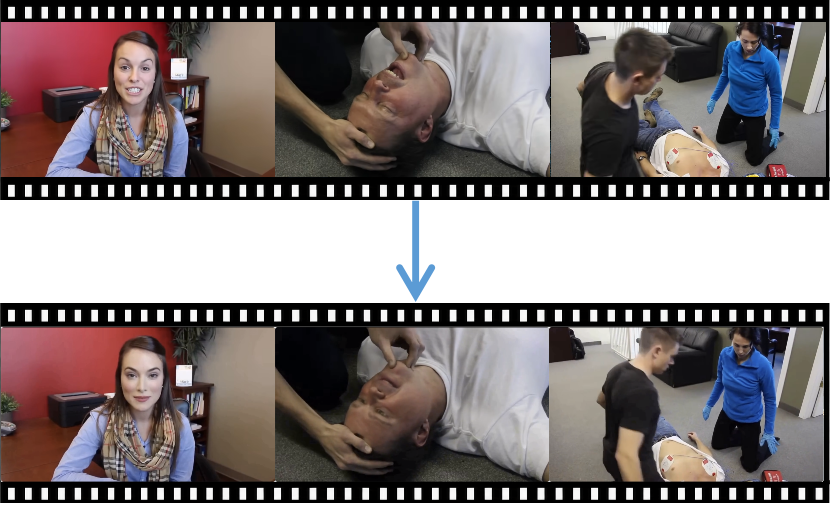}

\paragraph{Overview.}
Increasing the search-round budget from 6 to 8 does not simply add more supporting evidence; it can retrieve a second, loosely related video that reframes the entire analysis. The model then mixes correct observations from the original source with spurious comparisons drawn from the wrong candidate, diluting the true forgery explanation.

\paragraph{Task Description.}
We use a CPR/AED training video in which three instructors' faces are replaced with AI-generated identities while the underlying demonstration footage remains unchanged. The ground truth requires identifying these three face swaps:
\begin{quote}
``The fake video replaces her face with another person's face using AI.'' \\
``The fake video replaces his face with another person's face using AI.'' \\
``The fake video replaces her face with another person's face using AI.''
\end{quote}

\paragraph{Model Behavior.}
With 6 search rounds (480p, 64 frames), the pipeline retrieves one evidence video---\emph{CPR / AED Emergency Response Refresher}---and produces five forgery points centered on identity swap. It correctly describes, for example:
\begin{quote}\small\texttt{``Receptionist Face Swap: In Group B (Frame 65) ... In Group A (Frame 2), her face has been replaced with a different face via AI ...''}\end{quote}
\begin{quote}\small\texttt{``Narrator Face Swap: In Group B (Frame 113) ... In Group A (Frame 47), her face appears to be AI-generated or swapped ...''}\end{quote}
All three ground-truth points are matched (video-level score = 1.0).

With 8 search rounds under the same resolution and frame budget, the pipeline retains the original evidence video but additionally retrieves \emph{How to Give CPR and Use an AED | First Aid Tips}, a generic CPR tutorial from a different provider. Four of the first five collected points now come from this unrelated video and reinterpret the case through low-level artifacts rather than identity swap:
\begin{quote}\small\texttt{``Group A displays distorted and wavy text on the mannequin shirts (e.g., `CALL 911 GET AN AED' in Frame 58) ... hallmark artifacts of AI video generation ...''}\end{quote}
\begin{quote}\small\texttt{``Group A demonstrates the outdated `ABC' (Airway, Breathing, Compressions) protocol ... whereas Group B correctly follows the modern `CAB' protocol ...''}\end{quote}
\begin{quote}\small\texttt{``Group A presents a completely fabricated branding context (`Action First Aid') ... whereas Group B is authentically branded as the `Canadian Red Cross' ...''}\end{quote}
Only one of three ground-truth identity-swap points is recovered (video-level score = 0.33).

\paragraph{Result.}
The 8-round run does not fail because retrieval stops working---it fails because an extra round introduces a plausible but wrong comparison target. The model then explains the fake video as AI-generated instructional content or branding fraud against an unrelated Red Cross tutorial, rather than as face replacement on the original source.

\paragraph{Interpretation.}
This case shows that additional search rounds can hurt when sufficiency is judged too loosely. Once a second candidate enters the evidence pool, the model treats superficial instructional differences as primary forgery signals and underweights the actual manipulation mechanism. More retrieval is helpful only while it still converges on the correct source family.

\end{tcolorbox}

\Needspace{0.50\textheight}
\subsection{Higher Resolution Distracts from Narrative Forgery}

\begin{tcolorbox}[
    title=\textbf{Case Study: Higher resolution distracts from narrative forgery},
    fonttitle=\bfseries,
    breakable,
    fontupper=\small
]

\label{app:case_ablation_resolution}

\includegraphics[width=0.95\linewidth]{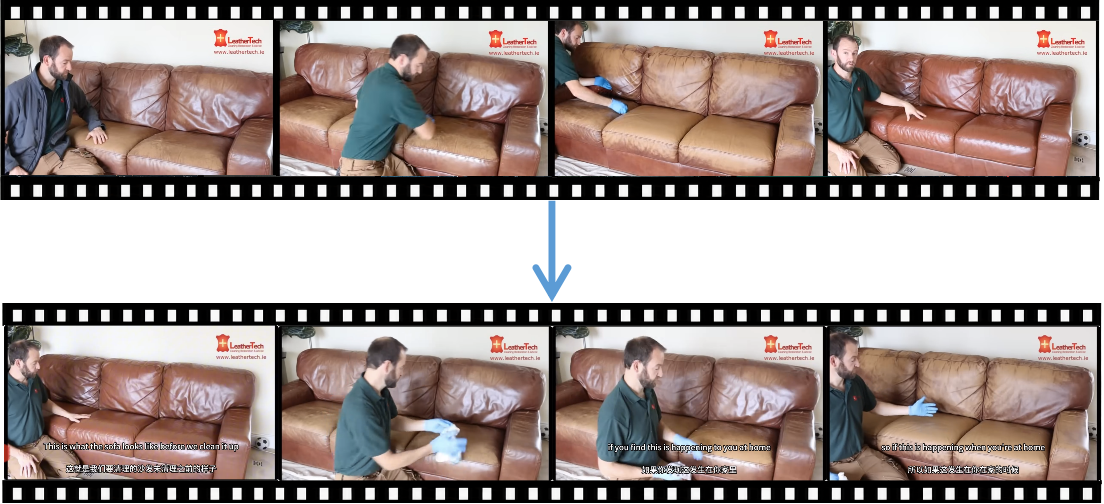}

\paragraph{Overview.}
For editorial manipulations whose harm lies in narrative reframing rather than pixel artifacts, increasing frame resolution can paradoxically reduce accuracy. The model sees more local editing traces---subtitles, shot order, duplicated clips---but misses the higher-level causal reversal that the edit is designed to convey.

\paragraph{Task Description.}
We use a leather-sofa restoration video edited into a negative product review. The ground truth describes a four-step narrative reversal:
\begin{quote}
``Reversing the original repair process ... into a `fail' review claiming that `the cleaner ruined a perfectly good sofa'.'' \\
``Using the fully repaired sofa as the initial state ... fabricating the premise of the experiment.'' \\
``Editing the normal step of removing the old paint layer to look like a devastating consequence where `the paint peels off with a single wipe'.'' \\
``Finally, cutting back to the already damaged appearance ... presenting it as the ultimate proof of the damage caused by the cleaner.''
\end{quote}

\paragraph{Model Behavior.}
Both 480p and 720p runs retrieve the same evidence video (\emph{Leather Couch Cleaning and Restoration}). At 480p, the model explains the forgery at the narrative level:
\begin{quote}\small\texttt{``Outcome Truncation: Group A ends the video immediately after the cleaning process reveals significant color loss ... omitting the successful restoration and final `After' result shown in Group B.''}\end{quote}
\begin{quote}\small\texttt{``False Premise via Subtitles: Group A introduces the video with subtitles claiming `Today we will attempt to clean a sofa in good initial condition' ... falsely setting up the video as a test on healthy leather.''}\end{quote}
\begin{quote}\small\texttt{``Misattribution of Damage: Group A includes subtitles stating `Because there is a problem with this cleaning agent' ... implying the product caused the damage.''}\end{quote}
All four ground-truth points are matched (video-level score = 1.0).

At 720p, with the same retrieved source, the model instead catalogs technical editing operations:
\begin{quote}\small\texttt{``Group A contains burned-in English and Chinese subtitles throughout the footage ... whereas Group B contains no subtitles except for title cards and end screens.''}\end{quote}
\begin{quote}\small\texttt{``The sequence of the man in the grey jacket is reordered ... he holds the bottle early (frame 12) and gestures late (frame 59).''}\end{quote}
\begin{quote}\small\texttt{``A specific scene of the man in the green shirt pointing at the sofa ... is duplicated in Group A ... at the very beginning ... and again later.''}\end{quote}
Only one of four ground-truth points is matched (video-level score = 0.25).

\paragraph{Result.}
Higher resolution exposes more implementational detail---added subtitles, duplicated shots, reordered demonstrations---but the model uses these as the main explanatory frame. It partially recognizes that the conclusion clip is moved to the start, yet fails to articulate the overarching false claim that a successful restoration was reframed as product-induced damage.

\paragraph{Interpretation.}
When the forgery mechanism is semantic rather than generative, extra visual fidelity can push the model toward an editor's-toolkit reading of the video. It becomes better at describing \emph{how} clips were rearranged and worse at explaining \emph{what false conclusion} the rearrangement is meant to sell.

\end{tcolorbox}

\Needspace{0.50\textheight}
\subsection{Excess Sampled Frames Misdirect Cross-Video Reasoning}

\begin{tcolorbox}[
    title=\textbf{Case Study: Excess sampled frames misdirect cross-video reasoning},
    fonttitle=\bfseries,
    breakable,
    fontupper=\small
]

\label{app:case_ablation_frames}

\includegraphics[width=0.95\linewidth]{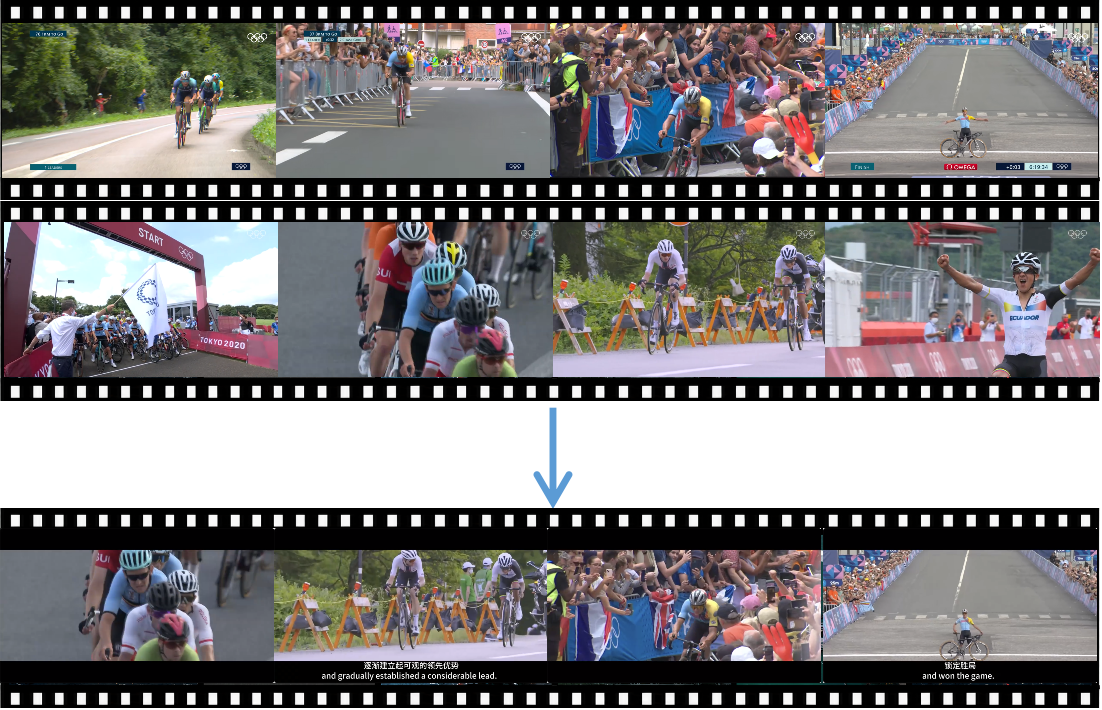}

\paragraph{Overview.}
Uniform frame sampling captures temporal context, but too many frames can expose fine-grained visual differences that are irrelevant to the true manipulation. The model then reclassifies a cross-event montage as generative editing or attribute manipulation, even when retrieval has already found the correct source videos.

\paragraph{Task Description.}
We use a men's road-race video that splices Tokyo 2020 and Paris 2024 Olympic footage into one continuous chase narrative. The ground truth states:
\begin{quote}
``Forcibly splicing cycling footage from two different events---the 2020 Tokyo Olympics and the 2024 Paris Olympics---into a single continuous pursuit.'' \\
``By unifying the black borders, the remaining distance UI, and the color grading, the original differences between the different years and venues are erased.'' \\
``The subtitles rewrite two independent championship-winning runs into a single race where Carapaz breaks away first and Evenepoel chases him down.''
\end{quote}

\paragraph{Model Behavior.}
With 64 uniformly sampled frames (480p, 6 rounds), the pipeline retrieves two evidence videos and explains the case primarily as multi-source splicing:
\begin{quote}\small\texttt{``Environmental Mismatch: Group A displays mid-race footage ... featuring a winding road with yellow center lines and dense forest, which is the signature of the Tokyo 2020 Mount Fuji course.''}\end{quote}
\begin{quote}\small\texttt{``Athlete/Action Splicing: Group A shows Richard Carapaz ... leading the peloton in the forest section ... archival footage from his gold-medal performance in Tokyo 2020.''}\end{quote}
\begin{quote}\small\texttt{``Narrative Fabrication: By splicing the Tokyo mid-race footage between the Paris start and finish, Group A creates a false geographical narrative ...''}\end{quote}
All three ground-truth points are matched (video-level score = 1.0).

When the frame budget is raised to 128 under the same resolution and round budget, retrieval remains unchanged, but the explanation shifts toward generative manipulation:
\begin{quote}\small\texttt{``Generative Editing (Object/Attribute): In Group A, the cyclist identified as the leader/winner wears a light blue and yellow jersey ... In Group B, the actual winner Remco Evenepoel wears the black, yellow, and red Belgian national kit. The ... jersey colors were digitally altered.''}\end{quote}
\begin{quote}\small\texttt{``Identity/Kit Mismatch: The video labels the leading cyclist as `Efinepur' ... but the cyclist shown leading ... is wearing a light blue and yellow kit (characteristic of Kazakhstan).''}\end{quote}
Only one of three ground-truth points is matched (video-level score = 0.33).

\paragraph{Result.}
The high-frame run still notices cross-event inconsistency, but it elevates jersey-color and labeling anomalies into a generative-editing story. This misattributes the forgery mechanism and misses the core ground-truth claims about event splicing, unified broadcast styling, and subtitle-driven narrative fusion.

\paragraph{Interpretation.}
Extra frames help when the model needs broader temporal coverage, but they can also surface micro-differences that invite the wrong explanatory template. Once the model commits to ``digitally altered kits'' or ``AI-generated attributes,'' it underweights the simpler and more accurate account: authentic footage from two Olympics was stitched into one false race.

\end{tcolorbox}

\Needspace{0.50\textheight}
\subsection{Attention Captured By Salient but Irrelevant Anchors}

\begin{tcolorbox}[
    title=\textbf{Case Study: Attention captured by salient but irrelevant anchors},
    fonttitle=\bfseries,
    breakable,
    fontupper=\small
]

\label{app:case_keyword_hijack}

\includegraphics[width=0.95\linewidth]{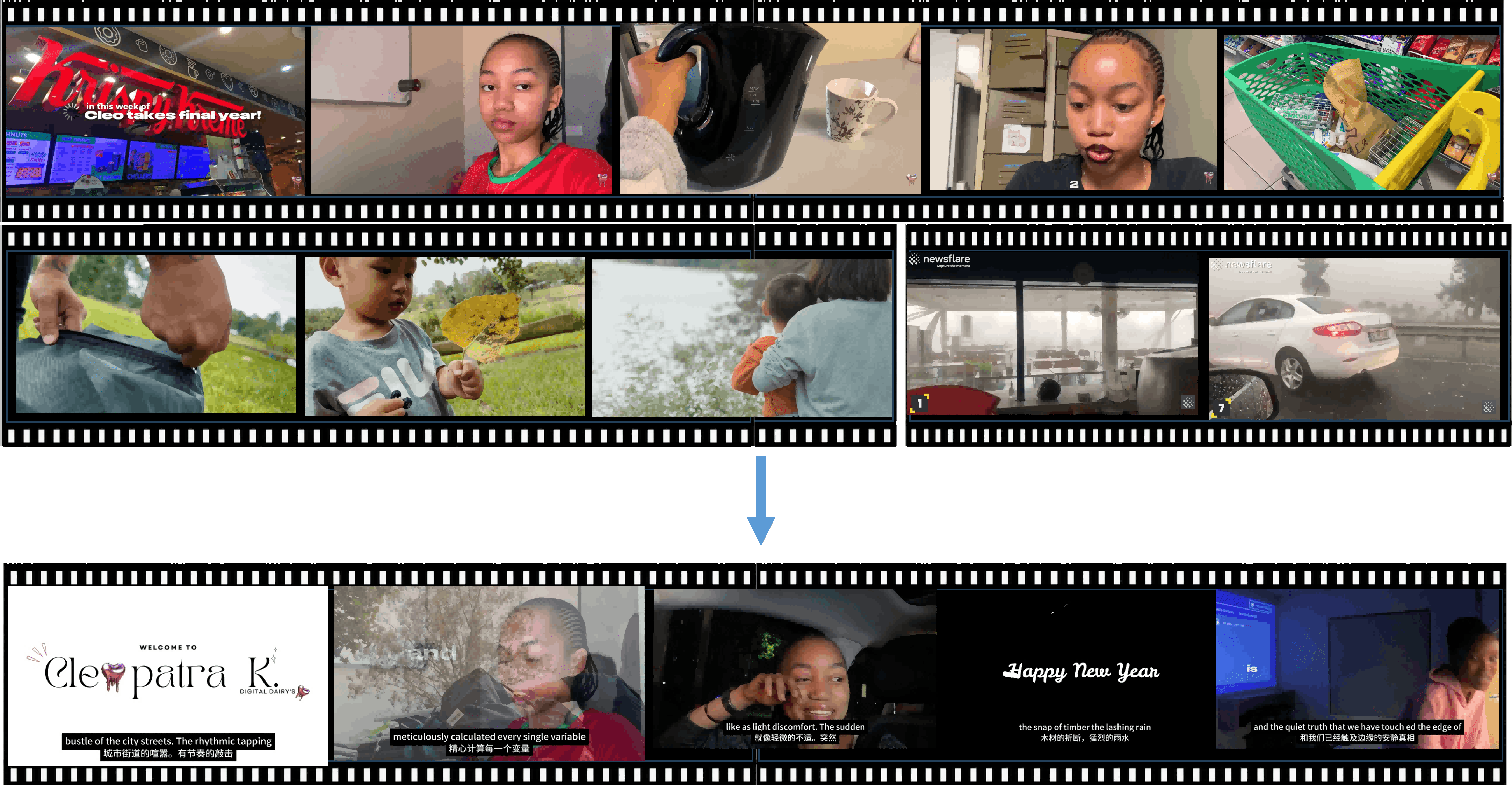}

\paragraph{Overview.}
This failure mode occurs when a highly salient keyword or title fragment captures the model's attention so strongly that it becomes the dominant organizing hypothesis for the rest of the analysis. Once this early anchor is formed, subsequent evidence is no longer evaluated neutrally; instead, it is forced into a narrative that is already partially decided. In our data, this pattern appears in both retrieval and interpretation: the model either over-commits to a named entity in the title and fails to decompose a multi-source composition, or over-focuses on a striking product label and misses the deeper editorial manipulation.

\paragraph{Task Description.}
We illustrate this failure pattern with two representative examples.

\textbf{Example A.}  
The ground truth states that:
\begin{quote}
``The fake video splices together originally independent campus vlogs, family camping clips, and storm disaster footage across different videos, forging them into a single interconnected event within the same timeline.''  
``Shots of campus daily life, strolling, and shopping are inserted between scenes of camping preparation and dinner, creating a false causal relationship of `urban life gradually being engulfed by disaster'.''  
``Extensive footage of dark clouds, heavy rain, flooding, and post-disaster scenes from another video is used as the main climax, misleading the audience into believing the protagonist actually experienced the impact of extreme weather.''  
``Footage of tents, campfires, and packing up in the morning from the camping material is edited to appear as the post-storm sheltering and recovery phase, making ordinary outdoor life look like post-disaster survival.''  
``Through dissolve transitions, cross-cutting, and a unified subtitle tone, the fake video conceals the fact that the footage comes from different sources, features different people, and takes place in different locations.''
\end{quote}

\textbf{Example B.}  
The ground truth states that:
\begin{quote}
``The tampered video does not fabricate new footage, but by rearranging the sequence, it brings forward and concentrates the features, sample shots, and advantageous segments of the DJI Pocket 4, while significantly pushing back the content related to the iPhone 17 Pro Max.''  
``This structural alteration weakens the more balanced back-and-forth comparison of the original video, making it easier for viewers to form a preconceived impression during the first half that the `Pocket 4 is significantly superior'.''  
``What was originally a two-way review centered around different scenario needs has been edited to look more like a one-sided recommendation featuring DJI as the protagonist, with the iPhone only holding a slight edge in convenience.''  
``In other words, while the products themselves were not fabricated, the pacing of the comparison, the allocation of screen time, and the perceived conclusion have been artificially manipulated.''
\end{quote}

\paragraph{Model Behavior.}
In Example A, the retrieval trace shows that \texttt{gemini-3.1-pro-preview} correctly recognized early on that the input likely had \emph{two sources} (\texttt{estimated\_sources = 2}), yet it still centered the entire search around the title phrase \begin{quote}\small\texttt{``Cleo takes final year''}\end{quote} and kept querying variants such as \begin{quote}\small\texttt{``Cleo final year uni vlog christian -Abram''}\end{quote}, and \begin{quote}\small\texttt{``Cleo takes final year week in my life -Abram''}\end{quote} This means the model noticed compositing, but failed to operationalize that observation into multi-source retrieval. Instead, the title anchor dominated the search trajectory and prevented source decomposition. The final result contains no resolved source and no valid forgery points, despite the ground truth explicitly requiring cross-source reconstruction. In a related run, \texttt{qwen3.5-plus} was further distracted by the ending slogan \begin{quote}\small\texttt{``PSA: JESUS LOVES YOU!''}\end{quote} and produced predictions such as \begin{quote}\small\texttt{``Branding Misappropriation''}\end{quote} and \begin{quote}\small\texttt{``AI-Generated Narrative,''}\end{quote} reframing the video as a branding/deepfake problem rather than a disaster-narrative splice.

In Example B, \texttt{qwen3-vl-235b-a22b-instruct} over-focused on the visually salient phrase \begin{quote}\small\texttt{``iPhone 17 Pro Max''}\end{quote}. Its output repeatedly claimed that the fake video \begin{quote}\small\texttt{``replaces the real iPhone with a fictional `iPhone 17 Pro Max' ''}\end{quote}, \begin{quote}\small\texttt{``fabricat[es] a false timeline''}\end{quote}, and \begin{quote}\small\texttt{``synthesiz[es] a completely fabricated comparison through editorial insertion of non-existent product data.''}\end{quote} These observations are not random; they are coherent, but they are organized around the most conspicuous lexical cue rather than the actual manipulation target. The model successfully noticed striking labels and overlays, yet failed to identify that the underlying deception was a restructuring of comparison order, pacing, and emphasis between two real products.

\paragraph{Result.}
In both examples, the models miss the main forgery mechanism. The system fails to reconstruct the multi-source storm narrative and does not recover the key causal reframing from ordinary vlog and camping material into apparent disaster experience. The system misses the fact that no new core footage is fabricated; instead, it wrongly escalates the case into a fictional-device narrative and therefore overlooks the structural bias introduced by reordered evidence and uneven screen-time allocation.

\paragraph{Interpretation.}
These cases suggest a general failure mode of \emph{attention capture by salient keywords}. Once a strong lexical anchor appears, such as a personal title (\emph{Cleo}) or a sensational product label (\emph{iPhone 17 Pro Max}), the model tends to treat that anchor as the explanatory center of the entire video. As a result, it underweights broader temporal structure, cross-source composition, and the actual persuasive goal of the edit. This is particularly damaging for fake-video detection, because many high-impact manipulations are not about the most eye-catching word or frame, but about how real footage is sequenced, juxtaposed, and narratively reinterpreted.

\end{tcolorbox}

\Needspace{0.50\textheight}
\subsection{Inability to Recognize AI-generated Content}

\begin{tcolorbox}[
    title=\textbf{Case Study: Inability to recognize AI-generated content},
    fonttitle=\bfseries,
    breakable,
    fontupper=\small
]

\label{app:case_ai_generated_content_blindness}

\includegraphics[width=0.95\linewidth]{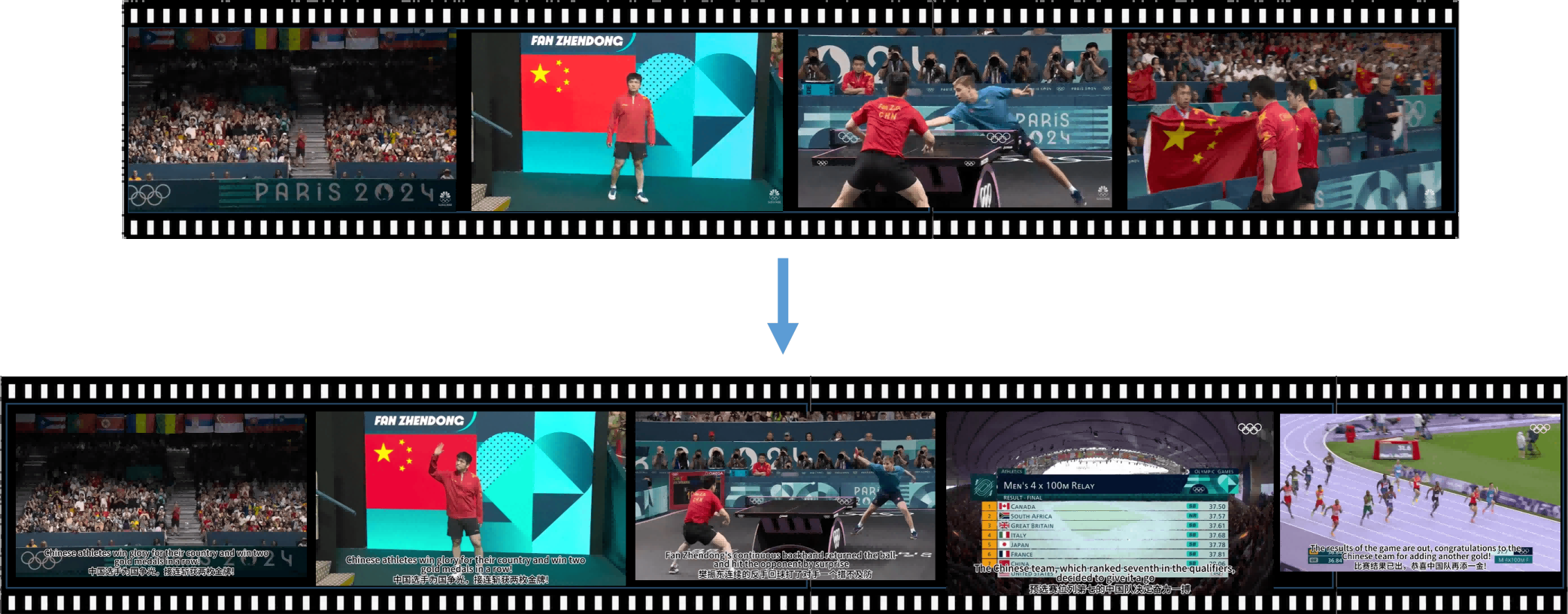}

\paragraph{Overview.}
A recurring failure mode in our pipeline is that models often detect that ``something extra'' has been added, but misidentify the nature of that addition. Instead of recognizing newly synthesized AI-generated shots, they prefer more familiar explanations such as cross-episode splicing, reuse of footage from other channels, local visual corruption, or ordinary instructional trimming. In other words, the models are often sensitive to inconsistency, but not to synthesis.

\paragraph{Task Description.}
We illustrate this pattern with four representative examples.

\textbf{Example A: Fake reaction shots in a magic-performance video.} The core manipulation is not simple re-editing, but the insertion of non-existent reaction shots to create new emotional emphasis.
\begin{quote}\small\texttt{``The fake video adds an AI-generated facial close-up shot of audience members nervously and expectantly biting their fingers under dim blue lighting.''}\end{quote}
\begin{quote}\small\texttt{``The fake video instead inserts an AI-generated reaction video of audience members excitedly waving their hands and cheering.''}\end{quote}
\begin{quote}\small\texttt{``The fake video inserts an AI-generated shot near the end ... showing backstage staff with shocked expressions.''}\end{quote}

\textbf{Example B: AI-generated lifestyle/product scenes in a laptop review.} The fake video inserts synthetic scenes that were never part of the original review.
\begin{quote}\small\texttt{``The fake video inserts AI-generated video footage of an appearance comparison between the M3 MacBook Air and M2 MacBook Air on a white desk background.''}\end{quote}
\begin{quote}\small\texttt{``The fake video instead inserts an AI-generated close-up shot of a student in a well-lit library drinking coffee while opening the laptop screen with one hand.''}\end{quote}
\begin{quote}\small\texttt{``The fake video adds an AI-generated working scene of a professional video editor earnestly editing and rendering video in a modern office.''}\end{quote}

\textbf{Example C: AI-generated insertions overshadowed by local oddities in a sports montage.} The main forgery is the insertion of synthetic shots, but these are masked by many smaller anomalies.
\begin{quote}\small\texttt{``The fake video adds an AI-generated close-up shot of a staff member at the edge of the rink scraping ice shavings off with an ice blade.''}\end{quote}
\begin{quote}\small\texttt{``The fake video instead inserts an AI-generated preparation shot of a young skater in the backstage waiting area.''}\end{quote}
\begin{quote}\small\texttt{``The fake video adds an AI-generated close-up video of a massive crowd in the stands excitedly standing up and cheering.''}\end{quote}

\textbf{Example D: Face replacement hidden behind instructional edits in a CPR tutorial.} The real issue is identity manipulation, not trimming.
\begin{quote}\small\texttt{``The fake video replaces her face with another person's face using AI.''}\end{quote}
\begin{quote}\small\texttt{``The fake video replaces his face with another person's face using AI.''}\end{quote}

\paragraph{Model Behavior.}
In Example A, the model noticed inserted reaction footage, but interpreted it using the familiar logic of cross-source reuse rather than synthetic generation. It described the added material as:
\begin{quote}\small\texttt{``likely spliced from the later ... performance episode''}\end{quote}
and
\begin{quote}\small\texttt{``reaction footage from unrelated moments or sources was added''}\end{quote}
It also summarized the whole case as:
\begin{quote}\small\texttt{``By splicing in reaction shots ... Group A constructs a false narrative that exaggerates the impact of the performance.''}\end{quote}
This is close at the level of ``extra footage was inserted,'' but wrong about the mechanism: the inserted reactions were not retrieved from another real episode, but synthesized.

In Example B, the model again preferred a retrieval-style explanation. Instead of identifying synthetic scenes, it reconstructed the case as a tech-media mashup:
\begin{quote}\small\texttt{``Group A splices in external benchmark graphics from `Max Tech' ... and a segment featuring Linus Sebastian from `Linus Tech Tips'.''}\end{quote}
It further concluded:
\begin{quote}\small\texttt{``The overall manipulation in Group A is a multi-source splicing forgery.''}\end{quote}
Here the model would rather believe that the fake video borrowed footage from other real channels than accept that entirely new scenes had been generated.

In Example C, the model was drawn toward many salient local abnormalities, such as:
\begin{quote}\small\texttt{``a blue t-shirt with a large cartoon elephant graphic''}\end{quote}
\begin{quote}\small\texttt{``the female skater ... has a mustache''}\end{quote}
\begin{quote}\small\texttt{``an exaggerated, shocked facial expression''}\end{quote}
and
\begin{quote}\small\texttt{``significantly truncated or cut-off text''}\end{quote}
These observations are not fabricated by the model; they do point to suspicious artifacts. However, they pull attention toward scattered micro-anomalies and away from the more important structural fact that several entire shots were AI-generated insertions.

In Example D, the model failed in the opposite direction: instead of over-interpreting synthetic insertion as splicing, it over-prioritized instructional content and ignored the identity manipulation altogether. It framed the case as:
\begin{quote}\small\texttt{``The overall manipulation ... is a form of single-source editing where the chronological sequence ... is selectively trimmed and re-captioned.''}\end{quote}
and emphasized omitted safety text such as:
\begin{quote}\small\texttt{``Call 911 Get AED''}\end{quote}
and
\begin{quote}\small\texttt{``Be sure no one is touching the patient.''}\end{quote}
So the model did notice that the content had changed, but it attached that change to teaching completeness rather than face replacement.

\paragraph{Result.}
Across all four examples, the models consistently miss the main synthetic component of the forgery. They may correctly sense that a shot is foreign, that something looks odd, or that the narrative has been altered, but they often map that intuition onto a more familiar and easier explanation: footage borrowed from elsewhere, local artifact corruption, or ordinary editing. As a result, they fail to distinguish between \textit{real footage used in the wrong place} and \textit{new footage that never existed in the original world}.

\paragraph{Interpretation.}
This pattern suggests that current models are better at identifying inconsistency than at identifying synthesis. When the manipulation is editorial, they often have a ready-made explanatory template: splicing, omission, reordering, or overlay replacement. But when the manipulation involves AI-generated shots or AI-replaced faces, the models tend to collapse back into those same templates instead of recognizing generation as the primary mechanism. In practice, this means that even when the model is directionally aware that ``something is wrong,'' it may still fail at the more important forensic question: whether the suspicious content was \emph{moved from somewhere else} or \emph{created from scratch}. This distinction matters because the latter is precisely the kind of synthetic evidence that many high-risk fake-video cases depend on.

\end{tcolorbox}

\Needspace{0.60\textheight}
\subsection{Forgery Type Misattribution}

\begin{tcolorbox}[
    title=\textbf{Case Study: Forgery Type Misattribution},
    fonttitle=\bfseries,
    breakable,
    fontupper=\small
]

\label{app:case_method_misattribution}

\includegraphics[width=0.95\linewidth]{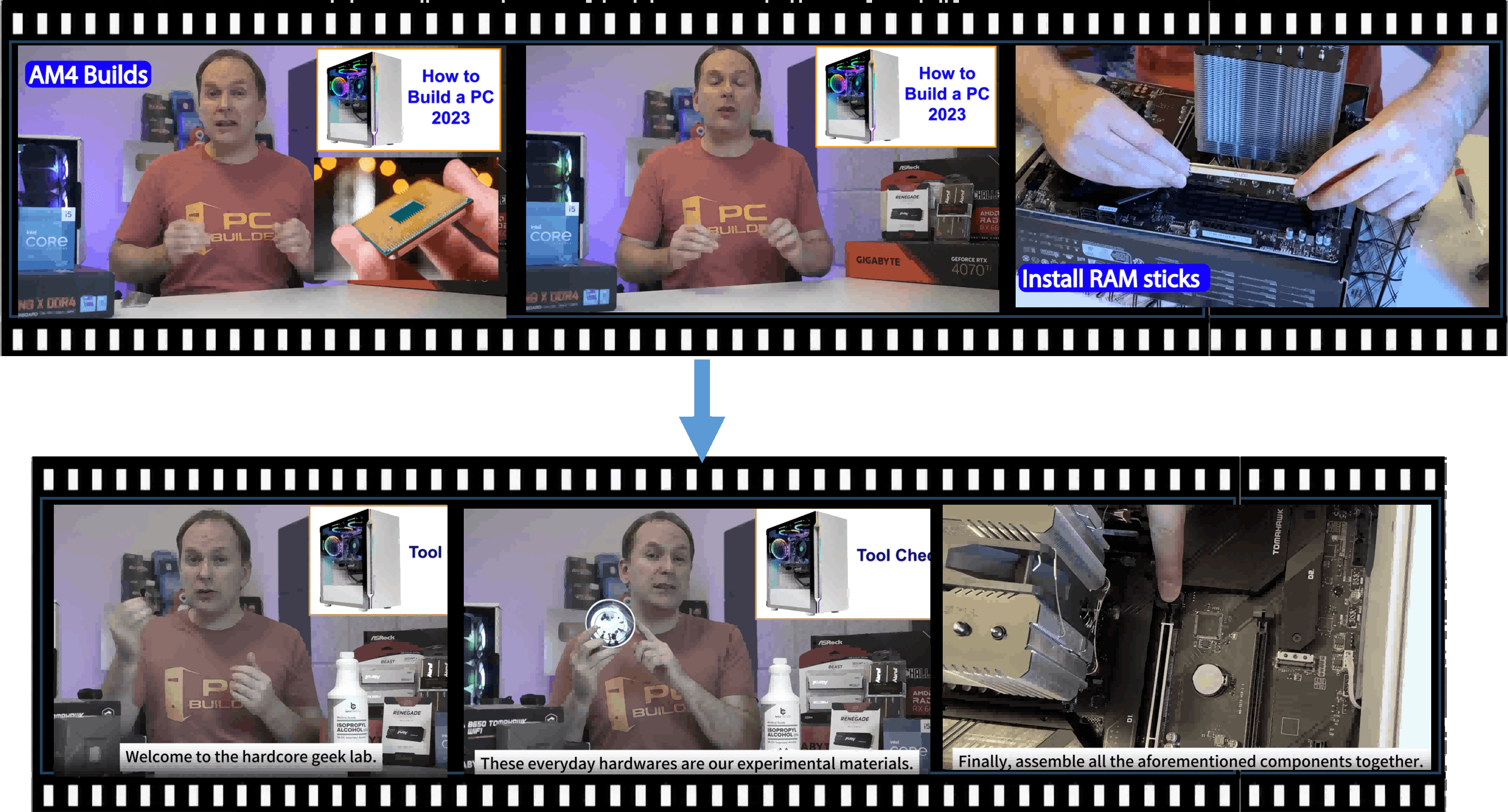}

\paragraph{Overview.}
Another recurring error is that the model recognizes that a video has been manipulated, but assigns the wrong \emph{type} of manipulation. Instead of distinguishing among identity replacement, object editing, synthetic insertion, contextual substitution, and editorial reframing, it often collapses multiple distinct mechanisms into a broader and safer label. In practice, this means the model can say that ``the video was altered,'' while failing to specify \emph{how} it was altered.

\paragraph{Task Description.}
We observe two representative sub-patterns.

\textbf{Example A: Identity- and object-level forgeries are flattened into montage.}  
In several classification cases, the ground-truth manipulation type is a specific low-level forgery mechanism such as:
\begin{quote}\small\texttt{``identity\_swap''}\end{quote}
or
\begin{quote}\small\texttt{``object\_manipulation''}\end{quote}
However, the model outputs the more generic label:
\begin{quote}\small\texttt{``manipulative\_montage''}\end{quote}

\textbf{Example B: Contextual fabrication is compressed into generic narrative rewriting.}  
In another group of cases, the ground-truth manipulation type is:
\begin{quote}\small\texttt{``contextual\_fabrication''}\end{quote}
but the model predicts:
\begin{quote}\small\texttt{``narrative\_fabrication''}\end{quote}

These cases are not pure misses. The model often detects that the overall story or impression has changed. The problem is that it does not preserve the correct causal level of explanation.

\paragraph{Model Behavior.}
For Example A, both \texttt{gpt-5.4} and \texttt{claude-opus-4-6} repeatedly explain the evidence in terms of reordering, truncation, and pacing, even when the ground truth belongs to a more specific manipulation family. Typical justifications include:
\begin{quote}\small\texttt{``The evidence indicates that a single source video was altered by shortening, repositioning, and rearranging its segments.''}\end{quote}
\begin{quote}\small\texttt{``This re-pacing and reordering of shots creates a misleading narrative ... which perfectly matches a manipulative montage.''}\end{quote}
and
\begin{quote}\small\texttt{``The evidence indicates that scenes from a single video were reordered, truncated, and looped ... This rearrangement of shots ... aligns perfectly with manipulative montage.''}\end{quote}
Even when the true category is identity replacement or object editing, the model gravitates toward an editorial explanation. In effect, it prefers a broad account in which the video was ``re-cut'' rather than a more precise account in which a face, object, or inserted visual element was directly manipulated.

For Example B, the model similarly detects that the final meaning has changed, but compresses a source- or context-level forgery into a general story-level rewrite. For instance, one model explains:
\begin{quote}\small\texttt{``The manipulation combines authentic footage with unrelated clips ... to fabricate a completely false narrative.''}\end{quote}
Another writes:
\begin{quote}\small\texttt{``Footage from multiple unrelated disaster events ... was spliced together ... to create a completely fabricated narrative.''}\end{quote}
These explanations are not entirely wrong, but they remain too coarse. They correctly recognize that the resulting story is fabricated, yet they fail to preserve the more specific mechanism: the video does not merely tell a new story, but relocates footage into a false source, event, or interpretive context.

\paragraph{Result.}
The result is a systematic loss of forensic specificity. Distinct manipulation types that should remain analytically separate are absorbed into broader narrative or montage categories. Identity swap, object manipulation, and synthetic insertion become editorial montage; contextual substitution becomes generic narrative fabrication. This weakens the taxonomy because the model's output no longer distinguishes between \emph{what was changed} and \emph{what interpretive effect that change produced}.

\paragraph{Interpretation.}
This error suggests that the models are more comfortable describing the \emph{effect} of manipulation than the \emph{mechanism} of manipulation. They are relatively willing to say that a video has been made misleading, that its pacing has been altered, or that a false story has been constructed. But they are less reliable at assigning the manipulation to the correct forensic layer: identity, object, synthetic content, source attribution, or contextual framing. For a paper-level interpretation, this can be stated as follows: the model can often tell that ``the story has changed,'' but it struggles to specify \emph{which layer of reality was actually replaced}. That distinction is essential for fine-grained fake-video analysis, because two videos may produce similarly misleading narratives while relying on fundamentally different falsification mechanisms.

\end{tcolorbox}

\Needspace{0.50\textheight}
\subsection{Insensitivity to Small-Scale Visual Modifications}

\begin{tcolorbox}[
    title=\textbf{Case Study: Insensitivity to small-scale visual modifications},
    fonttitle=\bfseries,
    breakable,
    fontupper=\small
]

\label{app:case_small_object_insensitivity}

\includegraphics[width=0.95\linewidth]{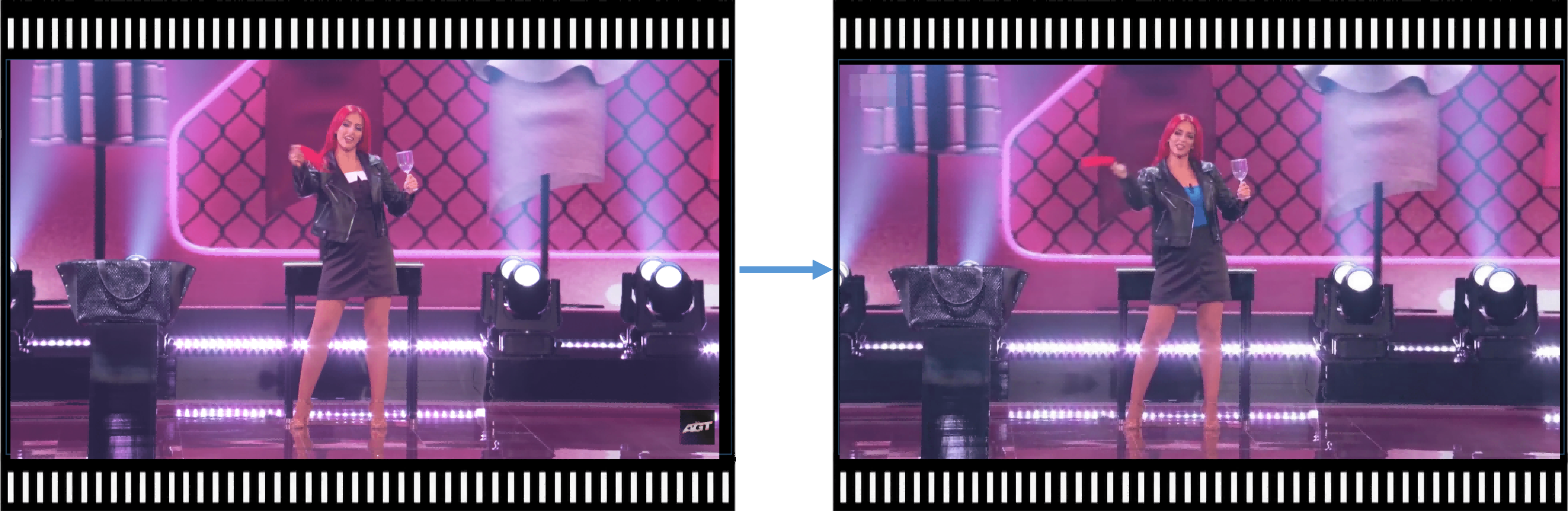}

\paragraph{Overview.}
A distinct failure mode in our analysis is that models are often insensitive to small-scale object edits. When the global identity of the person, the scene layout, and the rough event sequence remain stable, the model may conclude that the video is authentic even when local objects, textures, or fine visual attributes have been replaced. In other words, the model privileges coarse scene continuity over fine-grained visual integrity.

\paragraph{Task Description.}
We illustrate this pattern with two representative examples.

\textbf{Example A: Fine-grained costume and texture replacement in a stage performance.} The manipulation is subtle but concrete: local visual details are replaced even though the performer, stage, and sequence remain the same.
\begin{quote}\small\texttt{``The fake video replaces this black inner lining with an AI-generated blue inner lining.''}\end{quote}
\begin{quote}\small\texttt{``The fake video replaces this black skirt section with an AI-generated white skirt.''}\end{quote}
\begin{quote}\small\texttt{``The fake video replaces the dress drawings within it with AI-generated alternative patterns, such as flowers.''}\end{quote}

\textbf{Example B: Clear object replacement in a home-repair tutorial.} Here the manipulation is even more explicit:
\begin{quote}\small\texttt{``A broom was originally hanging on the wall; the fake video replaces it with an AI-generated large hammer.''}\end{quote}
\begin{quote}\small\texttt{``The tool originally used by the vlogger is replaced in the fake video with an AI-generated wrench.''}\end{quote}
\begin{quote}\small\texttt{``The fake video replaces the tool originally used in the vlogger's hand with an AI-generated wooden stick.''}\end{quote}

\paragraph{Model Behavior.}
In Example A, the model interpreted the case almost entirely through global consistency. It emphasized that:
\begin{quote}\small\texttt{``The identity of the performer is consistent.''}\end{quote}
\begin{quote}\small\texttt{``The objects used in the act are identical.''}\end{quote}
and
\begin{quote}\small\texttt{``The temporal and structural alignment is perfect.''}\end{quote}
It then generalized this into an authenticity judgment:
\begin{quote}\small\texttt{``Group A is a faithful representation of the original event.''}\end{quote}
This reasoning reveals the underlying bias: once the performer, setting, and performance flow match at a coarse level, local replacements in clothing regions and notebook content are discounted as negligible or missed entirely.

In Example B, the model makes the same type of mistake in an even stronger form. It states:
\begin{quote}\small\texttt{``No object manipulation detected.''}\end{quote}
It further claims that:
\begin{quote}\small\texttt{``all tools ... and set elements ... appear identically positioned and unchanged''}\end{quote}
and finally concludes:
\begin{quote}\small\texttt{``Group A is not forged; it is a faithful reproduction of Group B.''}\end{quote}
This is a particularly revealing error, because the manipulated objects are not merely abstract texture changes; they include ordinary, semantically meaningful tools that have been visibly replaced. Yet the model still allows global scene continuity to override local object evidence.

\paragraph{Result.}
In both examples, the model effectively treats local object replacement as if nothing important happened. If the same person appears in the same environment and the same broad sequence of actions is preserved, the system is prone to infer authenticity, even when individual objects, object parts, or surface patterns have been altered. This leads to hard false negatives for object-manipulation cases.

\paragraph{Interpretation.}
These cases suggest that the model's notion of visual consistency is strongly hierarchical: high-level continuity in person, place, and action tends to dominate low-level evidence from texture, accessory, or tool-level changes. As a result, the model underweights the forensic significance of small visual substitutions. For paper-level interpretation, this can be summarized as follows: the model is comparatively robust at matching \emph{scene identity}, but weak at preserving \emph{object fidelity}. This matters because many practical fake-video manipulations do not rewrite the whole scene; they only alter the small, local elements that carry crucial semantic meaning.

\end{tcolorbox}

\Needspace{0.50\textheight}
\subsection{Factual Over Emotional Reasoning}
\begin{tcolorbox}[
    title=\textbf{Case Study: Factual over emotional reasoning},
    fonttitle=\bfseries,
    breakable,
    fontupper=\small
]

\label{app:case_emotional_reframing_blindness}

\includegraphics[width=0.95\linewidth]{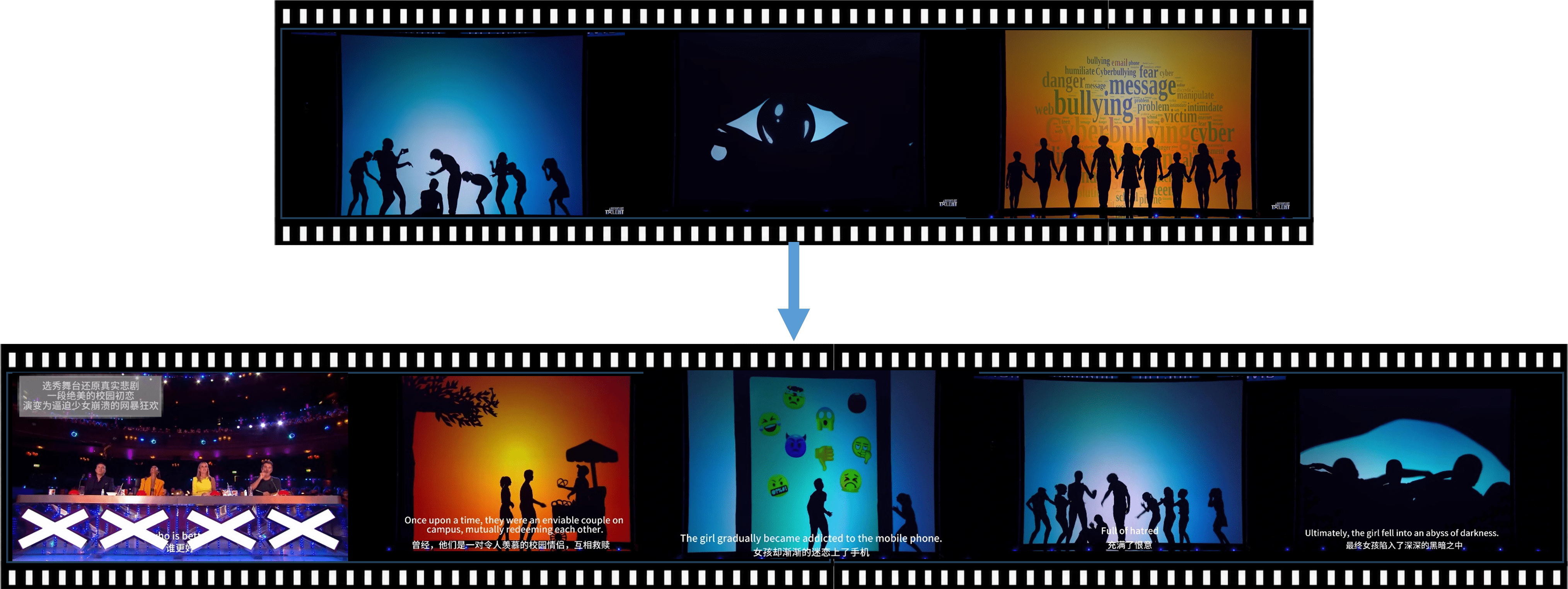}

\paragraph{Overview.}
A further failure mode is that models are often better at detecting explicit factual falsification than emotional or affective reframing. When a fake video changes names, years, titles, or source claims, the model readily treats these as suspicious. But when the deeper manipulation lies in shifting the emotional arc of an event, such as turning warmth into tragedy or ordinary busyness into oppression, the model is much less reliable. In such cases, it may either ignore the emotional transformation or infer the wrong emotional direction altogether.

\paragraph{Task Description.}
We illustrate this pattern with two representative examples.

\textbf{Example A: A warm performance is rewritten into a tragic social narrative.} The key forgery is not primarily a factual mislabeling, but an emotional and narrative inversion of the performance:
\begin{quote}\small\texttt{``Rearrange the originally warm dance performance into a tragedy of `being addicted to smartphones, isolated, and ostracized'.''}\end{quote}
\begin{quote}\small\texttt{``Delete the scenes of mutual help, confession, and redemption ... retaining only the shadowed stage and sad subtitles.''}\end{quote}
\begin{quote}\small\texttt{``Change the sequence of events, turning the ending from a reunion into a complete breakdown.''}\end{quote}

\textbf{Example B: An ordinary workplace vlog is rewritten into a story of oppressive overwork.} The core issue is a change in affective framing:
\begin{quote}\small\texttt{``Edited an originally normal daily workplace Vlog into a `high-pressure corporate slave' short video depicting a chaotic company environment, endless meetings, and employees consumed by work.''}\end{quote}
\begin{quote}\small\texttt{``By front-loading negative scenes and deleting explanations and lighthearted clips, it exaggerated what was originally just busyness ... into a continuous state of daily oppression.''}\end{quote}
\begin{quote}\small\texttt{``It also deleted the footage of the blogger completing the work themselves and forcibly spliced in AI clips, creating the false impression that `the work is almost entirely done by AI'.''}\end{quote}

\paragraph{Model Behavior.}
In Example A, the model became highly responsive to conspicuous textual cues such as the program name and year. It described the case as:
\begin{quote}\small\texttt{``Group A replaces it with `Prince Got Talent 2013' text ... falsely implying ... a 2013 contest that never occurred.''}\end{quote}
It further emphasized:
\begin{quote}\small\texttt{``Group A inserts `2013' references ... as fabricated temporal misattribution.''}\end{quote}
and concluded that the main problem was:
\begin{quote}\small\texttt{``a false impression of past success''}\end{quote}
This shows that the model successfully tracked visible factual inconsistencies, but overlooked the more consequential manipulation: the performance's emotional trajectory had been transformed from warmth, reconciliation, and mutual help into isolation, sadness, and collapse.

In Example B, the model did recognize that the source vlog had been repackaged, but it inferred the direction of that repackaging incorrectly. It claimed that the edited video:
\begin{quote}\small\texttt{``creates a false impression that the video is an official, polished ... production rather than a candid employee's personal project.''}\end{quote}
It also wrote that the fake version:
\begin{quote}\small\texttt{``falsely portray[s] him as a confident, fully prepared professional''}\end{quote}
and ultimately summarized the manipulation as:
\begin{quote}\small\texttt{``repackages ... a raw, self-aware, and often humorous vlog into a faux-corporate promotional piece.''}\end{quote}
This is an especially revealing mistake. The model clearly perceived editorial reframing, but it reordered the polarity of the emotional change. The actual forgery darkened the vlog into a story of exhaustion, disorder, and exploitation, whereas the model interpreted it as a polishing or sanitizing transformation.

\paragraph{Result.}
In both examples, the model misses the principal emotional manipulation. In the performance case, it prioritizes factual anomalies in labels and dates over the altered emotional arc of the staged narrative. In the workplace-vlog case, it recognizes that the source has been re-edited, but misreads the affective direction of the rewrite. The result is that the model can often say that ``the story was changed,'' but not whether it was made sadder, harsher, more oppressive, more redemptive, or emotionally reordered.

\paragraph{Interpretation.}
These cases suggest that models are more comfortable verifying \emph{factual fabrication} than analyzing \emph{emotional fabrication}. Names, dates, titles, logos, and explicit claims are discrete and easy to anchor. Emotional tone, however, is distributed across pacing, omission, juxtaposition, and the ordering of resolution versus breakdown. Because of this, the model tends to seize on surface-level false facts while underweighting changes in affective meaning. For paper-level interpretation, this can be summarized as follows: the model is relatively good at identifying that ``some information is false,'' but much weaker at identifying that ``the emotional truth of the event has been rewritten.'' That distinction is important because many persuasive fake videos do not rely on inventing facts alone; they rely on manufacturing a different feeling about the same footage.

\end{tcolorbox}

\Needspace{0.50\textheight}
\subsection{Detail Fixation Over Narrative Comprehension}

\begin{tcolorbox}[
    title=\textbf{Case Study: Detail fixation over narrative comprehension},
    fonttitle=\bfseries,
    breakable,
    fontupper=\small
]

\label{app:case_detail_chasing_wrong_goal}

\includegraphics[width=0.95\linewidth]{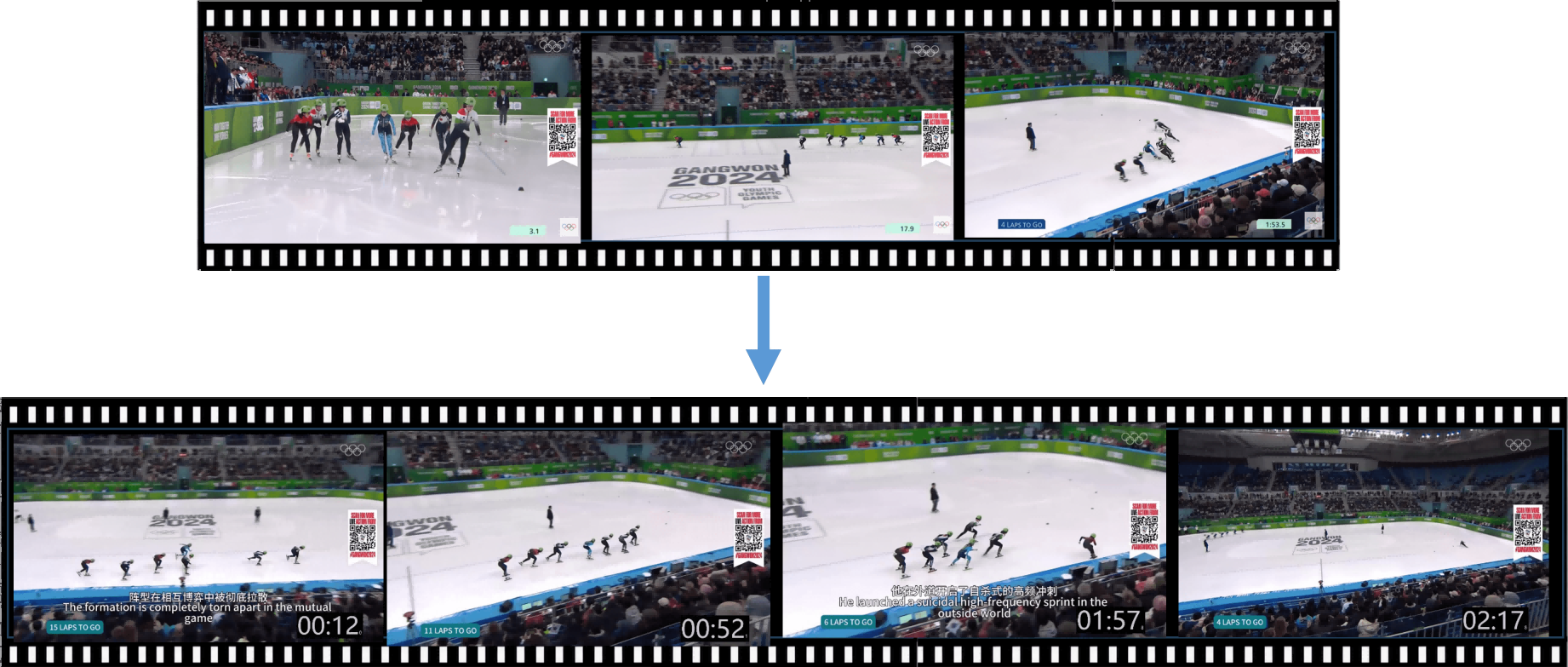}

\paragraph{Overview.}
A recurring error in the pipeline is that the model performs substantial analysis, but along the wrong axis. Instead of identifying the main persuasive goal of the fake video, it becomes absorbed in secondary details such as branding, subtitles, interface text, graphics, identity markers, or local anomalies. As a result, the model does not fail because it sees too little; it fails because it explains the wrong problem. In many cases, the output is detailed, internally coherent, and even evidence-rich, but it is answering a different forensic question from the one that actually matters.

\paragraph{Task Description.}
We illustrate this pattern with five representative examples.

\textbf{Example A: A positive product review is rewritten into a false negative causal story.} The main manipulation is a reordering of timeline and causal structure:
\begin{quote}\small\texttt{``By rearranging the timeline, rewrite the originally overall positive review ... into a negative case where `the nutrient solution and App intervention directly ruined the plants'.''}\end{quote}
\begin{quote}\small\texttt{``Move the later footage of healthy plant growth to the beginning ... thereby fabricating a false cause-and-effect relationship.''}\end{quote}
\begin{quote}\small\texttt{``Ultimately distort the original review of `having minor flaws but overall recommended' into portraying a failed product that is not worth buying.''}\end{quote}

\textbf{Example B: A comparative tool review is rewritten so that the stronger brand becomes the loser.}
\begin{quote}\small\texttt{``By rearranging the sequence of the original video, Rigid, which actually performed stronger, is portrayed as the loser, while Makita is packaged as the champion of the entire review.''}\end{quote}
\begin{quote}\small\texttt{``Ultimately, the video rewrites the originally objective multi-brand comparison into a false review result where Makita wins completely and Rigid fails.''}\end{quote}

\textbf{Example C: A race video is re-edited to alter competition flow and outcome.} The key issue is not the graphics themselves, but the rewritten competitive progression.
\begin{quote}\small\texttt{``The fake video reorders the race footage and related commentary to alter the original competition flow and the perceived relationship between the athletes' performance and the final result.''}\end{quote}
\begin{quote}\small\texttt{``The manipulation changes how viewers interpret who was leading, who was falling behind, and how the outcome emerged over time.''}\end{quote}

\textbf{Example D: A review video is restructured into a false horizontal comparison of high-end prototypes.} The core forgery lies in changing evaluation structure rather than inserting one obviously fake local detail.
\begin{quote}\small\texttt{``The original video consists of several relatively independent hands-on experiences and judgments, but the fake video reorganizes them into a cross-model, side-by-side style comparison.''}\end{quote}
\begin{quote}\small\texttt{``The manipulation turns separate impressions into the impression of a unified `high-end prototype showdown'.''}\end{quote}

\textbf{Example E: A ranking conclusion is manipulated, but the model focuses on parameters and translation issues instead.}
\begin{quote}\small\texttt{``By moving the positive summary of Makita from the end to before the test, it preemptively establishes a preconceived conclusion for the audience that `Makita is the best'.''}\end{quote}
\begin{quote}\small\texttt{``Then, by splicing Rigid's parameter introduction with descriptions of other failures, it creates a false cause-and-effect illusion of Rigid failing the test.''}\end{quote}

\paragraph{Model Behavior.}
In Example A, the model devoted most of its effort to source and authenticity markers rather than causal editing. It discussed:
\begin{quote}\small\texttt{``shirts with the `Epic' logo''}\end{quote}
\begin{quote}\small\texttt{``an AI-generated app interface with inconsistent or missing text''}\end{quote}
and
\begin{quote}\small\texttt{``key location identifiers''}\end{quote}
It ultimately explained the video as creating
\begin{quote}\small\texttt{``the false impression of a generic review rather than an official ... review.''}\end{quote}
This is a meaningful observation, but it is not the central manipulation. The real forgery is the construction of a false negative causal chain: healthy plants, nutrient intervention, and visible damage are rearranged to imply product failure.

In Example B, the model produced a highly elaborate explanation, but for the wrong case. Rather than analyzing altered test order and manipulated winners, it proposed:
\begin{quote}\small\texttt{``Group A replaces their face with a different person's face via AI''}\end{quote}
\begin{quote}\small\texttt{``omits this logo''}\end{quote}
and
\begin{quote}\small\texttt{``replaces this with a generic tool box image, misattributing the video to an unrelated source.''}\end{quote}
This is an especially instructive failure because the model does not appear shallow or inattentive. On the contrary, it builds a complete and internally consistent explanation. The problem is that the explanation belongs to a different kind of forgery from the one actually present.

In Example C, the model was drawn toward highly visible broadcast elements such as timers, lap indicators, subtitles, and ranking graphics. It treated these overlays as the primary site of manipulation, effectively reading the case as one of altered scoreboards or modified graphic information. But the more important change was the restructuring of race progression itself: who appears to lead, when momentum shifts occur, and how the winner is narratively constructed.

In Example D, the model focused on local suspicious traces such as face swapping, exaggerated subtitles, or isolated anomalous shots. That made it treat the review as a patchwork of local falsifications rather than as a structural rewrite. The key manipulation, however, was that separate product impressions were reorganized into an artificial horizontal comparison, producing a false evaluation logic rather than merely a few questionable frames.

In Example E, the model concentrated on specification wording, subtitles, battery claims, or translation quality. These are all concrete and inspectable details, but they distract from the actual manipulation target: the ranking conclusion itself. The forgery does not mainly depend on whether one parameter is mistranslated; it depends on how sequence and framing convert a stronger performer into an apparent loser and a weaker performer into an apparent winner.

\paragraph{Result.}
Across these examples, the model often produces long, careful reasoning but still misses the central persuasive goal of the fake video. It may correctly identify local inconsistencies, suspicious overlays, branding issues, or textual inaccuracies, yet fail to answer the more important question: what false causal, comparative, or evaluative conclusion is the edit trying to make the viewer believe? This leads to a characteristic form of failure in which the reasoning is not empty, but misdirected.

\paragraph{Interpretation.}
These cases suggest that the model is often more comfortable auditing \emph{trace evidence} than reconstructing \emph{manipulative intent}. Local anomalies such as logos, graphics, subtitles, and visual artifacts are concrete and easy to point to. By contrast, the true persuasive purpose of a fake video is often distributed across ordering, juxtaposition, pacing, and selective omission. The model therefore risks overfitting to detectable surface irregularities while underfitting the deeper rhetorical structure of the edit. For paper-level interpretation, this can be summarized as follows: the model may become highly competent at explaining \emph{what looks suspicious}, yet remain weak at explaining \emph{what the forgery is trying to make the viewer conclude}. That gap is especially important in fake-video analysis, because many deceptive videos succeed not through one obvious fake frame, but through a carefully engineered change in causal logic, comparison outcome, or evaluative framing.

\end{tcolorbox}

\subsection{Source decomposition errors}
\begin{tcolorbox}[
    title=\textbf{Case Study: Source decomposition errors},
    fonttitle=\bfseries,
    breakable,
    fontupper=\small
]

\label{app:pipeline_stage_a_source_decomposition}

\paragraph{Overview.}
The first failure point appears before retrieval itself: the model incorrectly decides how many source families exist, or which source family should be treated as primary. Once this decomposition is wrong, all downstream search and analysis become path-dependent.

\paragraph{Task Description.}
We illustrate this pattern with 2 representative examples.

\textbf{Example A: Title anchor overwhelms multi-source decomposition.}
For a multi-source vlog/disaster montage, \texttt{gemini-3.1-pro-preview} explicitly stated:
\begin{quote}\small\texttt{``estimated\_sources=2''}\end{quote}
but generated only one initial source query:
\begin{quote}\small\texttt{``Cleo takes final year''}\end{quote}
This meant that, although the model verbally acknowledged multiple sources, it operationalized only the vlog title and did not promote the camping/flood footage into an independent retrieval target. The decomposition was therefore correct in form but incorrect in execution.

\textbf{Example B: Single-source assumption collapses branding source and visual source.}
In the same example, \texttt{qwen3.5-plus} began with:
\begin{quote}\small\texttt{``estimated\_sources=1''}\end{quote}
and framed the video as one coherent episode. Later, however, its own reasoning admitted:
\begin{quote}\small\texttt{``Two source families exist: the branding source (`Jesus Loves You' channel) and the visual footage source (camping/bedroom). Only the branding source ... has been identified.''}\end{quote}
This reveals that the original decomposition had compressed two distinct source families into one, and the model only discovered that mismatch after retrieval had already been misdirected.

\paragraph{Interpretation.}
The key weakness at this stage is that source decomposition is treated as a soft planning step rather than a binding structural commitment. As a result, the model may say that a case is multi-source while still behaving as if it were single-source. When this happens, later stages do not repair the mistake; they inherit it.

\end{tcolorbox}
\subsection{Query reflection trapped in error clusters}
\begin{tcolorbox}[
    title=\textbf{Case Study: Query reflection trapped in error clusters},
    fonttitle=\bfseries,
    breakable,
    fontupper=\small
]

\label{app:pipeline_stage_b_query_reflection}

\paragraph{Overview.}
The second failure point is not the absence of search, but the inability of search to escape a wrong semantic cluster. Reflection may appear active, yet still remain trapped inside the same mistaken hypothesis.

\paragraph{Task Description.}
We illustrate this pattern with 2 representative examples.

\textbf{Example A: Reflection changes wording but not direction.}
For the same montage case, \texttt{gemini-3.1-pro-preview} kept refining the query:
\begin{quote}\small\texttt{``Cleo takes final year''}\end{quote}
\begin{quote}\small\texttt{``Cleo takes final year'' vlog -Abram''}\end{quote}
\begin{quote}\small\texttt{``Cleo takes final year'' morning routine uni vlog -Abram''}\end{quote}
\begin{quote}\small\texttt{``Cleo `final year' uni vlog South Africa -Abram''}\end{quote}
This looks like iterative improvement, but in reality all rounds stayed within the same wrong title-centered cluster. Reflection refined surface wording without changing the retrieval hypothesis.

\textbf{Example B: Queries drift into irrelevant object domains.}
With \texttt{qwen3.5-plus}, later rounds became increasingly driven by local cues:
\begin{quote}\small\texttt{``pink Bratz bedding vlog camping New Year 2021''}\end{quote}
\begin{quote}\small\texttt{``Cleo takes final year Campus Superior coffee camping''}\end{quote}
\begin{quote}\small\texttt{``echo chamber of our daily existence'' camping vlog''}\end{quote}
Here the model was not passive. It searched energetically, but along ever more distracting details, causing retrieval to drift into unrelated domains such as toys, coffee, or religious media.

\paragraph{Interpretation.}
This stage shows that reflection alone does not guarantee recovery. If the model lacks a mechanism for abandoning a bad hypothesis, it may simply optimize within that hypothesis, producing a sequence of increasingly specific but equally irrelevant queries.

\end{tcolorbox}

\subsection{Forgery extraction misaligned with ground truth.}
\begin{tcolorbox}[
    title=\textbf{Case Study: Forgery extraction misaligned with ground truth},
    fonttitle=\bfseries,
    breakable,
    fontupper=\small
]

\label{app:pipeline_stage_d_fine_extraction}

\paragraph{Overview.}
This is the most important error stage in the pipeline. In many difficult cases, the system has already found the correct source, yet still extracts the wrong forgery mechanism. It sees real differences, but explains them along the wrong axis.
\paragraph{Task Description.}
We illustrate this pattern with 4 representative examples.

\textbf{Example A: subtitles and specs instead of ranking rewrite.}
After finding the exact source of a tool-review video, \texttt{qwen3.5-plus} summarized the manipulation as:
\begin{quote}\small\texttt{``The primary forgery mechanism is Single-Source Editing via the addition of misleading and inaccurate text overlays.''}\end{quote}
This explanation is not empty, but it mistakes local textual corruption for the central manipulation. The actual forgery was that test order and comparative framing were reorganized to make one brand look like the winner.

\textbf{Example B: branding/context instead of causal restructuring.}
In a hydroponics review, \texttt{qwen3-vl-235b-a22b-think} matched the original source immediately, but still concluded:
\begin{quote}\small\texttt{``key branding and contextual elements are systematically removed or altered to misattribute the video to an unknown source''}\end{quote}
Again, this is coherent but off-target. The real manipulation was a timeline rearrangement that fabricated a false causal chain in which nutrient solution and app intervention appeared to ruin the plants.

\textbf{Example C: fictional product claims instead of comparative pacing bias.}
In a camera comparison case, \texttt{qwen3-vl-235b-a22b-instruct} summarized the manipulation as:
\begin{quote}\small\texttt{``AI-generated text (`17' references) and a fabricated 2025 article are spliced into the footage''}\end{quote}
This explanation centers on salient labels and future-dated claims, while missing the actual manipulation target: the reallocation of attention and screen time across a real comparison.

\textbf{Example D: Synthetic insertions are reconstructed as a plausible super-edit.}
In a laptop review case, \texttt{gemini-3-flash-preview} quickly found the correct source but then built a different story:
\begin{quote}\small\texttt{``Group A splices in external benchmark graphics from `Max Tech' ... and a segment featuring Linus Sebastian from `Linus Tech Tips'.''}\end{quote}
It ultimately concluded:
\begin{quote}\small\texttt{``The overall manipulation in Group A is a multi-source splicing forgery.''}\end{quote}
The explanation is internally self-consistent, but the inserted scenes were actually AI-generated synthetic footage rather than a multi-channel editorial mashup.

\paragraph{Interpretation.}
This stage shows the pipeline's central forensic weakness: finding the source is not the same as understanding the manipulation. Fine extraction is highly vulnerable to salience bias, local anomaly overfitting, and preference for familiar explanation templates such as subtitles, branding, or cross-source splicing.

\end{tcolorbox}

\subsection{Premature stopping}
\begin{tcolorbox}[
    title=\textbf{Premature stopping},
    fonttitle=\bfseries,
    breakable,
    fontupper=\small
]

\label{app:pipeline_stage_e_stop_decision}

\paragraph{Overview.}
The final stage decides whether the current evidence is already sufficient. This is a particularly fragile point, because once the model declares the case ``sufficient,'' the pipeline stops and the current hypothesis is effectively frozen. If the hypothesis is already wrong, the stop decision converts an intermediate mistake into the final answer.

\paragraph{Task Description.}
We illustrate this pattern with 3 representative examples.

\textbf{Example A: Early stopping locks in the wrong verification targets.}
In the hydroponics case, \texttt{qwen3-vl-235b-a22b-think} stopped after one round, claiming:
\begin{quote}\small\texttt{``The evidence collected ... covers all key verification targets''}\end{quote}
However, those targets were themselves misdefined around logo, interface, location, and attribution. The actual manipulation target --- the fabricated causal narrative --- had never been recovered. The stop decision therefore validated the wrong story.

\textbf{Example B: ``Source found'' becomes equivalent to ``forgery understood.''}
In the tool-review case, \texttt{qwen3.5-plus} stopped because:
\begin{quote}\small\texttt{``Since the source is from the official channel and matches frame-for-frame, further searching is unnecessary.''}\end{quote}
This reveals a structural weakness in the stopping criterion: source resolution is treated as if it implied manipulation coverage, even though a perfectly matched source can still coexist with a completely wrong forgery explanation.

\textbf{Example C: Stopping checks self-consistency, not correctness.}
In the laptop review case, \texttt{gemini-3-flash-preview} continued searching longer, but its final stopping logic still remained trapped in the wrong hypothesis:
\begin{quote}\small\texttt{``Without the Max Tech source video, I cannot confirm if the input is a Max Tech video with a swapped end card or an MKBHD video with swapped graphics.''}\end{quote}
This shows that the stop module does not ask whether the hypothesis is fundamentally off-task. It asks only whether that hypothesis has been developed enough to feel complete.

\paragraph{Interpretation.}
The stop stage is less a verifier of truth than a verifier of local narrative completeness. Once the model has constructed a sufficiently coherent story, the pipeline may terminate even if that story addresses the wrong manipulation. This makes early stopping a particularly high-leverage failure point in retrieval-based fake-video analysis.

\end{tcolorbox}

\begingroup
\setlength{\parskip}{0pt}
\section{Licenses and Terms of Use}
\label{sec:appendix_license}

EVID-Bench will be publicly released for non-commercial academic research under the Creative Commons Attribution-NonCommercial-ShareAlike 4.0 International (CC BY-NC-SA 4.0) license. To comply with the YouTube Terms of Service regarding media distribution, authentic source videos are provided as YouTube Video IDs and URLs rather than raw MP4 files. Source links may become unavailable over time; we therefore release a manifest with each sample's YouTube ID and collection timestamp, distribute the manipulated benchmark videos directly, and will maintain a mirror to improve long-term availability.

External models, tools, and datasets are governed by their respective licenses and terms of use. Generation tools include FaceFusion (MIT License) and Seedance~2.0~\cite{seedance2026seedance20advancingvideo} (subject to ByteDance's applicable terms). Open-weight Qwen models are used under their model licenses. Closed-source APIs used for generation filtering and evaluation---OpenAI GPT~\cite{singh2025openai}, Google Gemini~\cite{gemini3pro2025}, and Anthropic Claude~\cite{claudeops,claudeson}---are accessed under each provider's API Terms of Service and acceptable-use policies, for non-commercial academic research only.

\vspace{-\baselineskip}
\section{Content Filtering and Privacy Considerations}
\label{sec:appendix_content}

During the dataset construction and human review phase, we implemented strict filtering criteria. We ensured that no offensive, harmful, or Not Safe For Work (NSFW) content was included in the benchmark. Furthermore, we verified that the dataset does not expose sensitive Personally Identifiable Information (PII) of private individuals. However, strict visual anonymization (such as blurring faces or redacting on-screen text) was purposefully not applied for most forgery types. Because EVID-Bench evaluates evidence-dependent video forgery (e.g., contextual fabrication and brand manipulation), masking public figures, brand logos, or event details would destroy the essential visual evidence required for the cross-video reasoning task. All source materials are based on initially public footage from YouTube. Donor faces are taken solely from the public CelebA face dataset, not from private individuals.

\endgroup
\end{document}